\documentclass[letterpaper, 10 pt, journal, twoside]{IEEEtran} 

\IEEEoverridecommandlockouts                              

\usepackage[dvipsnames]{xcolor}
\definecolor{dark-green}{RGB}{12,80,12}

\usepackage{url}
\usepackage{dsfont}
\usepackage{amsmath}
\usepackage{amssymb}
\usepackage{cuted}
\usepackage{multirow} 
\usepackage{multicol}
\usepackage[shortlabels]{enumitem}
\newcommand{\tableref}[1]{Table~\ref{#1}}
\usepackage{tabularx}
\newcolumntype{C}[1]{>{\centering\let\newline\\\arraybackslash\hspace{0pt}}m{#1}} 
\newcolumntype{L}[1]{>{\let\newline\\\arraybackslash\hspace{0pt}}m{#1}} 

\usepackage{hyperref}
\usepackage[mathscr]{euscript}

\usepackage[export]{adjustbox}
\newcolumntype{P}[1]{>{\centering\arraybackslash}p{#1}}

\iffalse 
  \newcommand{\whyekit}[1]{\noindent}
  \newcommand{\holger}[1]{\noindent}
  \newcommand{\lubing}[1]{\noindent}
  \newcommand{\rohit}[1]{\noindent}
  \newcommand{\juana}[1]{\noindent}
  \newcommand{\abhi}[1]{\noindent}
  \newcommand{\todo}[1]{\noindent}
\else
  \newcommand{\whyekit}[1]{\textcolor{blue}{\bf [WF: #1]}}
  \newcommand{\holger}[1]{\textcolor{orange}{\bf [HC: #1]}}
  \newcommand{\lubing}[1]{\textcolor{purple}{\bf [LZ: #1]}}
  \newcommand{\rohit}[1]{\textcolor{brown}{\bf [RM: #1]}}
  \newcommand{\juana}[1]{\textcolor{ForestGreen}{\bf [HV: #1]}}
  \newcommand{\abhi}[1]{\textcolor{green}{\bf [AV: #1]}}
  \newcommand{\todo}[1]{\textcolor{red}{\bf [Todo: #1]}}
  
\fi

\usepackage{booktabs}
\usepackage{xspace}
\usepackage{caption}
\usepackage{graphicx} 
\usepackage{color}
\usepackage{siunitx}
\usepackage[hang,flushmargin,symbol]{footmisc}
\renewcommand{\thefootnote}{\fnsymbol{footnote}}
\usepackage{microtype}
\usepackage{rotating}
\newcommand{\secref}[1]{Sec.~\ref{#1}}
\renewcommand{\eqref}[1]{Eq.~(\ref{#1})}
\newcommand{\figref}[1]{Fig.~\ref{#1}}
\newcommand{\tabref}[1]{Tab.~\ref{#1}}

\usepackage{tabularx}
\usepackage{colortbl} 
\newcolumntype{Y}{>{\centering\arraybackslash}X}
\newcolumntype{Z}{>{\raggedleft\arraybackslash}X}

\usepackage{array}
\usepackage{multirow}
\usepackage{pifont}
\newcommand{\cmark}{\ding{51}}%
\newcommand{\xmark}{\ding{55}}%
\usepackage{cite}
\usepackage{flushend}
\usepackage[export]{adjustbox}
\usepackage[resetlabels]{multibib}
\newcites{New}{References}

\newcommand{\newmet}{PAT}

\DeclareSIUnit{\rad}{rad}

\renewcommand{\baselinestretch}{0.99}

\title{\LARGE \bf
Panoptic nuScenes: A Large-Scale Benchmark for\\LiDAR Panoptic Segmentation and Tracking
}

\author{Whye Kit Fong*$^{,1}$, Rohit Mohan*$^{,2}$, Juana Valeria Hurtado$^2$, Lubing Zhou$^1$,\\ Holger Caesar$^1$, Oscar Beijbom$^1$, and Abhinav Valada$^2$
\thanks{$^1$ Motional.\\
$^2$ Department of Computer Science, University of Freiburg, Germany.\\
* Authors contributed equally.}%
}

\begin{document}

\maketitle
\thispagestyle{empty}
\pagestyle{empty}

\begin{abstract}
Panoptic scene understanding and tracking of dynamic agents are essential for robots and automated vehicles to navigate in urban environments. As LiDARs provide accurate illumination-independent geometric depictions of the scene, performing these tasks using LiDAR point clouds provides reliable predictions. However, existing datasets lack diversity in the type of urban scenes and have a limited number of dynamic object instances which hinders both learning of these tasks as well as credible benchmarking of the developed methods. In this paper, we introduce the large-scale Panoptic nuScenes benchmark dataset that extends our popular nuScenes dataset with point-wise groundtruth annotations for semantic segmentation, panoptic segmentation, and panoptic tracking tasks. To facilitate comparison, we provide several strong baselines for each of these tasks on our proposed dataset. Moreover, we analyze the drawbacks of the existing metrics for panoptic tracking and propose the novel instance-centric PAT metric that addresses the concerns. We present exhaustive experiments that demonstrate the utility of Panoptic nuScenes compared to existing datasets and make the online evaluation server available at \url{nuScenes.org}. We believe that this extension will accelerate the research of novel methods for scene understanding of dynamic urban environments.\looseness=-1
\end{abstract}

\section{Introduction}

Autonomous vehicles~(AVs) have made significant strides in the past decade, in large part due to the advent of deep learning techniques and large-scale perception datasets.
%
The majority of the existing AV datasets~\cite{Geiger2013IJRR, caesar2020nuscenes, chang2019argoverse, houston2020one, sun2020scalability}, model objects as 3D bounding boxes with 7 degrees of freedom~(DoF) and constant size, tracked over time.
%
While bounding boxes form a good approximation for the shape of a car, they fail to tightly include extremities such as mirrors and antennas. Bounding boxes are also less suitable for other classes such as pedestrians with extended arms. When modeling the interactions of two pedestrians with bounding boxes, we lose most of the information about their spatial relation and contact points.
%
This is exacerbated by the common assumption that object size remains constant across a track, which is not the case for articulated objects like pedestrians.
%
Finally, the use of 7~DoF boxes means that the pitch of an object is not adjusted to the inclination of the road, which results in less accurate annotations.
%
Furthermore, bounding boxes are only suitable for foreground objects (\emph{thing}), but not for amorphous background regions (\emph{stuff}) such as road or grass.

In the LiDAR semantic segmentation task~\cite{thomas2019kpconv,wu2018squeezeseg, milioto2019rangenet++}, we label each LiDAR point with a semantic label rather than bounding boxes. This enables a much higher level of granularity. It also includes labels for \emph{stuff} classes that are missing in object detection and tracking.
However, the lack of instance labels means that the interaction between objects of the same class can still not be accurately modeled.
%
In LiDAR panoptic segmentation~\cite{behley2019semantickitti, sirohi2021efficientlps}, we assign each LiDAR point with a semantic label and each point belonging to \emph{thing} classes with an instance label. However, existing datasets and metrics ignore the temporal dimension by evaluating only on a single scan.
%
The novel panoptic tracking task~\cite{hurtado2020mopt} combines panoptic segmentation and tracking into a single coherent scene understanding problem.

\begin{figure}
    \centering
     \includegraphics[width=0.88\linewidth]{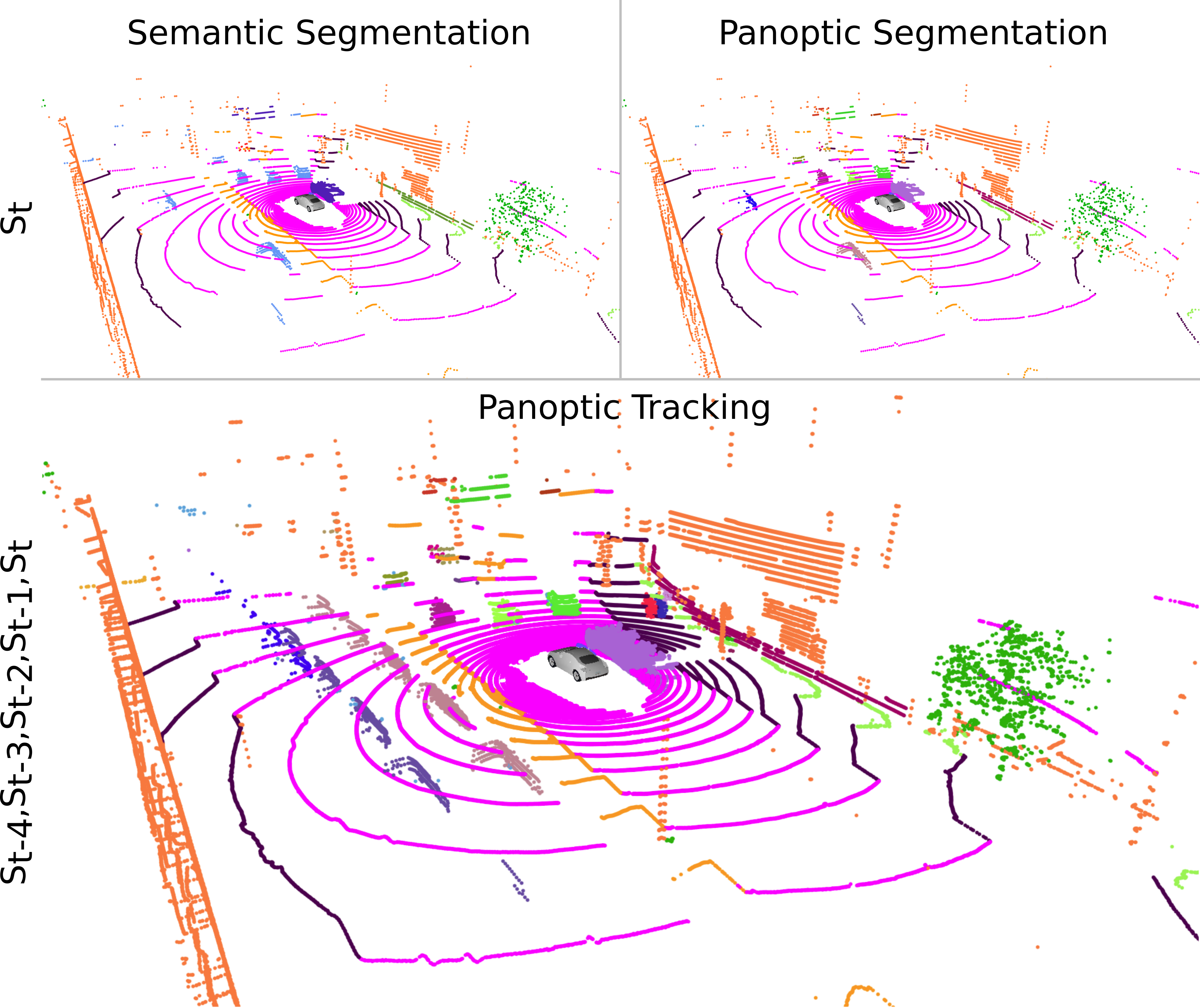}
     \caption{LiDAR scans from the Panoptic nuScenes dataset showing semantic segmentation labels with point-wise semantic class annotations, panoptic segmentation labels with additional instance IDs, and panoptic tracking labels with additional temporally consistent instance IDs. Note that in panoptic tracking, the same instances in consecutive scans have the same color indicating tracking.\looseness=-1}
     \label{fig:paper-teaser}
     \vspace{-0.3cm}
\end{figure}

\begin{table*}
    \footnotesize
    \centering
    \caption{A comparison of leading datasets for LiDAR segmentation. 
    We show the number of cities, number of semantic LiDAR segmentation classes, number of scan-wise moving and total instances (given in brackets), annotations types, availability of temporally consistent instance annotations, and the size of the dataset in hours. We state values for the full dataset where possible.
    ($^{\dagger}$) For the number of scan-wise moving and total instances, only the \emph{train-val} split is considered.
    ($^{\ddagger}$) For A2D2, the point-level instances were projected from the image to the point cloud.}
    \begin{tabular}{l|l c c c l c c c}
    \toprule
         Dataset                                      & Cities      & Classes & Scenes & Instances$^{\dagger}$   & Annotation            & Sequential & {Density} & Size \\
         \midrule
         Toronto-3D~\cite{tan2020toronto3d}           & 1x Canada   & 8       &  -     & -           & Point                 & \xmark  &   {-} & -     \\
         A2D2~\cite{geyer2020a2d2}                    & 3x Germany  & 38      &  -     & -           & Pixel$^{\ddagger}$, Box            & \xmark &   {-}     & -     \\
         PandaSet~\cite{pandaset}                                     & 2x USA      & 37      &  100   & -           & Point, Box            & \cmark    &   {-}  & 0.2 h \\
         SemanticKITTI~\cite{behley2021panoptickitti} & 1x Germany  & 28      &  22    & 23k (270k)  & Point, Instance      & \cmark  &   {122k}    & 1.5 h \\
         Panoptic nuScenes (ours)                     & Boston, SG  & 32      &  1000  & 300k (1.2M) & Point, Box, Instance & \cmark  &   {35k}    & 5.5 h \\
         \bottomrule
    \end{tabular}
    \vspace{-4mm}
    \label{tab:datasets}
\end{table*}

In this paper, we introduce Panoptic nuScenes, the first benchmark dataset for panoptic tracking\footnote[2]{An equivalent task~\cite{aygun20214d} was concurrently proposed on SemanticKITTI.}.
%
Many AV datasets only support a small number of AV tasks. Rather than developing a new dataset for each task, we extend the existing nuScenes~\cite{caesar2020nuscenes} dataset that already features object detection, bounding box tracking, and prediction tasks. We manually annotate 40,000 keyframes with 32 semantic classes, amounting to a total of 1.1B labeled LiDAR points.
%
The nuScenes dataset contains 1,000 scenes from 4 locations in Singapore and Boston. Contrary to the SemanticKITTI~\cite{behley2019semantickitti} dataset, nuScenes has been primarily collected in dense urban environments with many dynamic agents such as different vehicles and pedestrians. The AV that was used to collect the dataset was equipped with sensors that cover the full 360 degrees of the surroundings and include radar. Moreover, the nuScenes dataset is much more diverse with 1,000 scenes compared to 21 scenes that are present in SemanticKITTI. Furthermore, the semantic class annotations are more fine-grained with 23 compared to 7 \emph{thing} classes.\looseness=-1

Our main contributions in this work are as follows:
\begin{enumerate}[topsep=0pt]
    \item We manually annotate 1.1B points in 1,000 scenes with 23 object classes, resulting in the most diverse LiDAR segmentation dataset to date.
    \item We extend nuScenes to a holistic LiDAR benchmark that includes semantic segmentation, panoptic segmentation, and panoptic tracking tasks as illustrated in \figref{fig:paper-teaser}.
    \item We propose a new instance-level Panoptic Tracking (\newmet) metric that incorporates both panoptic quality and tracking quality while penalizing track fragmentation.
    \item We provide extensive baselines for each of the three tasks and present ablation studies that demonstrate the novelty of our benchmark dataset.
\end{enumerate}

The supplementary material can be found at \url{https://arxiv.org/abs/2109.03805}.

\section{Related Work}
\label{sec:relatedWork}

\noindent\textit{Datasets and Benchmarks}: LiDAR datasets for autonomous driving have contributed significantly to the progress in the industry. However, most of these datasets focus solely on 3D object detection rather than point-level segmentation which provides a more holistic and richer scene understanding. To fill this gap, recent driving datasets provide LiDAR point-level semantic annotations. However, they either do not provide instance annotations, have only a few moving object instances, or ignore the temporal dimension. \tabref{tab:datasets} provides an overview of relevant LiDAR semantic segmentation datasets. Overall, there is a lack of LiDAR datasets that provide annotations for panoptic segmentation and tracking. SemanticKITTI~\cite{behley2021panoptickitti} introduced a dataset that provides fine-grained semantic and temporally consistent instance annotations. 
It provides 1.5 hours of data recorded in a single city. 
Recently, Aygün~\textit{et~al.}~\cite{aygun20214d} extended this benchmark and proposed a point-centric evaluation metric for panoptic tracking~\cite{hurtado2020mopt}.

In this work, we annotate the significantly more diverse nuScenes~\cite{caesar2020nuscenes} dataset with LiDAR segmentation labels for 1.1B points. This dataset is referred to as \emph{nuScenes-lidarseg}.
Since the initial release of nuScenes-lidarseg, some works have attempted to create a panoptic dataset by combining the 3D bounding boxes from nuScenes~\cite{caesar2020nuscenes} and the LiDAR segmentation labels from nuScenes-lidarseg. 
DS-Net~\cite{hong2021dynamic} and EfficientLPS~\cite{sirohi2021efficientlps} assign instance labels to each point based on the instance of the bounding box it is inside of. 
EfficientLPS~\cite{sirohi2021efficientlps} ignores instances that have less than 15 points. PolarSeg-Panoptic~\cite{zhou2021panoptic} creates the panoptic instance annotation by assigning \emph{thing} object points to its closest 3D bounding box of the same label and removes outliers by omitting the \emph{thing} object points that are more than \SI{5}{\meter} from the centroid of the nearest bounding box.
Only instances with at least 20 points are evaluated.
Due to the lack of an official dataset and evaluation protocol, the performances reported by these works are not comparable. Moreover, none of these works have made their dataset public nor have they released a standardized evaluation protocol or public evaluation server, making it difficult to reproduce or build on their works. In this work, we build upon nuScenes and introduce \emph{nuScenes-lidarseg} and \emph{Panoptic nuScenes}. With Panoptic nuScenes, we introduce official panoptic annotations, a benchmarking protocol, and an open challenge for panoptic segmentation and tracking.

\noindent\textit{Methods}: There have been significant breakthroughs in \emph{LiDAR semantic segmentation} in recent years. Methods such as KPConv~\cite{thomas2019kpconv} and RandLA~\cite{hu2020randla} utilize an encoder-decoder architecture that directly operates on the point cloud whereas approaches such as Squeezeseg~\cite{wu2018squeezeseg} and RangeNet++~\cite{milioto2019rangenet++} use spherical projections of point clouds to enable usage of 2D CNN architectures. These methods typically employ a KNN-based post-processing step to account for reprojection errors. PolarNet~\cite{zhang2020polarnet} on the other hand, transforms the point cloud into polar BEV space to account for the imbalanced distribution of points. Cylinder3D~\cite{zhu2021cylindrical} opts for cylindrical partition and asymmetrical 3D convolution networks.

\textit{LiDAR panoptic segmentation and tracking} approaches are broadly categorized into proposal-free and proposal-based methods. Proposal-free methods~\cite{milioto2020lidar, gasperini2021panoster, aygun20214d} generally infer semantic segmentation before detecting instances through keypoint/center estimation or vote/learnable clustering in the projection or BEV domain. Whereas, proposal-based methods~\cite{sirohi2021efficientlps, hurtado2020mopt} first generate region proposals from encoded features and then detect the instances in parallel to perform semantic segmentation. EfficientLPS~\cite{sirohi2021efficientlps} introduces several novel network modules to incorporate 3D information explicitly into a 2D panoptic segmentation architecture. PanopticTrackNet~\cite{hurtado2020mopt} further builds upon EfficientPS~\cite{mohan2020efficientps} and adds a tracking head to obtain temporally consistent instance labels for panoptic tracking. The subsequently proposed 4D-PLS~\cite{aygun20214d} employs point clustering to effectively leverage the sequential nature of several consecutive LiDAR scans. Given that LiDAR panoptic segmentation and tracking are critical tasks for autonomous driving and they are considerably less explored than the image-based approaches, we believe that our large-scale benchmark dataset will encourage innovative research that addresses these tasks.\looseness=-1

\section{Dataset}
\label{sec:dataset}

In this section, we describe the Panoptic nuScenes dataset and the protocol that we employ for annotating the groundtruth for semantic segmentation, panoptic segmentation, and panoptic tracking tasks. Panoptic nuScenes is an extension of the nuScenes dataset~\cite{caesar2020nuscenes} which contains 1000 scenes collected across multiple cities with both left and right-hand driving, and several dynamic agents such as different moving vehicles and pedestrians. As a result, Panoptic nuScenes is large-scale and geographically diverse with point-level semantic and instance segmentation annotations that are temporally consistent. Moreover, the highly varied scenes and a large number of moving agents in each scene make it a challenging dataset suitable for benchmarking the panoptic tracking task.

\vspace{-2mm}
\subsection{Data Annotation}
\label{sec:annotation_protocol}
We annotate the Panoptic nuScenes dataset to consist of 32 semantic classes, with 23 \emph{thing} and 9 \emph{stuff} classes as shown in \figref{fig:panoptic_chart}. We manually annotate the semantic label of each LiDAR point. To reduce the annotation effort, we use the 3D boxes to initialize the semantic labels of the points for the \emph{thing} classes. Thereafter, we manually refine points that are included in multiple thing boxes or close to stuff points (e.g. vehicle wheels close to the ground plane). This approach eliminates a significant number of points and reduces the effort required to semantically annotate the remainder of the points, which belong to \emph{stuff} classes. 
We then perform multiple rounds of validation to achieve high-quality LiDAR segmentation annotations. As part of the process, we rendered camera frames with the corresponding point cloud and segmentation labels overlaid to ensure that the labels match up with what is seen in the camera view. We render these frames sequentially into a video to help identify temporal inconsistencies in the point-wise semantic labels during the review process.

We combine the point-level labels with the 3D bounding boxes from nuScenes~\cite{caesar2020nuscenes} to obtain instance labels for each point. An instance consists of the points that fall within a 3D bounding box and have the same segmentation type as the box. For bounding boxes that overlap, we resolve them by labeling the overlapping points as \emph{noise}. The percentage of points that are present in such overlapping regions is zero for 9 classes, and it is less than $0.8\%$ for all classes. This is even the case for classes such as `bendy bus` which is usually made up of multiple potentially overlapping bounding boxes. 
Using the track ID we ensure that instances are temporally consistent.

\vspace{-2mm}
\subsection{Dataset Analysis}

\begin{figure}
\centering
\includegraphics[width=\linewidth]{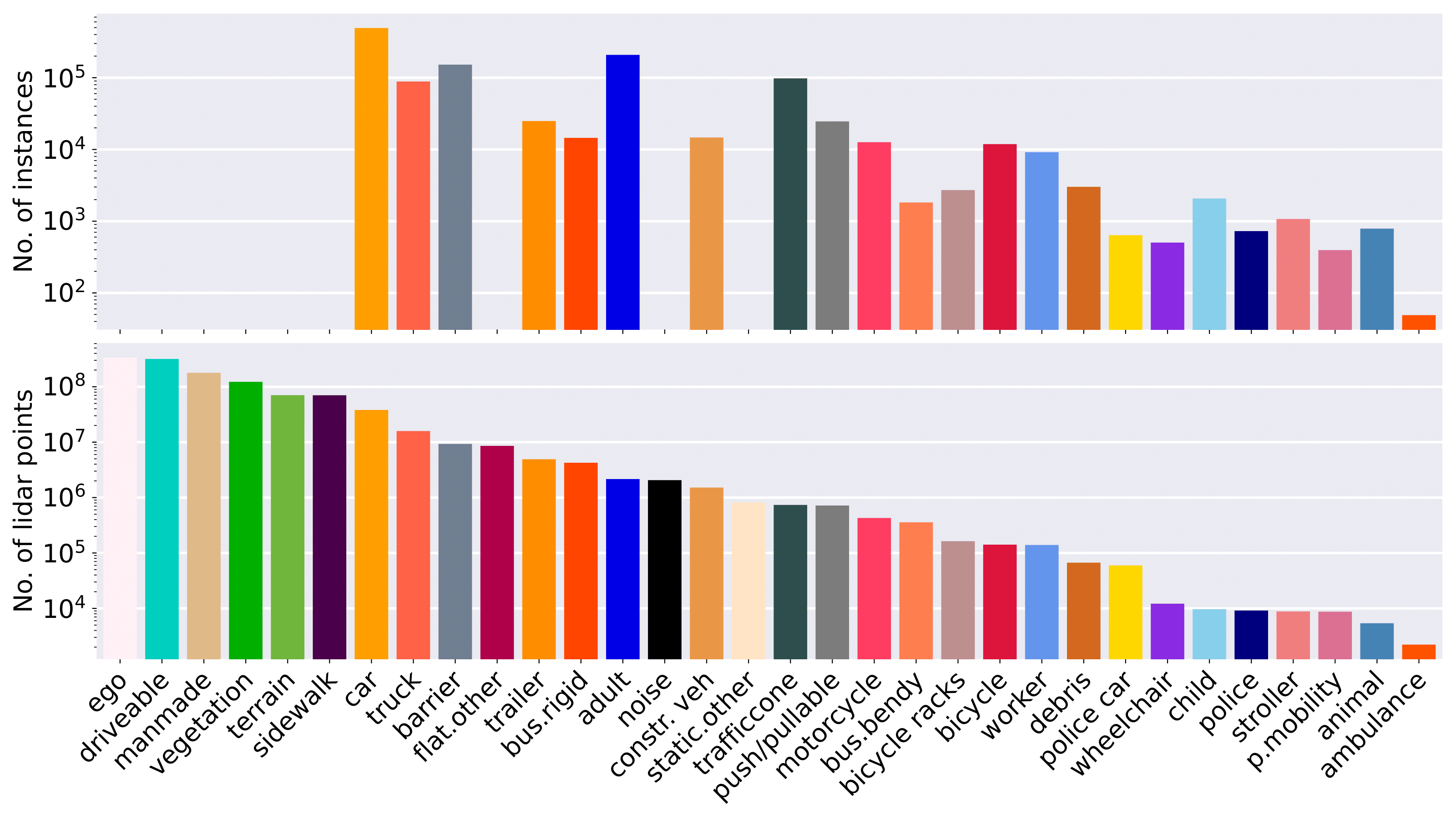}
\caption{(Bottom) Number of LiDAR points for each class in Panoptic nuScenes. 
(Top) Corresponding number of scan-wise instances for each \emph{thing} class.}
\label{fig:panoptic_chart}
\vspace{-0.2cm}
\end{figure}

\begin{figure}
\centering
\includegraphics[width=1.0\linewidth]{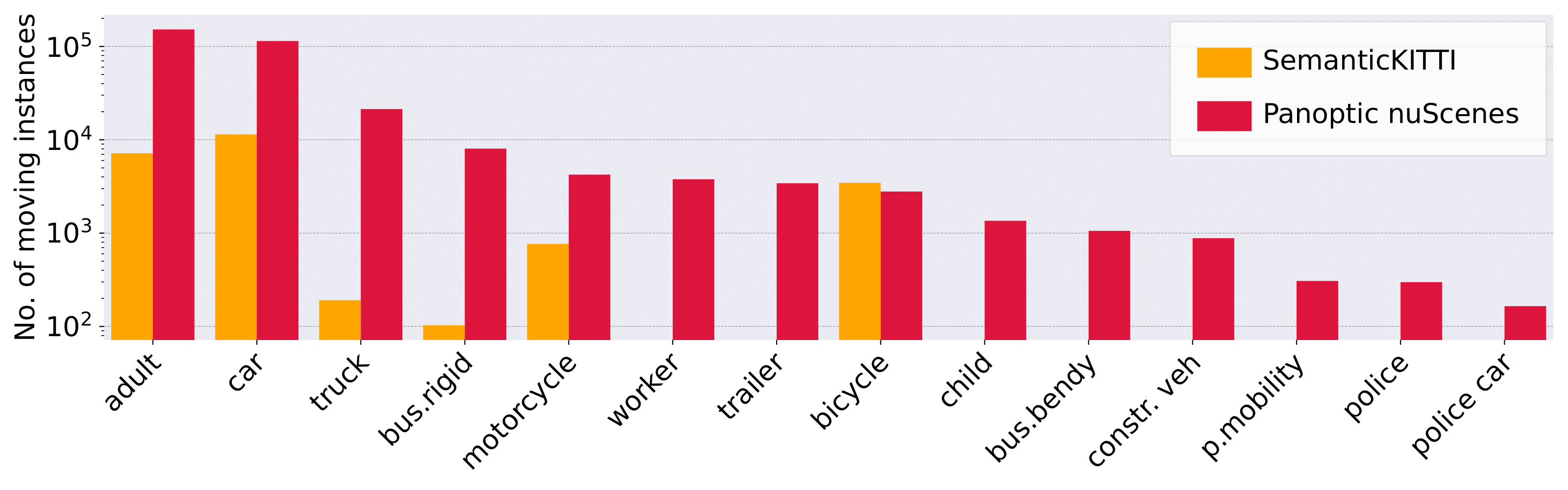}
\caption{Number of scan-wise moving instances in SemanticKITTI and Panoptic nuScenes. 
We only show bicycles/motorcycles with a rider as a proxy for the moving attribute. We exclude the rare on-rails and other\_vehicle classes.}
\label{fig:moving_instances_semkitti_nusc}
\vspace{-0.2cm}
\end{figure}

Our dataset contains 1.1B LiDAR points annotated with one of 32 semantic labels and temporally consistent instance IDs for thing classes. \figref{fig:panoptic_chart} shows the number of points for each semantic class in our dataset and the number of instances for the relevant classes. The most common \emph{stuff} classes are drivable surface, manmade, and vegetation, which can be useful for mapping and ground plane estimation~\cite{cattaneo2021lcdnet}. For the \emph{thing} classes, instances of dynamic object classes such as cars and adult pedestrians occur most frequently. On the other end of the distribution, we have rare classes such as police and ambulances, which represent the long-tail challenging categories. \figref{fig:moving_instances_semkitti_nusc} compares the number of moving object instances in SemanticKITTI~\cite{behley2021panoptickitti} and Panoptic nuScenes. 
With the exception of the bicycle class, Panoptic nuScenes contains significantly more moving object instances. Additionally, the variety of moving object classes in Panoptic nuScenes is also greater than SemanticKITTI~\cite{behley2021panoptickitti}.

\section{Tasks and Metrics}
\label{sec:tasks_metrics}

In this section, we describe the three benchmark tasks in Panoptic nuScenes and the corresponding evaluation metrics. For all the tasks, we only consider object instances that have more than 15 LiDAR points for evaluation. We also merge similar classes and remove rare or void classes, resulting in 10 \emph{thing} and 6 \emph{stuff} classes as shown in \tabref{table:class_map} (supplementary).

\subsection{LiDAR Semantic Segmentation}

Bounding box-based 3D object detection and tracking are two of the main perception tasks in autonomous driving. Due to the release of many large-scale public benchmark datasets~\cite{Geiger2013IJRR,caesar2020nuscenes,sun2020scalability}, the performance of methods that address these tasks has progressed rapidly. Similar to the evolution from 2d bounding box detection to semantic segmentation with images, 3d bounding box detection evolved to point-level LiDAR segmentation and has been attracting more and more attention in recent years. On one hand, there are numerous use cases for LiDAR segmentation in autonomous driving, such as drivable surface segmentation, ground plane prediction, and vehicle door segmentation. On the other hand, the emergence of better and faster segmentation models enables real-time point-level semantic segmentation. To further boost the development, we set up a LiDAR semantic segmentation benchmark using the Panoptic nuScenes dataset. The goal is to predict the semantic categories of each LiDAR point for both \emph{thing} and \emph{stuff} classes. The benchmark supports both LiDAR only and multi-modal methods. To enable a fair comparison, the benchmark is split into the LiDAR track and open track based on whether only the LiDAR or multi-modal data is used. More details can be found at \url{http://www.nuscenes.org/lidar-segmentation}. To evaluate the performance of semantic segmentation methods in the benchmark, we primarily use the Intersection-over-Union (IoU) metric and rank methods based on the average IoU (mIoU) over all the classes. We also report the frequency-weighted IoU (fwIoU) that weights the relevance of each class depending on its point-level frequency.

\subsection{LiDAR Panoptic Segmentation}

While the semantic segmentation task requires predicting the semantic categories for each point, it does not distinguish different object instances. Similar to image panoptic segmentation~\cite{kirillov2019panoptic}, the LiDAR panoptic segmentation task requires predicting the semantic categories for each point as well as instance IDs for \emph{thing} classes. The panoptic segmentation benchmark also consists of the LiDAR track and open track. More details can be found at \url{http://www.nuscenes.org/panoptic}. We primarily use the panoptic quality (PQ)~\cite{kirillov2019panoptic} metric to evaluate the panoptic segmentation performance in the benchmark. We individually compute the performance for \emph{thing} (PQ$^{\text{Th}}$,SQ$^{\text{Th}}$, RQ$^{\text{Th}}$) and \emph{stuff} (PQ$^{\text{St}}$, SQ$^{\text{St}}$, RQ$^{\text{St}}$) classes, and also report the PQ$^\dagger_\text{c}$~\cite{porzi2019seamless} metric for completeness. 

\subsection{LiDAR Panoptic Tracking}
\label{sec:panoptic_tracking}
While panoptic segmentation resolves the point-level and instance-level prediction of the whole scene, another critical aspect for autonomous driving is the temporally consistent perception of the environment. The recently introduced panoptic tracking task~\cite{hurtado2020mopt} addresses this problem by unifying panoptic segmentation and multi-object tracking into a single coherent scene understanding task. The goal of this task is to predict temporally consistent semantic categories of each point and instance IDs for \emph{thing} classes. While panoptic segmentation focuses on static frames, panoptic tracking additionally enforces temporal coherence and pixel-level association over time. Due to the lack of existing public benchmarks, we introduce a LiDAR panoptic tracking benchmark. Similar to the other tasks, the panoptic tracking benchmark also consists of the LiDAR track and open track. More details can be found at \url{http://www.nuscenes.org/panoptic}.\looseness=-1

Evaluating LiDAR panoptic tracking requires assessing the performance of both panoptic segmentation and instance association across frames. Unlike panoptic segmentation, for which we use the well-established PQ metric, evaluating multi-object tracking (MOT) has typically been challenging. Simultaneously measuring detection and association can lead to a higher influence of either detection or association in the metric. Moreover, measuring the performance of panoptic tracking is even more challenging since it combines panoptic segmentation and MOT.\looseness=-1

Recent works have proposed different metrics to evaluate this task. Hurtado~\textit{et~al.}~\cite{hurtado2020mopt} propose panoptic tracking quality (PTQ) that adapts PQ to account for the incorrectly tracked objects. Weber~\textit{et~al.}~\cite{weber2021step} analyzes the drawbacks of PTQ and demonstrates that it assigns more weight to segmentation than association, as it depends on the correct segmentation to assess tracking quality. To deal with this imbalance, Aygun~\textit{et~al.}~\cite{aygun20214d} adapt the STQ metric to LiDAR Segmentation and Tracking Quality (LTSQ). LSTQ disentangles semantics and temporal association and avoids penalizing tracking when an instance is incorrectly classified. LSTQ is computed at the point-level and evaluates if each point is assigned the right semantic class and if \emph{thing} class objects are assigned the right instance IDs. However, LSTQ also does not account for temporal information and is invariant to frame permutation. Importantly, it does not penalize the score in case a tracking prediction is fragmented. In the context of autonomous driving, penalizing track fragmentation is more important than the correct point-level spatio-temporal prediction. A fragmented track prediction can lead to imprecise velocity estimates. Even if the segmentation is not entirely accurate and some points are misclassified, tracking the entire moving object instance is essential for decision making. Although point-level evaluation of this task enriches the granularity of segmentation, measuring the performance of tracking at the instance-level, which is robust to track fragmentation, is more suitable for certain applications. 

With this in mind and based on the requirements of a good metric identified in \cite{weber2021step}, we define a new instance-level metric with the following features:
\begin{enumerate}
    \item Error-type differentiability: Allows separate analysis of tracking and panoptic segmentation components.
    \item Penalize fragmentation of tracking: The metric is not invariant to frame permutation.
    \item Long-term track consistency: The metric promotes long-term track consistency considering Association Recall.
    \item Penalize ID transfer: The metric can detect association errors accounting for Association Precision.
    \item Penalize the removal of correctly segmented instances with incorrect IDs: Metric robustness.
\end{enumerate}

\begin{table*}
\setlength\tabcolsep{3.0pt}
\centering
\caption{Comparison of LiDAR semantic segmentation performance on the Panoptic nuScenes dataset. All scores are in [\%].}
\label{tab:NuscenesSemantic}
\footnotesize
\begin{tabular}
{ll|cccccccccc|cccccc|cc}
\toprule
& Method & \begin{sideways}barrier\end{sideways} & \begin{sideways}bicycle\end{sideways} & \begin{sideways}bus\end{sideways} & \begin{sideways}car\end{sideways} & \begin{sideways}cvehicle\end{sideways} & \begin{sideways}motorcycle\end{sideways} & \begin{sideways}pedestrian\end{sideways} & \begin{sideways}traffic cone\end{sideways} & \begin{sideways}trailer\end{sideways} & \begin{sideways}truck\end{sideways} & \begin{sideways}driveable\end{sideways} & \begin{sideways}other flat\end{sideways} & \begin{sideways}sidewalk\end{sideways} & \begin{sideways}terrain\end{sideways} & \begin{sideways}man-made\end{sideways} & \begin{sideways}vegetation\end{sideways} & mIoU & fwIoU \\
\toprule
\multirow{7}{*}{\rotatebox[origin=c]{90}{val set}}
& RangeNet++~\cite{milioto2019rangenet++} & 66.0 & 21.3 & 77.2 & 80.9 & 30.2 & 66.8 & 69.6 & 52.1 & 54.2 & 72.3 & 94.1 & 66.6 & 63.5 & 70.1 & 83.1 & 79.8 & 65.5 & {83.0}\\
& EfficientLPS~\cite{sirohi2021efficientlps} & 72.6 & 29.3 & 81.2 & 81.3 & 28.0 & 60.1 & 66.4 & 57.8 & 50.3 & 66.6 & 94.9 & 69.2 & 73.4 & 72.6 & 86.7 & 85.1 & 67.3 & 86.0\\
& PolarNet~\cite{zhang2020polarnet} &  74.7 & 28.2 & 85.3 & 90.9 & 35.1 & 77.5 & 71.3 & 58.8 & 57.4 & 76.1 & 96.5 & 71.1 & 74.7 & 74.0 & 87.3 & 85.7 & 71.0 & {87.6}\\
& AMVNet~\cite{liong2020amvnet} & \textbf{79.8} & 32.4 & 82.2 & 86.4 & \textbf{62.5} & \textbf{81.9} & 75.3 & \textbf{72.3} & \textbf{83.5} & 65.1 & \textbf{97.4} & 67.0 & \textbf{78.8} & 74.6 & \textbf{90.8} & \textbf{87.9} & \textbf{76.1} & \textbf{89.5} \\
& Cylinder3D~\cite{zhu2021cylindrical} & 76.4 & \textbf{40.3} & \textbf{91.2} & \textbf{93.8} & 51.3 & 78.0 & \textbf{78.9} & 64.9 & 62.1 & \textbf{84.4} & 96.8 & \textbf{71.6} & 76.4 & \textbf{75.4} & 90.5 & 87.4 & \textbf{76.1} & {89.3}\\
\midrule
\multirow{5}{*}{\rotatebox[origin=c]{90}{test set}} & EfficientLPS~\cite{sirohi2021efficientlps} &  77.0 & 20.9 & 69.0 & 78.8 & 40.0 & 66.3 & 64.6 & 60.8 & 72.0 & 57.6 & 95.6 & 65.3 & 76.4 & 72.9 & 87.5 & 86.1 & 68.2 & 86.8\\
& PolarNet~\cite{zhang2020polarnet} &  80.1 & 19.9 & 78.6 & 84.1 & 53.2 & 47.9 & 70.5 & 66.9 & 70.0 & 56.7 & 96.7 & 68.7 & 77.7 & 72.0 & 88.5 & 85.4 & 69.8 & 87.7 \\
& AMVNet~\cite{liong2020amvnet} & 80.6 & 32.0 & 81.7 & 88.9 & 67.1 & \textbf{84.3} & 76.1 & \textbf{73.5} & \textbf{84.9} & 67.3 & 97.5 & 67.4 & 79.4 & \textbf{75.5} & \textbf{91.5} & \textbf{88.7} & 77.3 & 89.7 \\
& Cylinder3D~\cite{zhu2021cylindrical} & \textbf{82.8} & 33.9 & 84.3 & \textbf{89.4} & 69.6 & 79.4 & \textbf{77.3} & 73.4 & 84.6 & 69.4 & \textbf{97.7} & \textbf{70.2} & \textbf{80.3} & \textbf{75.5} & 90.4 & 87.6 & 77.9 & \textbf{89.9} \\
& (AF)\textsuperscript{2}-S3Net~\cite{cheng20212} & 78.9 & \textbf{52.2} & \textbf{89.9} & 84.2 & \textbf{77.4} & 74.3 & \textbf{77.3} & 72.0 & 83.9 & \textbf{73.8} & 97.1 & 66.5 & 77.5 & 74.0 & 87.7 & 86.8 & \textbf{78.3} & 88.5\\
\bottomrule
\end{tabular}
\vspace{-0.2cm}
\end{table*}

Accordingly, we introduce the Panoptic Tracking (\newmet)~metric based on two separable components that are explicitly related to the task and allow straightforward interpretation. \newmet~is computed as the harmonic mean of PQ and TQ as
\begin{equation}
    \newmet =  \frac{2 \times PQ \times TQ}{PQ + TQ}   \in [0,1],
\end{equation}
where PQ is the panoptic quality~\cite{kirillov2019panoptic} and TQ is our proposed tracking quality that we compute as the following class agnostic geometric mean for \emph{thing} classes:
\begin{align}
TQ(g) &= \sum([1-\frac{IDS(g)}{N_{IDS}(g)}]\times AS(g))^{\frac{1}{2}} , \label{eq:track_quality_1} \\
TQ &= \frac{1}{gt_{tracks}}\sum_{g\in gt_{tracks}} TQ(g), \in [0,1], \label{eq:track_quality_2}
\end{align}
where $gt_{tracks}$ is the set of unique instance IDs. TQ is comprised of two components. First is the association score $AS(g)$ that we define by adapting $S_{assoc}$~\cite{aygun20214d} to compute it at the instance-level. We compute $AS(g)$ using the true positive association ($TPA$), false negative association ($FNA$), and false positive association ($FPA$) sets for \emph{thing} classes. More precisely, $TPA$ is a set between a $gt_{tracks}$ of an object $g$  and any other object prediction with ID $p$ that have their mask overlap greater than $0.5$ IoU. Subsequently, $FNA$ is a set of $g$ $gt_{tracks}$ that has no matching predictions or have matching predictions with a ID other than $p$, and $FPA$ is a set of $p$ predictions that has no matching $gt_{tracks}$ or is matched with some $g'$ $gt_{tracks}$ instead of $g$. $AS(g)$ is then defined as
\begin{align}
    AS(g) &= \frac{1}{|g|}\sum_{p,|p \cap g|\ne 0}TPA(p,g)\times IoU_{a}(p,g), 
\end{align}
where the $IoU_{a}$ is obtained with $TPA$, $FNA$ and $FPA$.
This component of TQ encourages long-term track consistency and penalizes the removal of correctly segmented instances. The second component of TQ penalizes track fragmentation for $gt_{tracks}$ $g$ with the rate of ID switches where $IDS$ is the number of ID switches, and $N_{IDS}$ is the maximum possible ID switches over $|g|$ track length. We define an ID switch when either of the two consecutive frames of the instance $g$ does not have a matching prediction or are associated with predictions with inconsistent IDs. In \secref{subsec:new_metric_analysis} we present detailed analysis of our proposed metric that proves our proposed metric satisfies all the outlined requirements of being a good metric at instance-level.\looseness=-1

\section{Experimental Evaluation}
\label{sec:experiments}

In this section, we present extensive quantitative comparisons and benchmarking results for the semantic segmentation, panoptic segmentation and panoptic tracking tasks defined in \secref{subsec:baseline_results}. We then present a detailed analysis of our \newmet~metric for panoptic tracking in \secref{subsec:new_metric_analysis} and ablation studies that demonstrate the utility of our proposed dataset in \secref{subsec:ablation}. We present additional details and analysis in the supplementary. 

\subsection{Baseline Results}
\label{subsec:baseline_results}

We follow standard protocols to create/select baselines for the three tasks. In the case of LiDAR semantic segmentation, we frequently organize challenges for the task on Panoptic nuScenes and thus, report the test set results of the published challenge submissions as the baselines. However, for the validation set, we report the results of approaches from their respective published sources. For panoptic segmentation and panoptic tracking, we create two groups of baselines: end-to-end approaches and an independent combination of task-specific methods that are benchmarked on our challenge server. We trained the end-to-end approaches using the official implementations that have been publicly released by the authors after further tuning of hyperparameters to the best of our ability and report their results for both validation and test set.

\subsubsection{Semantic Segmentation}

Results for the semantic segmentation benchmark are shown in \tabref{tab:NuscenesSemantic}. We observe that \emph{stuff} classes such as manmade and drivable surface are comparatively easier to segment, achieving an IoU of about 90\%. Additionally, these classes generally exhibit negligible IoU degradation with variation in distance from the ego vehicle. Among the \emph{thing} classes, bicycle appears to be the hardest class as only a few approaches achieve an IoU greater than 40\%. (AF)\textsuperscript{2}-S3Net~\cite{cheng20212} achieves the highest bicycle class IoU of 52.2\% and an overall mIoU of 78.3\% due to the novel multi-branch attentive feature fusion module in the encoder and a unique adaptive feature selection module with feature map re-weighting in the decoder that enables more accurate segmentation of smaller instances.

\begin{table*}
\centering
\caption{Comparison of LiDAR panoptic segmentation performance on the Panoptic nuScenes dataset. All scores are in [\%].}
\label{tab:nuscenesPanoptic}
\footnotesize
\begin{tabular}
{ll|cccc|ccc|ccc|c}
\toprule
& Method & PQ & PQ$^\dagger$  & SQ & RQ & PQ\textsuperscript{Th} & SQ\textsuperscript{Th} & RQ\textsuperscript{Th} & PQ\textsuperscript{St} & SQ\textsuperscript{St} & RQ\textsuperscript{St} & mIoU \\
\toprule
\multirow{3}{*}{\rotatebox[origin=c]{90}{val set}} & PanopticTrackNet~\cite{hurtado2020mopt} & 51.4 & 56.2 & 80.2 & 63.3 & 45.8 & 81.4 & 55.9 & 60.4 & 78.3 & 75.5 & 58.0 \\
& EfficientLPS~\cite{sirohi2021efficientlps} & 62.0 & 65.6 & 83.4 & 73.9 & 56.8 & 83.2 & 68.0 & 70.6 & 83.8 & 83.6 & 65.6 \\
& PolarSeg-Panoptic~\cite{zhou2021panoptic} & \textbf{63.4} &  \textbf{67.2} & \textbf{83.9} &  \textbf{75.3} &  \textbf{59.2} &  \textbf{84.1} &  \textbf{70.3} &  \textbf{70.4} &  \textbf{83.6} &  \textbf{83.5} &  \textbf{66.9} \\
\midrule
\multirow{6}{*}{\rotatebox[origin=c]{90}{test set}} & PanopticTrackNet~\cite{hurtado2020mopt} & 51.6 & 56.1 & 80.4 & 63.3 & 45.9 & 81.4 & 56.1 & 61.0 & 79.0 & 75.4 & 58.9 \\
& EfficientLPS~\cite{sirohi2021efficientlps} & 62.4 & 66.0 & 83.7 & 74.1 & 57.2 & 83.6 & 68.2 & 71.1 & 83.8 & 84.0 & 66.7 \\
& PolarSeg-Panoptic~\cite{zhou2021panoptic} & 63.6 & 67.1 & 84.3 & 75.1 & 59.0 & 84.3 & 69.8 & 71.3 & 84.2 & 83.9 & 67.0 \\
& SPVNAS~\cite{tang2020searching} + CenterPoint~\cite{yin2021center} & 72.2 & 76.0 & 88.5 & 81.2 & 71.7 & 89.7 & 79.4 & 73.2 & 86.4 & 84.2 & 76.9 \\
& Cylinder3D++~\cite{zhu2021cylindrical} + CenterPoint~\cite{yin2021center} & 76.5 & 79.4 & \textbf{89.6} & 85.0 & 76.8 & 91.1 & 84.0 & \textbf{76.0} & \textbf{87.2} & \textbf{86.6} & 77.3 \\
& (AF)\textsuperscript{2}-S3Net~\cite{cheng20212} + CenterPoint~\cite{yin2021center} & \textbf{76.8} & \textbf{80.6} & 89.5 & \textbf{85.4} & \textbf{79.8} & \textbf{91.8} & \textbf{86.8} & 71.8 & 85.7 & 83.0 & \textbf{78.8} \\
\bottomrule
\end{tabular}
\vspace{-0.2cm}
\end{table*}

\begin{table*}
\centering
\caption{Comparison of LiDAR panoptic tracking performance on the Panoptic nuScenes dataset. All scores are in [\%].}
\label{tab:nuscenesTrack}
\footnotesize
\begin{tabular}
{ll|ccc|cccc}
\toprule
& Method & {\textbf{\newmet}} & LSTQ & PTQ & PQ & TQ & S$_{\text{cls}}$ & S$_{\text{assoc}}$ \\
\toprule
\multirow{3}{*}{\rotatebox[origin=c]{90}{val set}} & PanopticTrackNet~\cite{hurtado2020mopt} & 44.0 & 43.4 & 50.9 
& 51.6 & 38.5 & 58.4 & 32.3\\
& 4D-PLS~\cite{aygun20214d}& 59.2 & 56.1 & 55.5 
& 56.3 & 62.3 & 61.2 & 51.4 \\
& EfficientLPS~\cite{sirohi2021efficientlps} + Kalman Filter & \textbf{64.6} & \textbf{62.0} & \textbf{60.6} & \textbf{62.0} & \textbf{67.6} & \textbf{65.6} & \textbf{58.6} \\
\midrule
\multirow{6}{*}{\rotatebox[origin=c]{90}{test set}} & PanopticTrackNet~\cite{hurtado2020mopt} & 45.7 & 44.8 & 51.6 & 51.7 & 40.9 & 58.9 & 36.7 \\
& 4D-PLS~\cite{aygun20214d} & 60.5 & 57.8 & 55.6 & 56.6 & 64.9 & 62.3 & 53.6 \\
& AMVNet~\cite{liong2020amvnet} + OGR3MOT~\cite{zaech2021ogr3mot} & 63.2 & 61.7 & 61.5 & 61.9 & 64.7 & 63.6 & 59.9 \\
& Cylinder3D++~\cite{zhu2021cylindrical} + OGR3MOT~\cite{zaech2021ogr3mot} & 62.7 & 61.7 & 61.3 & 61.6 & 63.8 & 64.0 & 59.4 \\
& (AF)\textsuperscript{2}-S3Net~\cite{cheng20212} + OGR3MOT~\cite{zaech2021ogr3mot} & 62.9 & 62.4 & 60.9 & 61.3 & 64.5 & 65.0 & 59.9 \\
& EfficientLPS~\cite{sirohi2021efficientlps} + Kalman Filter & \textbf{67.1} & \textbf{63.7} & \textbf{62.3} & \textbf{63.6} & \textbf{71.2} & \textbf{67.4} & \textbf{60.2} \\
\bottomrule
\end{tabular}
\vspace{-0.2cm}
\end{table*}

\subsubsection{Panoptic Segmentation}

To create the combination of task-specific baselines for panoptic segmentation, we merge submissions from the past nuScenes LiDAR segmentation and detection challenges. We do so, by assigning unique instance IDs to all points lying within the predicted bounding boxes along with the semantic class ID of the box. Further, we apply a heuristic similar to \cite{kirillov2019panoptic} to resolve bounding box overlaps. While combining the detection predictions, we investigated the filtering of low confidence detection boxes based on the max F1 score per class or different threshold values. Interestingly, applying no filtering on the detected boxes achieves the highest performance for each method, on average improving the PQ score by $0.62\%$ for each class compared to F1 thresholding.

By combining 70 detection submissions and 21 semantic segmentation submissions, we generate a total of 1470 independently combined panoptic segmentation baselines as shown in \figref{fig:mIoU_mAP_AMOTA_PQ_LSTQ}(a) of the supplementary material and compare it with three state-of-the-art end-to-end approaches~\cite{hurtado2020mopt, sirohi2021efficientlps, zhou2021panoptic}. In the former, we observe a general trend of baselines with combinations of higher-performing detection and semantic segmentation methods achieving higher PQ scores, with more emphasis on stronger detection approaches. This can be attributed to the fact that the detection methods are primarily responsible for the instance segmentation performance. \tabref{tab:nuscenesPanoptic} presents the benchmarking results for this task where we report only the top three combination baselines that have published sub-task methods. Among the end-to-end baselines, PolarSeg-Panoptic achieves the highest PQ score of $63.4\%$ indicating that the polar bird's-eye-view representation is more suitable for LiDAR panoptic segmentation compared to other projection-based baselines. 
Overall, the independently combined models significantly outperform the end-to-end approaches, demonstrating the need for more research on end-to-end LiDAR panoptic segmentation methods. Thus, we believe that our challenging benchmark will pave the way for many innovative solutions.\looseness=-1  

\subsubsection{Panoptic Tracking}

\begin{figure}
\centering
\includegraphics[width=0.7\linewidth]{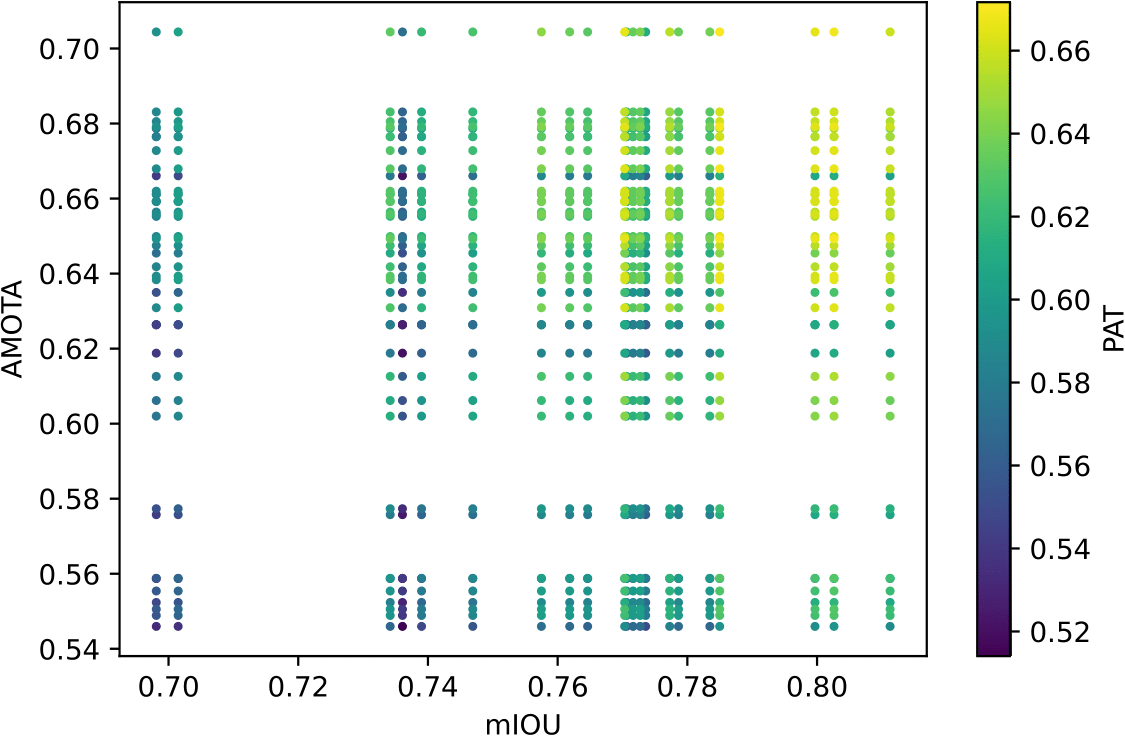}
\caption{Analysis of how the PAT metric measures the performance with different combinations of LiDAR semantic segmentation and tracking methods. Each dot represents a single combination. A total of 924 combinations are shown.}
\label{fig:mIoU_amota_PAT}
\vspace{-0.2cm}
\end{figure}

Following a similar protocol for creating baselines, for panoptic tracking, we generate a total of 924 task-specific combination baselines by merging 21 submissions from the nuScenes LiDAR segmentation challenge and 44 submissions from the tracking challenge. Results from this experiment are presented in \figref{fig:mIoU_amota_PAT}. Note that three \emph{thing} classes (barrier, construction vehicle, and traffic cone) are not included in the tracking challenge and hence are absent in the merged panoptic tracking prediction. Essentially, the contribution of these classes is counted as zero in the relevant evaluation metrics. On the other hand, for the unified end-to-end~\cite{hurtado2020mopt, aygun20214d} approaches, we train inclusive of all classes as they will serve as the main baselines for any future work in this task. We observe that the task-specific combination baselines are positively correlated to the performance of both semantic segmentation and tracking methods, similar to the observation in the panoptic segmentation results. \tabref{tab:nuscenesTrack} presents the benchmarking results for this task where we report only the top three combination baselines that have published methods. Among all the baselines, EfficientLPS with Kalman Filter achieves the highest \newmet~score of $64.6\%$ and $67.1\%$ on the validation and test set respectively. This shows that the initialization of initial and noise state covariance matrices in the filter with training set statistics is crucial for the filter's convergence which yields accurate data associations. We also observe that the unified approaches are outperformed by the task-specific combination baselines. We expect that our benchmark dataset will accelerate research on unified end-to-end approaches that will surpass the performance of the independently combined baselines.

\subsection{Analysis of Panoptic Tracking Metrics}
\label{subsec:new_metric_analysis}

To evaluate the effectiveness of our proposed \newmet~metric for panoptic tracking, we consider five challenging tracking scenarios, each corresponding to the metric requirements described in \secref{sec:panoptic_tracking}. First, we analyze the capability of the metric to decouple panoptic and tracking errors. Subsequently, we analyze four challenging tracking scenarios or cases in which we assume perfect segmentation for seven consecutive scans and study the performance of the tracking metrics. \tabref{tab:metric} illustrates the four cases in the figure and the adjacent table presents the tracking scores comparing \newmet~with the previous proposed LSTQ and PTQ metrics.

\subsubsection{Error-Type Differentiability}

In \tabref{tab:nuscenesTrack}, we present the panoptic tracking results of three baselines with the same tracking method OGR3MOT combined with different panoptic segmentation methods AMVNet, (AF)\textsuperscript{2}-S3Net, and Cylinder3D++. The ranking of the methods differs between \newmet~and LSTQ. While LSTQ only considers segmentation quality with mIoU, our metric also accounts for instance identification by means of PQ. Since the panoptic tracking task aims to predict \emph{stuff} and \emph{thing} classes in the scene while preserving the \emph{thing} IDs across frames, we consider \newmet~to be more informative for the task. In this case, our metric yields a higher score for the method with the better panoptic segmentation performance, which facilitates interpretability. Additionally, we observe in \tabref{tab:correlations} that the correlation of AMOTA and mIOU to \newmet~is 0.48 and 0.57. This represents an improvement with respect to PTQ (0.23 and 0.69), implying that similar to LSTQ (0.55 and 0.62), \newmet~is a more balanced metric.\looseness=-1


\begin{table}
\centering
\caption{Illustration of four challenging tracking cases (figure) along with their corresponding evaluation scores using different metrics (table). We assume correct LiDAR instance segmentation in seven consecutive scans where the color of the vehicle depicts the instance ID. Our \newmet~is the only metric that successfully addresses all the four tracking challenges.}
\label{tab:metric}
\footnotesize
\begin{tabular}{l|p{0.3cm}p{0.3cm}c| p{0.3cm}c| p{0.3cm}p{0.3cm}c| p{0.3cm}p{0.3cm}c}
\toprule
Metric &  \multicolumn{10}{c}{Prediction} \\
\cmidrule{2-12}
 & \multicolumn{3}{c|}{Case 1} & \multicolumn{2}{c|}{Case 2} & \multicolumn{3}{c|}{Case 3} & \multicolumn{3}{c}{Case 4} \\
\cmidrule{2-12}
& 1 & 2 &   & 3 &  & 4 & 5 &  & 6 & 7 &   \\
\toprule
{\textbf{\newmet}} & 66.7 & 89.6 & \cmark & 81.6 & \cmark & 92.3 & 89.6 & \cmark & 89.6 & 76.8 & \cmark\\
LSTQ & 87.2 & 87.2 & \xmark & 70.7 & \cmark & 91.6 & 87.2 & \cmark & 87.2 & 69.7 & \cmark\\
PTQ & 42.9 & 86.7 & \cmark & 100.0 & \xmark & 85.7 & 85.7 & \xmark & 85.7 & 83.3 & \cmark\\
\bottomrule
\end{tabular}
\begin{tabular}{p{1cm}}
     ~\vspace{0.2cm}\\
\end{tabular}
\includegraphics[width=0.9\linewidth]{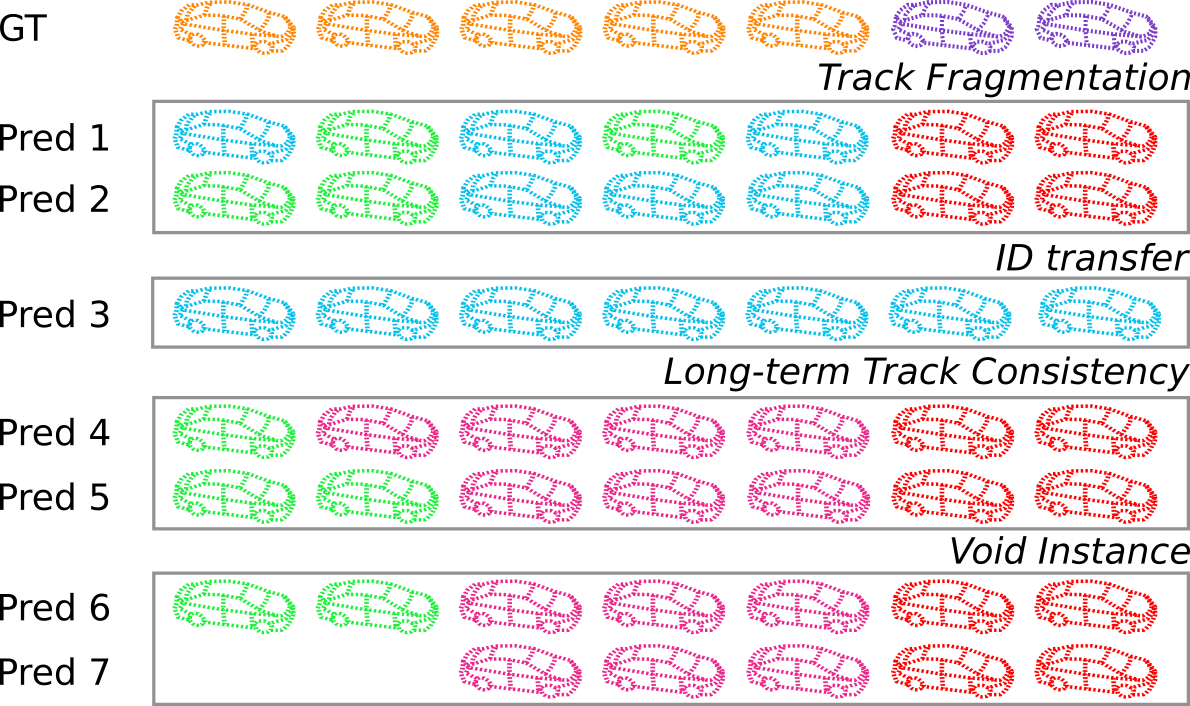}
\vspace{-0.3cm}
\end{table}

\subsubsection{Track Fragmentation}

In \tabref{tab:metric}, Pred 1 and 2 represent tracking predictions with the same number of incorrect track IDs but with permuted frames. We observe that \newmet~and PTQ are able to penalize the track fragmentation while LSTQ remains invariant to frame permutation. In the particular case of autonomous driving, we aim to provide a metric that encourages tracking consistency so that approaches correctly represent the dynamics of the scene.

\subsubsection{ID-Transfer}

Prediction 3 of case 2 shown in \tabref{tab:metric} refers to when a track finishes and a new track corresponding to a different object instance begins. Ideally, the metric should be able to penalize the predictions that do not reflect the track changes to account for association precision. We observe that both \newmet~and LSTQ correctly penalize this case, while PTQ incorrectly assesses it as the same track.

\subsubsection{Long-Term Tracking Consistency}

Predictions 4 and 5 of case 3 shown in \figref{tab:metric} exemplify a tracking error in which a single track is predicted as two different tracks. The difference between these predictions is the track duration. We aim to provide a metric that accounts for association recall stimulating long-term track consistency. The results show that the longer track prediction is evaluated with a higher score for our \newmet~and LSTQ while PTQ fails in this case.

\subsubsection{Void Instance Prediction}

As previously discussed in \cite{weber2021step}, a suitable metric should be unsusceptible to prediction processing, which affects the metric by yielding a better result without improving the quality of predictions. One such processing is ignoring the correctly segmented instances with the incorrect track IDs during evaluation. We represent this case 4 with predictions 6 and 7 in \tabref{tab:metric} having instances assigned with the incorrect track ID presented and ignored for evaluation respectively. The results show that all three metrics are able to penalize ignoring the incorrectly tracked instances.

\subsection{Ablation Study}
\label{subsec:ablation}

In this section, we present ablation studies that demonstrate the utility of our Panoptic nuScenes benchmark dataset.

\subsubsection{Influence of Pre-Training}
Transfer learning by pre-training on a large-scale dataset followed by fine-tuning on the target dataset generally improves the performance of complex image segmentation tasks. Due to the lack of LiDAR panoptic segmentation datasets, transfer learning for panoptic segmentation has not been investigated thus far. To demonstrate the utility of our diverse Panoptic nuScenes dataset, we performed transfer learning experiments with SemanticKITTI, using our EfficientLPS model with a Kalman Filter.
%
%
\tabref{tab:pretrainingNew} shows that pre-training on one dataset and fine-tuning on the other, results in increased performance on the respective datasets. Overall, we observe an increase of $1.1\%$ in PAT, $1.8\%$ in PQ, and $1.6\%$ in mIoU scores for the model pre-trained on SemanticKITTI and fine-tuned on Panoptic nuScenes. Whereas, the conversely trained model achieves a improvement of $0.6\%$ in PAT, $0.8\%$ in PQ, and $0.7\%$ in mIoU scores. The stronger performance gain when pre-training on SemanticKITTI is expected, as Panoptic nuScenes is sparser than SemanticKITTI. Thus, the rich feature representation captured during pre-training translates more efficiently from dense to sparse datasets. Nevertheless, the presence of two distinct datasets is highly beneficial for the research community as each can leverage the other and at the same time it facilitates studying the generalization ability of proposed methods.

\begin{table}
\centering
\caption{Comparison of panoptic tracking performance while pre-training on different datasets using the EfficientLPS model with Kalman Filter. Results are reported on the validation set of the respective fine-tuning datasets. The same pre-training and fine-tuning dataset indicates only training once on the dataset without any transfer learning. All scores are in [\%].}
\label{tab:pretrainingNew}
\footnotesize
\begin{tabular}{l|ccc|ccc}
\toprule
Pre-Training & \multicolumn{6}{c}{Fine-Tuning Dataset} \\
\cmidrule{2-7}
Dataset &  \multicolumn{3}{c|}{Panoptic nuScenes} & \multicolumn{3}{c}{SemanticKITTI} \\
\cmidrule{2-7}
& {\textbf{\newmet}} & PQ & mIoU & {\textbf{\newmet}} & PQ & mIoU  \\
\midrule
Panoptic nuScenes & 64.6 & 62.0  & 58.6 & \textbf{58.5} & \textbf{58.2}  & \textbf{62.1} \\
SemanticKITTI & \textbf{65.7} & \textbf{63.8} & \textbf{60.2} & 57.9 & 57.4  & 61.4 \\
\bottomrule
\end{tabular}
\vspace{-0.2cm}
\end{table}

\subsubsection{Impact of Diverse Scenes in Training Set}
\label{sec:generalize}
In this section, we study the generalization ability of an approach trained on a dataset consisting of diverse scenes and objects. We train two EfficeintLPS models for panoptic segmentation, one on the Panoptic nuScenes and the other on SemanticKITTI. We evaluate these models on the unseen validation set of PandaSet~\cite{pandaset}. Please note that PandaSet does not provide official panoptic segmentation annotations, we generate these annotations using a similar approach described in \secref{sec:annotation_protocol}. For this experiment, we only consider the classes that are common among all three datasets. 
We find that the model trained on Panoptic nuScenes outperforms that trained on SemanticKITTI on all the metrics. We observe the highest improvement of $1.3\%$ in the RQ score which can be attributed to the presence of diverse scenes in Panoptic nuScenes that consist of different types of \emph{thing} object classes which helps in better generalization. The improved instance segmentation capability consequently results in an overall higher panoptic quality.

\begin{table}
\centering
\caption{Comparison of the generalization ability of EfficientLPS trained on a panoptic segmentation dataset with diverse scenes and evaluated in unseen environments. All scores are in [\%].\looseness=-1}
\label{tab:generalization}
\footnotesize
\begin{tabular}{l|ccccc}
\toprule
Training Dataset & \multicolumn{5}{c}{Evaluation Dataset} \\
\cmidrule{2-6}
&  \multicolumn{5}{c}{PandaSet} \\
\cmidrule{2-6}
& PQ & PQ$^\dagger$ & SQ & RQ & mIoU  \\ 
\toprule
SemanticKITTI & 27.6 & 29.9 & 65.8 & 40.8 & 33.2\\
Panoptic nuScenes & \textbf{28.4} & \textbf{30.3} & \textbf{66.2} & \textbf{42.1} & \textbf{33.9}\\
\bottomrule
\end{tabular}
\vspace{-0.3cm}
\end{table}

\section{Conclusion}

In this paper, we introduce Panoptic nuScenes, a large-scale public LiDAR benchmark dataset that facilitates research on semantic segmentation, panoptic segmentation, and panoptic tracking tasks. We released an evaluation server on a hidden test set for fair comparisons of the aforementioned tasks. Panoptic nuScenes alleviates the lack of a diverse urban dataset with a large number of scenes that contain many moving objects and point-wise annotations for the aforementioned tasks. We performed an extensive analysis of existing panoptic tracking metrics and proposed the novel PAT metric that addresses the shortcomings of existing metrics. We presented exhaustive benchmarking results of several baselines for all three tasks in Panoptic nuScenes. The results demonstrate the need for further research on end-to-end learning methods that effectively address these tasks in a coherent manner. We believe that this work 
will pave the way for novel research on scene understanding of dynamic urban environments.

\section*{Acknowledgements}

This work was partly funded by Eva Mayr-Stihl Stiftung. The Panoptic nuScenes dataset was annotated by Scale.ai and we thank Jon Wilfong, Gerard Roy, and Galen Bertozzi. We thank Serene Chen at Motional for data inspection and quality control.\looseness-1

\typeout{}
\footnotesize
\bibliographystyle{IEEEtran}
\bibliography{references.bib}

\begin{thebibliography}{10}
\providecommand{\url}[1]{#1}
\csname url@rmstyle\endcsname
\providecommand{\newblock}{\relax}
\providecommand{\bibinfo}[2]{#2}
\providecommand\BIBentrySTDinterwordspacing{\spaceskip=0pt\relax}
\providecommand\BIBentryALTinterwordstretchfactor{4}
\providecommand\BIBentryALTinterwordspacing{\spaceskip=\fontdimen2\font plus
\BIBentryALTinterwordstretchfactor\fontdimen3\font minus
  \fontdimen4\font\relax}
\providecommand\BIBforeignlanguage[2]{{%
\expandafter\ifx\csname l@#1\endcsname\relax
\typeout{** WARNING: IEEEtran.bst: No hyphenation pattern has been}%
\typeout{** loaded for the language `#1'. Using the pattern for}%
\typeout{** the default language instead.}%
\else
\language=\csname l@#1\endcsname
\fi
#2}}

\bibitem{Geiger2013IJRR}
A.~Geiger, P.~Lenz, C.~Stiller, and R.~Urtasun, ``Vision meets robotics: The
  {KITTI} dataset,'' \emph{Int.~Journal of Robotics Research}, 2013.

\bibitem{caesar2020nuscenes}
H.~Caesar, V.~Bankiti, A.~H. Lang, S.~Vora, V.~E. Liong, Q.~Xu, A.~Krishnan,
  Y.~Pan, G.~Baldan, and O.~Beijbom, ``{nuScenes}: A multimodal dataset for
  autonomous driving,'' in \emph{Proc.~of the IEEE Conf.~on Computer Vision and
  Pattern Recognition}, 2020, pp. 11\,621--11\,631.

\bibitem{chang2019argoverse}
M.-F. Chang, J.~Lambert, P.~Sangkloy, J.~Singh, S.~Bak, \emph{et~al.},
  ``Argoverse: 3d tracking and forecasting with rich maps,'' in \emph{Proc.~of
  the IEEE Conf.~on Computer Vision and Pattern Recognition}, 2019, pp.
  8748--8757.

\bibitem{houston2020one}
J.~Houston, G.~Zuidhof, L.~Bergamini, Y.~Ye, L.~Chen, A.~Jain, S.~Omari,
  V.~Iglovikov, and P.~Ondruska, ``One thousand and one hours: Self-driving
  motion prediction dataset,'' \emph{arXiv preprint arXiv:2006.14480}, 2020.

\bibitem{sun2020scalability}
P.~Sun, H.~Kretzschmar, X.~Dotiwalla, A.~Chouard, \emph{et~al.}, ``Scalability
  in perception for autonomous driving: Waymo open dataset,'' in \emph{Proc.~of
  the IEEE Conf.~on Computer Vision and Pattern Recognition}, 2020.

\bibitem{thomas2019kpconv}
H.~Thomas, C.~R. Qi, J.-E. Deschaud, B.~Marcotegui, F.~Goulette, and L.~J.
  Guibas, ``Kpconv: Flexible and deformable convolution for point clouds,'' in
  \emph{Int.~Conf.~on Computer Vision}, 2019, pp. 6411--6420.

\bibitem{wu2018squeezeseg}
B.~Wu, A.~Wan, X.~Yue, and K.~Keutzer, ``Squeezeseg: Convolutional neuralnets
  with recurrent {CRF} for real-time road-object segmentation from 3d lidar
  point cloud,'' in \emph{Int.~Conf.~on Robotics and Automation}, 2018.

\bibitem{milioto2019rangenet++}
A.~Milioto, I.~Vizzo, J.~Behley, and C.~Stachniss, ``Rangenet++: Fast and
  accurate lidar semantic segmentation,'' in \emph{Int.~Conf.~on Intelligent
  Robots and Systems}, 2019, pp. 4213--4220.

\bibitem{behley2019semantickitti}
J.~Behley, M.~Garbade, A.~Milioto, J.~Quenzel, S.~Behnke, C.~Stachniss, and
  J.~Gall, ``{SemanticKITTI}: A dataset for semantic scene understanding of
  lidar sequences,'' in \emph{Int.~Conf.~on Computer Vision}, 2019.

\bibitem{sirohi2021efficientlps}
K.~Sirohi, R.~Mohan, D.~B{\"u}scher, W.~Burgard, and A.~Valada,
  ``{EfficientLPS}: Efficient lidar panoptic segmentation,'' \emph{arXiv
  preprint arXiv:2102.08009}, 2021.

\bibitem{hurtado2020mopt}
J.~V. Hurtado, R.~Mohan, W.~Burgard, and A.~Valada, ``{MOPT}: Multi-object
  panoptic tracking,'' \emph{The IEEE/CVF Conference on Computer Vision and
  Pattern Recognition Workshops}, 2020.

\bibitem{tan2020toronto3d}
W.~Tan, N.~Qin, L.~Ma, Y.~Li, J.~Du, G.~Cai, K.~Yang, and J.~Li, ``Toronto-3d:
  {A} large-scale mobile lidar dataset for semantic segmentation of urban
  roadways,'' \emph{arXiv preprint arXiv:2003.08284}, 2020.

\bibitem{geyer2020a2d2}
J.~Geyer, Y.~Kassahun, M.~Mahmudi, X.~Ricou, R.~Durgesh, A.~S. Chung,
  \emph{et~al.}, ``{A2D2: Audi Autonomous Driving Dataset},'' \emph{arXiv
  preprint arXiv:2004.06320}, 2020.

\bibitem{pandaset}
\BIBentryALTinterwordspacing
``{PandaSet}.'' [Online]. Available: \url{https://pandaset.org}
\BIBentrySTDinterwordspacing

\bibitem{behley2021panoptickitti}
J.~Behley, A.~Milioto, and C.~Stachniss, ``{A Benchmark for LiDAR-based
  Panoptic Segmentation based on KITTI},'' in \emph{Int.~Conf.~on Robotics and
  Automation}, 2021.

\bibitem{aygun20214d}
M.~Aygun, A.~Osep, M.~Weber, M.~Maximov, C.~Stachniss, J.~Behley, and
  L.~Leal-Taix{\'e}, ``4d panoptic lidar segmentation,'' in \emph{Proc.~of the
  IEEE Conf.~on Computer Vision and Pattern Recognition}, 2021.

\bibitem{hong2021dynamic}
F.~Hong, H.~Zhou, X.~Zhu, H.~Li, and Z.~Liu, ``Lidar-based panoptic
  segmentation via dynamic shifting network,'' in \emph{Proc.~of the IEEE
  Conf.~on Computer Vision and Pattern Recognition}, 2021, pp.
  13\,090--13\,099.

\bibitem{zhou2021panoptic}
Z.~Zhou, Y.~Zhang, and H.~Foroosh, ``Panoptic-polarnet: Proposal-free lidar
  point cloud panoptic segmentation,'' in \emph{Proc.~of the IEEE Conf.~on
  Computer Vision and Pattern Recognition}, 2021, pp. 13\,194--13\,203.

\bibitem{hu2020randla}
Q.~Hu, B.~Yang, L.~Xie, S.~Rosa, Y.~Guo, Z.~Wang, N.~Trigoni, and A.~Markham,
  ``Randla-net: Efficient semantic segmentation of large-scale point clouds,''
  in \emph{Proc.~of the IEEE Conf.~on Computer Vision and Pattern Recognition},
  2020, pp. 11\,108--11\,117.

\bibitem{zhang2020polarnet}
Y.~Zhang, Z.~Zhou, P.~David, X.~Yue, Z.~Xi, B.~Gong, and H.~Foroosh,
  ``Polarnet: An improved grid representation for online lidar point clouds
  semantic segmentation,'' in \emph{Proc.~of the IEEE Conf.~on Computer Vision
  and Pattern Recognition}, 2020, pp. 9601--9610.

\bibitem{zhu2021cylindrical}
X.~Zhu, H.~Zhou, T.~Wang, F.~Hong, Y.~Ma, W.~Li, H.~Li, and D.~Lin,
  ``Cylindrical and asymmetrical 3d convolution networks for lidar
  segmentation,'' in \emph{Proc.~of the IEEE Conf.~on Computer Vision and
  Pattern Recognition}, 2021, pp. 9939--9948.

\bibitem{milioto2020lidar}
A.~Milioto, J.~Behley, C.~McCool, and C.~Stachniss, ``Lidar panoptic
  segmentation for autonomous driving,'' in \emph{Int.~Conf.~on Intelligent
  Robots and Systems}, 2020, pp. 8505--8512.

\bibitem{gasperini2021panoster}
S.~Gasperini, M.-A.~N. Mahani, A.~Marcos-Ramiro, N.~Navab, and F.~Tombari,
  ``Panoster: End-to-end panoptic segmentation of lidar point clouds,''
  \emph{IEEE Robotics and Automation Letters}, vol.~6, no.~2, 2021.

\bibitem{mohan2020efficientps}
R.~Mohan and A.~Valada, ``Efficientps: Efficient panoptic segmentation,''
  \emph{Int.~Journal of Computer Vision}, vol. 129, pp. 1551 -- 1579, 2021.

\bibitem{cattaneo2021lcdnet}
D.~Cattaneo, M.~Vaghi, and A.~Valada, ``Lcdnet: Deep loop closure detection and
  point cloud registration for lidar slam,'' \emph{arXiv preprint
  arXiv:2103.05056}, 2021.

\bibitem{kirillov2019panoptic}
A.~Kirillov, K.~He, R.~Girshick, C.~Rother, and P.~Doll{\'a}r, ``Panoptic
  segmentation,'' in \emph{Proc.~of the IEEE Conf.~on Computer Vision and
  Pattern Recognition}, 2019, pp. 9404--9413.

\bibitem{porzi2019seamless}
L.~Porzi, S.~R. Bulo, A.~Colovic, and P.~Kontschieder, ``Seamless scene
  segmentation,'' in \emph{Proc.~of the IEEE Conf.~on Computer Vision and
  Pattern Recognition}, 2019, pp. 8277--8286.

\bibitem{weber2021step}
M.~Weber, J.~Xie, M.~Collins, Y.~Zhu, P.~Voigtlaender, H.~Adam, B.~Green,
  A.~Geiger, B.~Leibe, D.~Cremers, \emph{et~al.}, ``Step: Segmenting and
  tracking every pixel,'' \emph{arXiv preprint arXiv:2102.11859}, 2021.

\bibitem{liong2020amvnet}
V.~E. Liong, T.~N.~T. Nguyen, S.~Widjaja, D.~Sharma, and Z.~J. Chong,
  ``{AMVNet}: Assertion-based multi-view fusion network for lidar semantic
  segmentation,'' \emph{arXiv preprint arXiv:2012.04934}, 2020.

\bibitem{cheng20212}
R.~Cheng, R.~Razani, E.~Taghavi, E.~Li, and B.~Liu, ``{AF2-S3Net}: Attentive
  feature fusion with adaptive feature selection for sparse semantic
  segmentation network,'' in \emph{Proc.~of the IEEE Conf.~on Computer Vision
  and Pattern Recognition}, 2021, pp. 12\,547--12\,556.

\bibitem{tang2020searching}
H.~Tang, Z.~Liu, S.~Zhao, Y.~Lin, J.~Lin, H.~Wang, and S.~Han, ``Searching
  efficient 3d architectures with sparse point-voxel convolution,'' in
  \emph{Europ.~Conf.~on Computer Vision}, 2020, pp. 685--702.

\bibitem{yin2021center}
T.~Yin, X.~Zhou, and P.~Krahenbuhl, ``Center-based 3d object detection and
  tracking,'' in \emph{Proc.~of the IEEE Conf.~on Computer Vision and Pattern
  Recognition}, 2021, pp. 11\,784--11\,793.

\bibitem{zaech2021ogr3mot}
J.~Zaech, D.~Dai, A.~Liniger, M.~Danelljan, and L.~V. Gool, ``Learnable online
  graph representations for 3d multi-object tracking,'' \emph{arXiv preprint
  arXiv:2104.11747}, 2021.

\end{thebibliography}

\clearpage
\renewcommand{\baselinestretch}{1}
\setlength{\belowcaptionskip}{0pt}

\begin{strip}
\begin{center}
\vspace{-5ex}
\textbf{\LARGE \bf
Panoptic nuScenes: A Large-Scale Benchmark for\\\vspace{0.5ex}LiDAR Panoptic Segmentation and Tracking} \\
\vspace{2ex}

\Large{\bf- Supplementary Material -}\\
\vspace{0.4cm}
\normalsize{Whye Kit Fong*$^{,1}$, Rohit Mohan*$^{,2}$, Juana Valeria Hurtado$^2$, Lubing Zhou$^1$,\\ Holger Caesar$^1$, Oscar Beijbom$^1$, and Abhinav Valada$^{2}$}
\end{center}
\end{strip}

\setcounter{section}{0}
\setcounter{equation}{0}
\setcounter{figure}{0}
\setcounter{table}{0}
\setcounter{page}{1}
\makeatletter

\renewcommand{\thesection}{S.\arabic{section}}
\renewcommand{\thesubsection}{S.\arabic{subsection}}
\renewcommand{\thetable}{S.\arabic{table}}
\renewcommand{\thefigure}{S.\arabic{figure}}


\let\thefootnote\relax\footnote{$^1$ Motional.\\
$^2$ Department of Computer Science, University of Freiburg, Germany.\\
* Authors contributed equally.}%

\normalsize

In this supplementary material, we (i) provide more details about the dataset and the challenges that we are organizing based on the Panoptic nuScenes benchmark that we introduce in this work, (ii) present class-wise panoptic segmentation results, and (iii) extend the correlation analysis between semantic segmentation, object detection, multi-object tracking, and panoptic tracking. Following, (iv) we compare our proposed panoptic tracking evaluation metric to existing ones. We then (v) provide more details for our ablation study on the impact of diverse scenes. Lastly, we present qualitative evaluations for the task of panoptic segmentation and tracking.

\section{Additional Dataset Details}
This section presents further details about the Panoptic nuScenes dataset and annotations.

\begin{figure}
\centering
\includegraphics[width=1.0\linewidth]{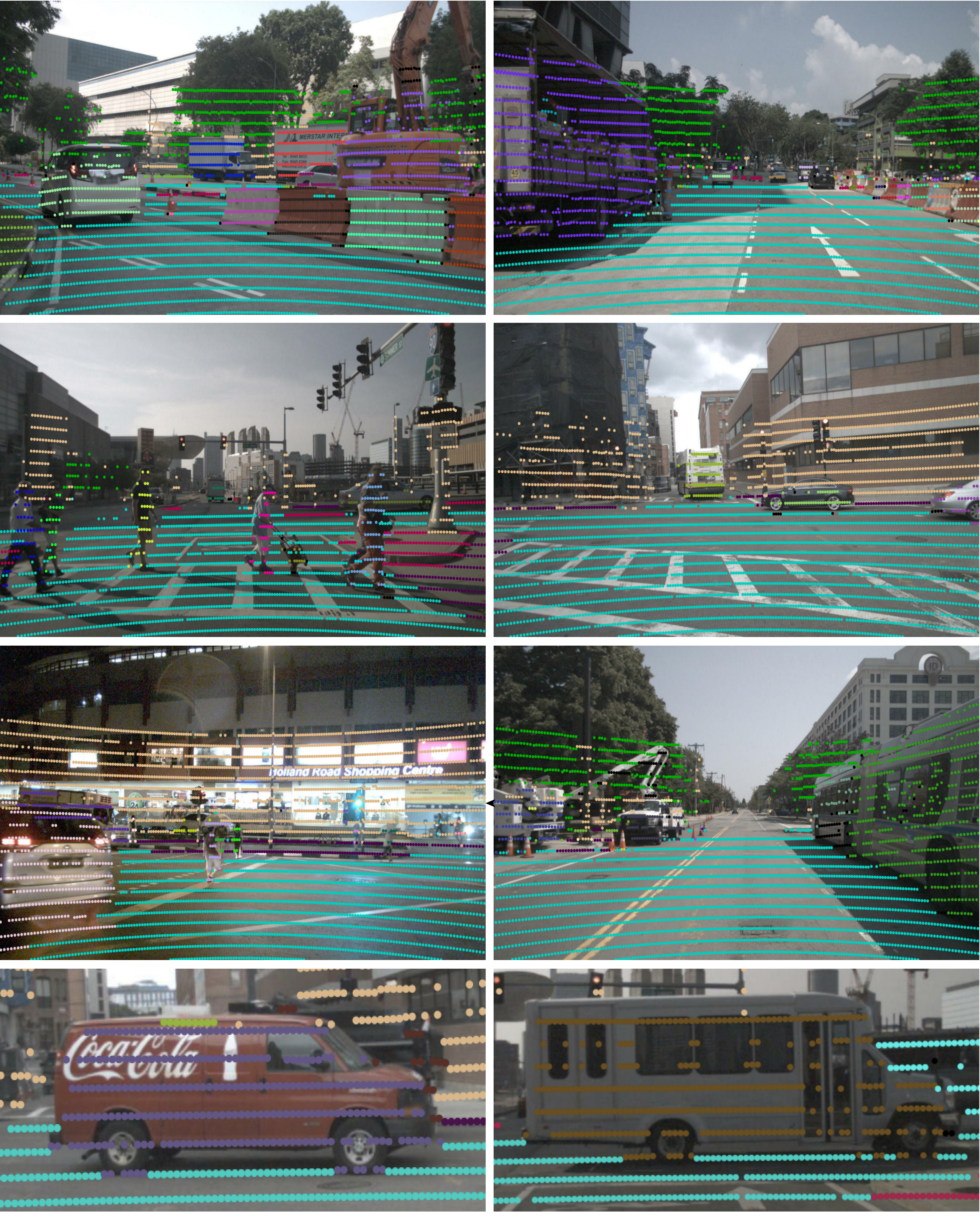}
\caption{Front camera view of panoptic annotation examples, including construction zones (row 1), junctions (row 2), nighttime (row 3 left) and bendy bus (row 3 right). 
We can see that the annotations accurately outline vehicle wheels, rather than including nearby ground points (row 4).}
\label{fig:annnotation_example}
\end{figure}

\noindent\textit{Dynamic and Diverse Scenes}: Panoptic nuScenes was primarily collected in dense urban environments with many dynamic agents. Such scenes include those near intersections and construction sites, which are of high traffic density and have the potential for interesting driving situations (e.g. jaywalkers, lane changes, turning). The scenes are also diverse in terms of, among others, geographical location (i.e. left-hand versus right-hand drive), weather, and lighting conditions. \figref{fig:annnotation_example} shows example panoptic segmentation annotations overlaid on the camera image.

\begin{figure*}
\centering
\footnotesize
{\renewcommand{\arraystretch}{1}
\begin{tabular}{P{0.4cm}P{5.5cm}P{5.5cm}P{5.5cm}}
\multicolumn{4}{c}{Semantic Segmentation} \\
&\raisebox{-0.4\height}{\includegraphics[width=\linewidth,frame]{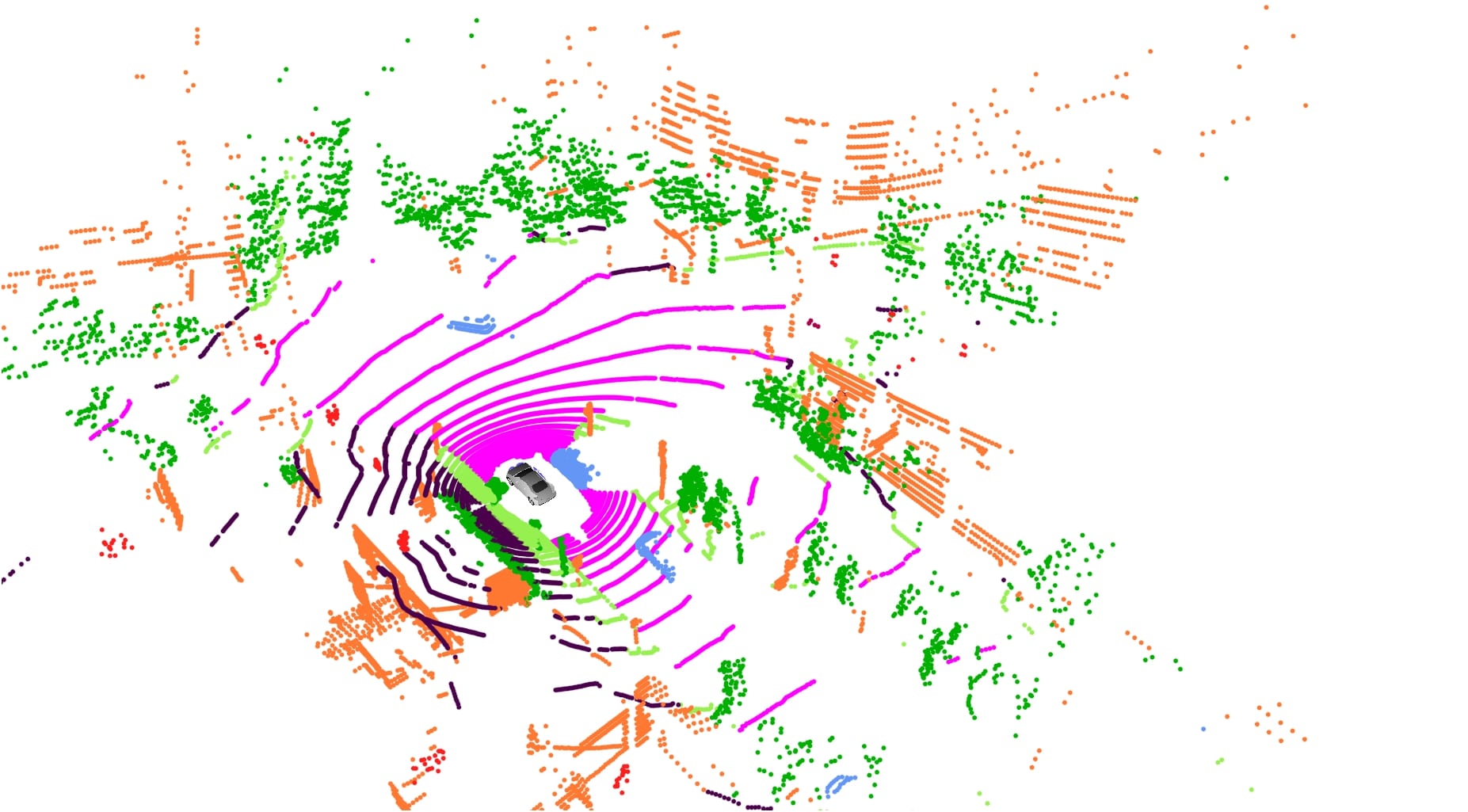}} & \raisebox{-0.4\height}{\includegraphics[width=\linewidth,frame]{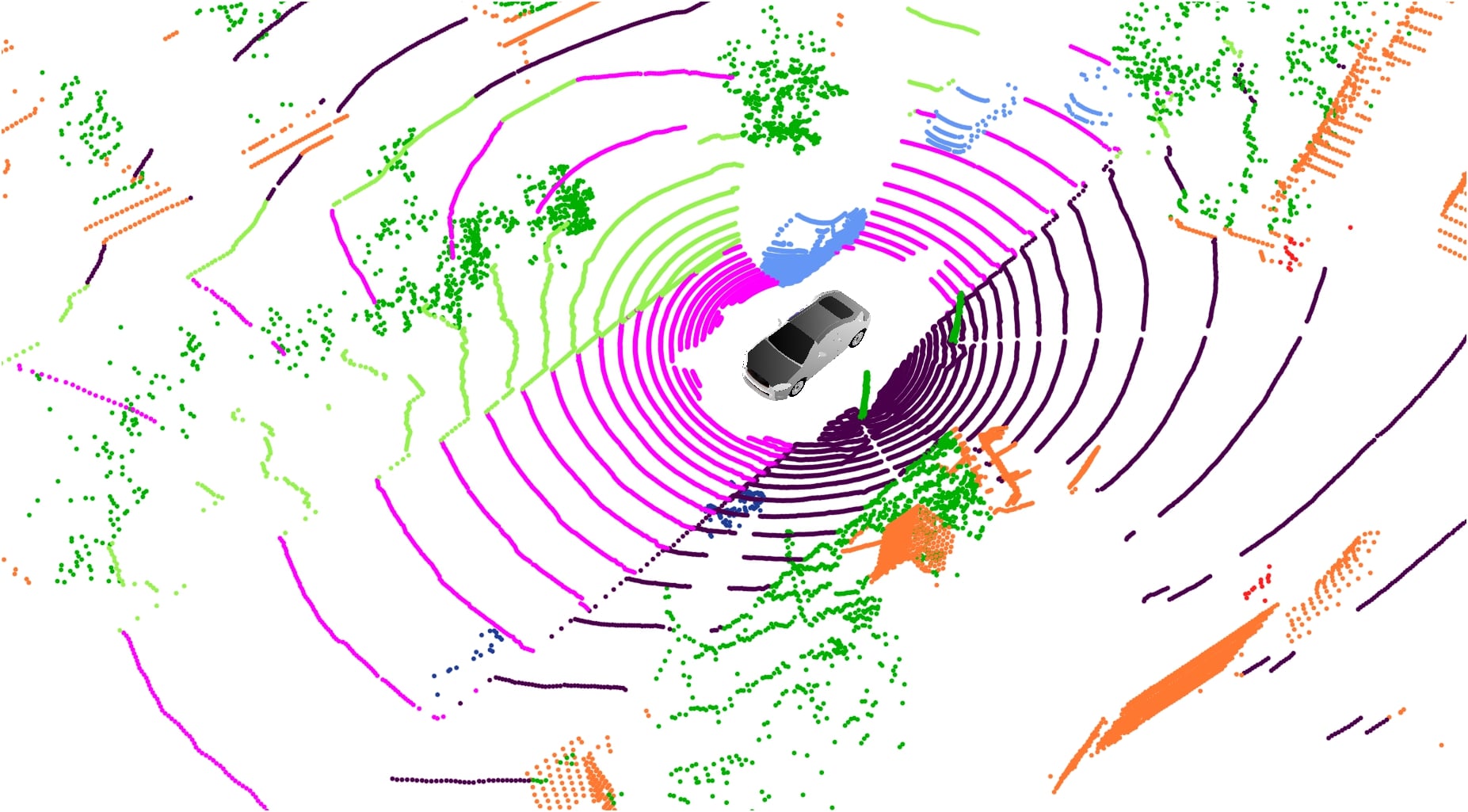}} & \raisebox{-0.4\height}{\includegraphics[width=\linewidth,frame]{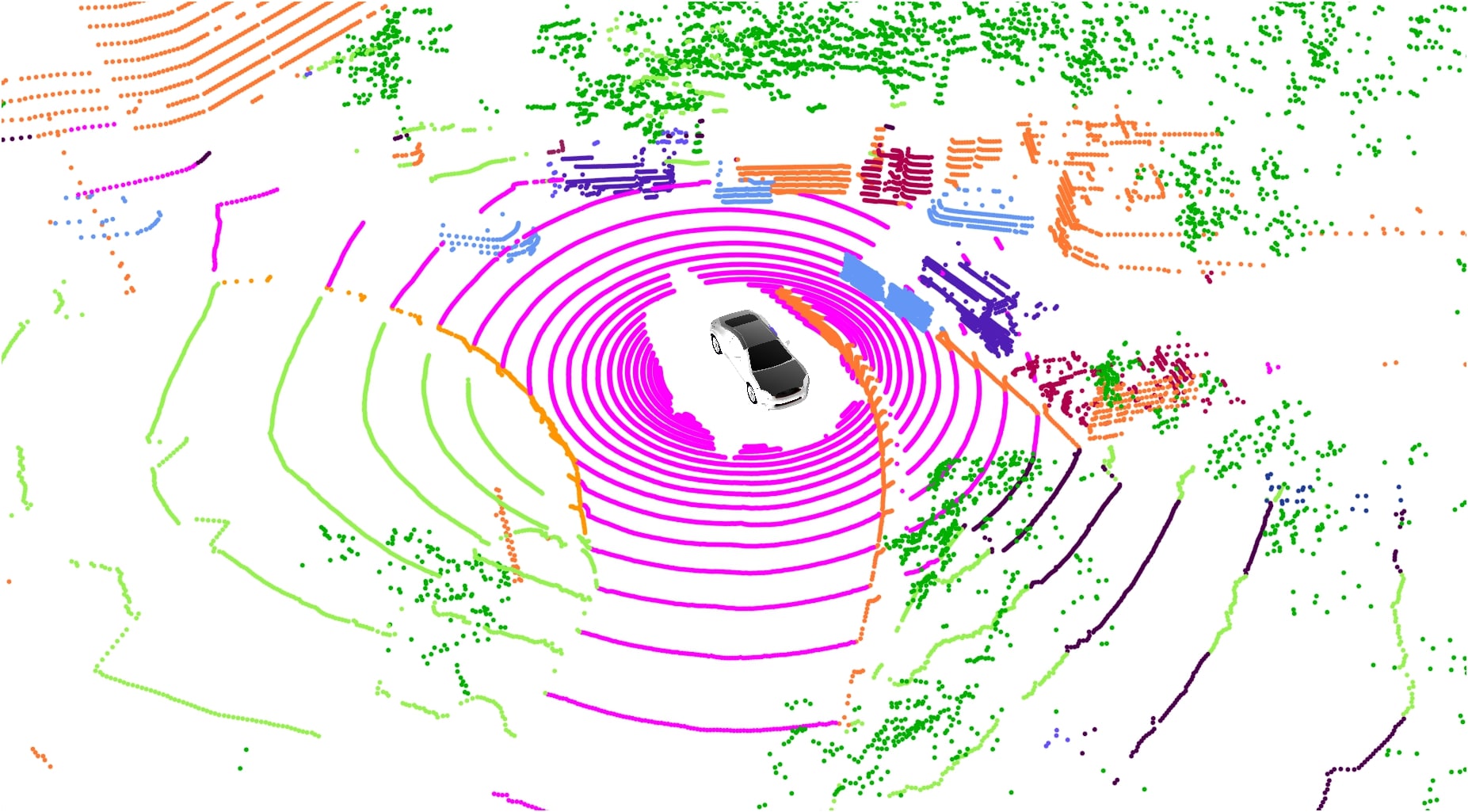}} \\
&\raisebox{-0.4\height}{(a)} & \raisebox{-0.4\height}{(b)} &  \raisebox{-0.4\height}{(c)} \\
\\
&\raisebox{-0.4\height}{\includegraphics[width=\linewidth,frame]{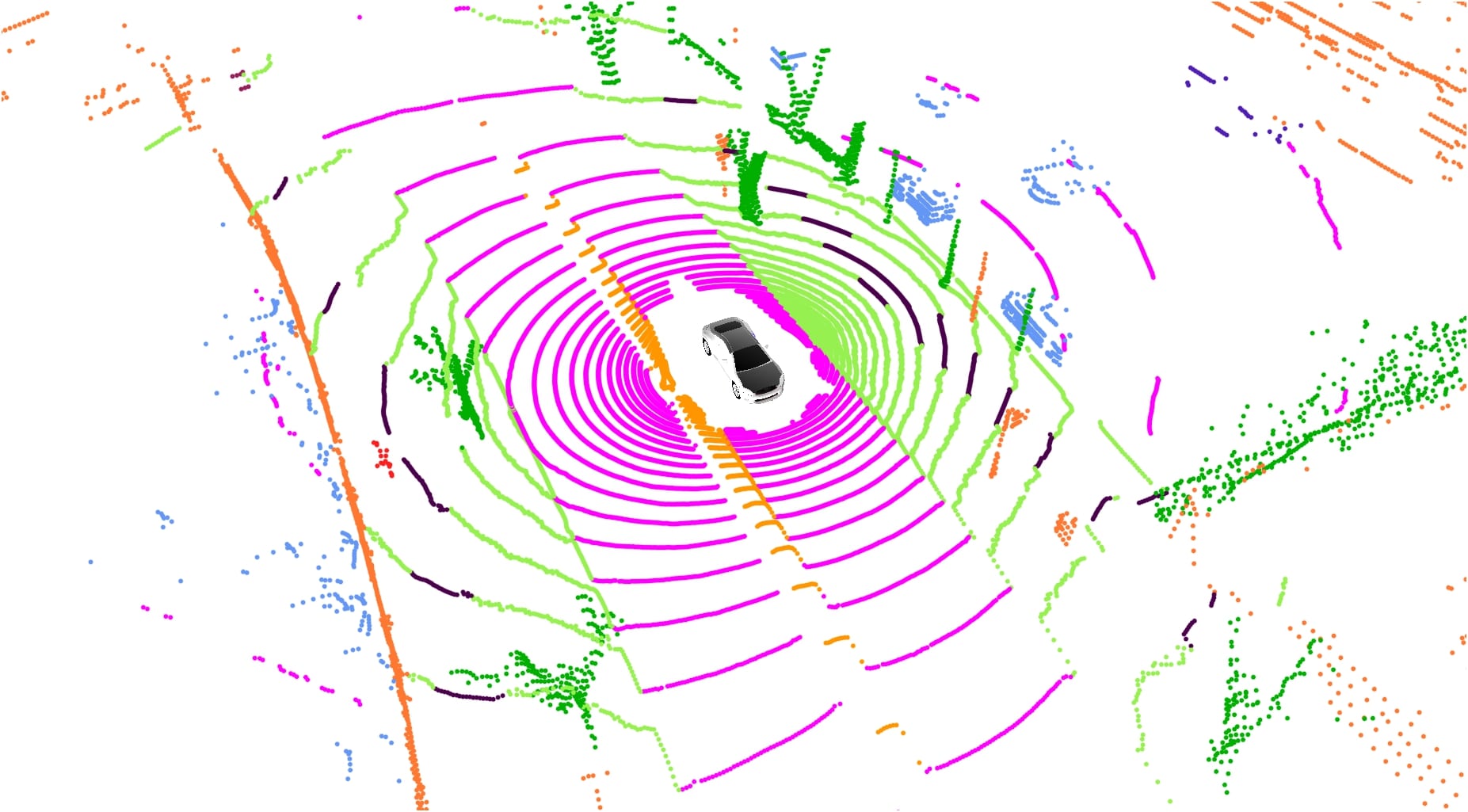}} & \raisebox{-0.4\height}{\includegraphics[width=\linewidth,frame]{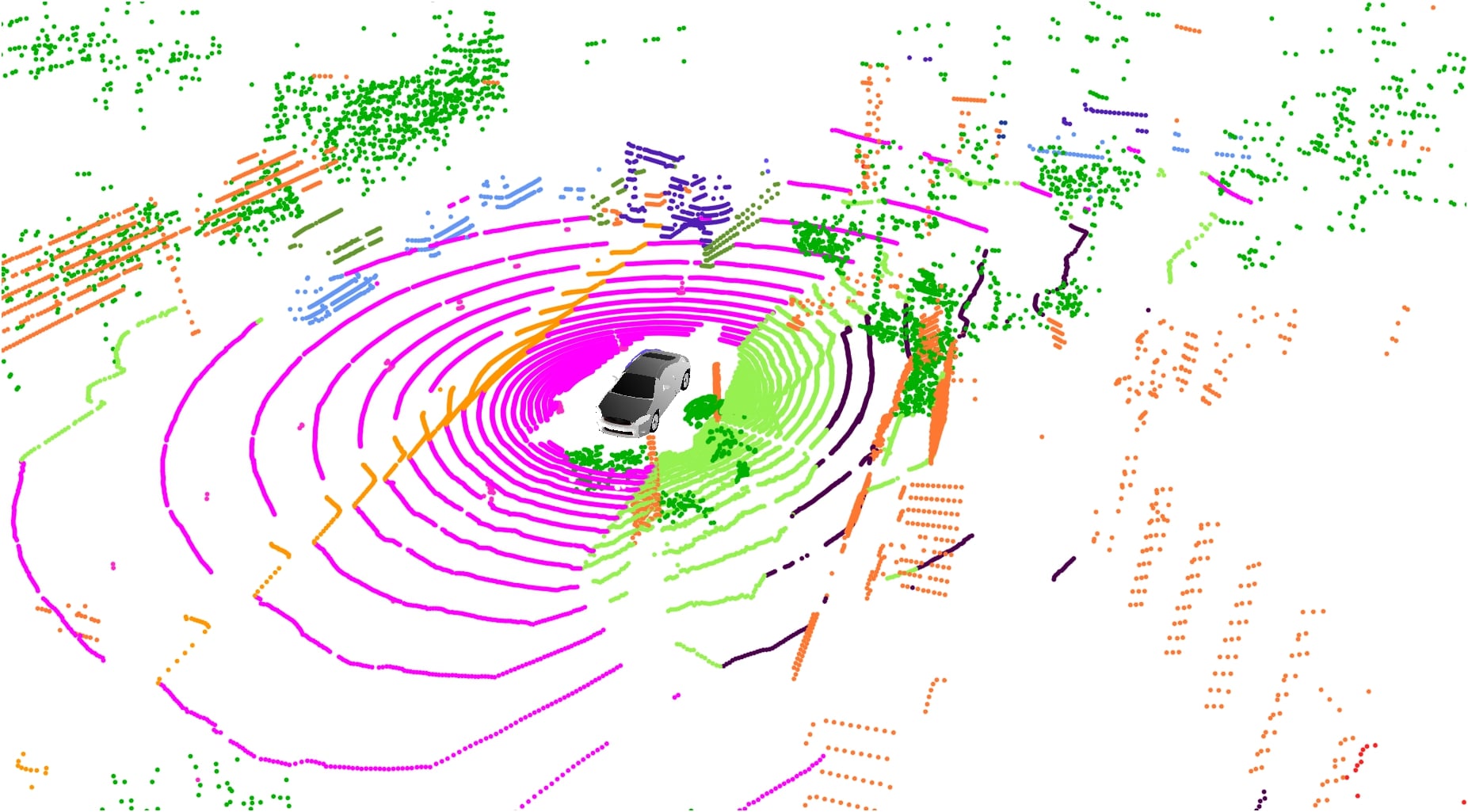}} & \raisebox{-0.4\height}{\includegraphics[width=\linewidth,frame]{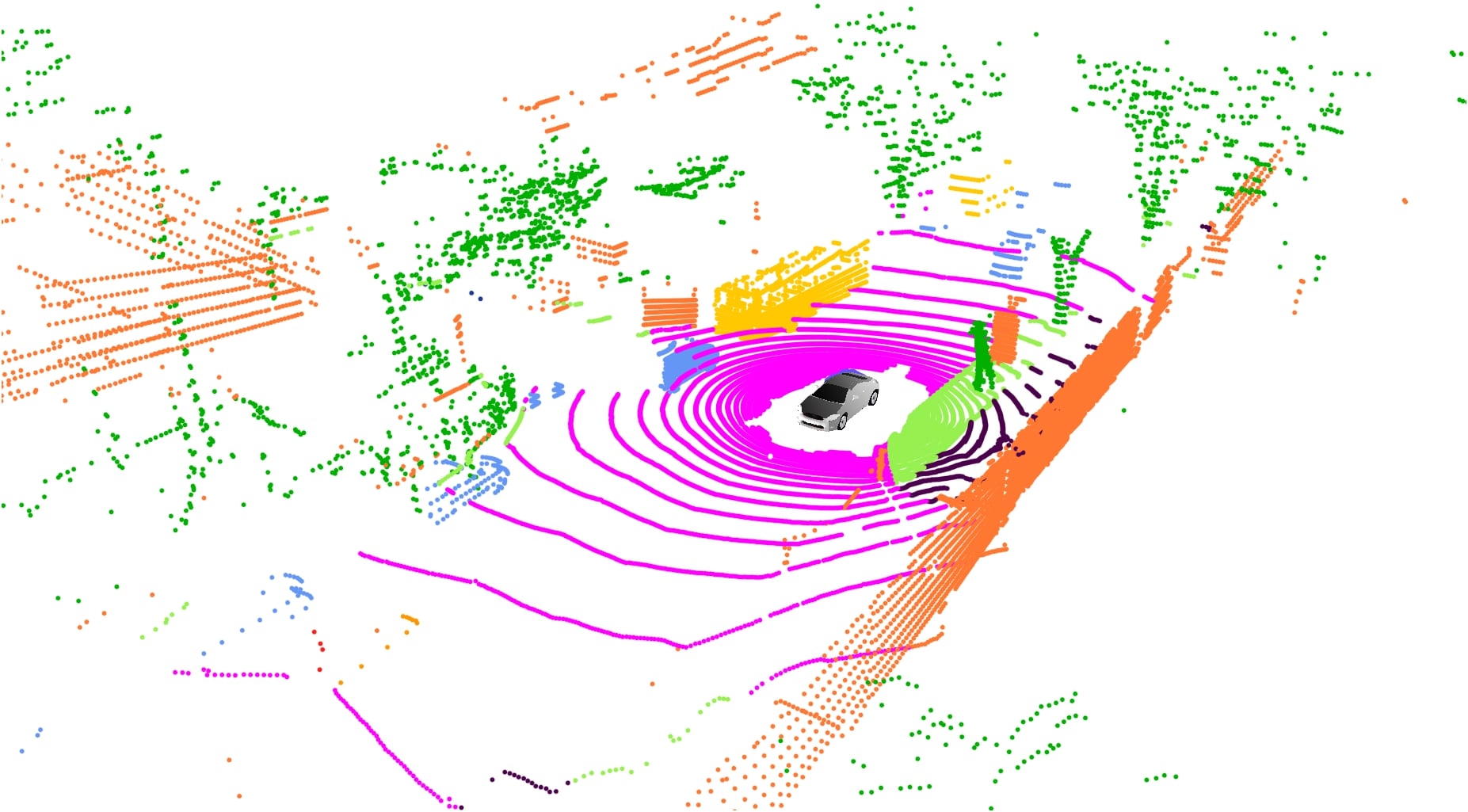}} \\
&\raisebox{-0.4\height}{(d)} & \raisebox{-0.4\height}{(e)} &  \raisebox{-0.4\height}{(f)} \\
\\\midrule
\multicolumn{4}{c}{Panoptic Segmentation} \\
&\raisebox{-0.4\height}{\includegraphics[width=\linewidth,frame]{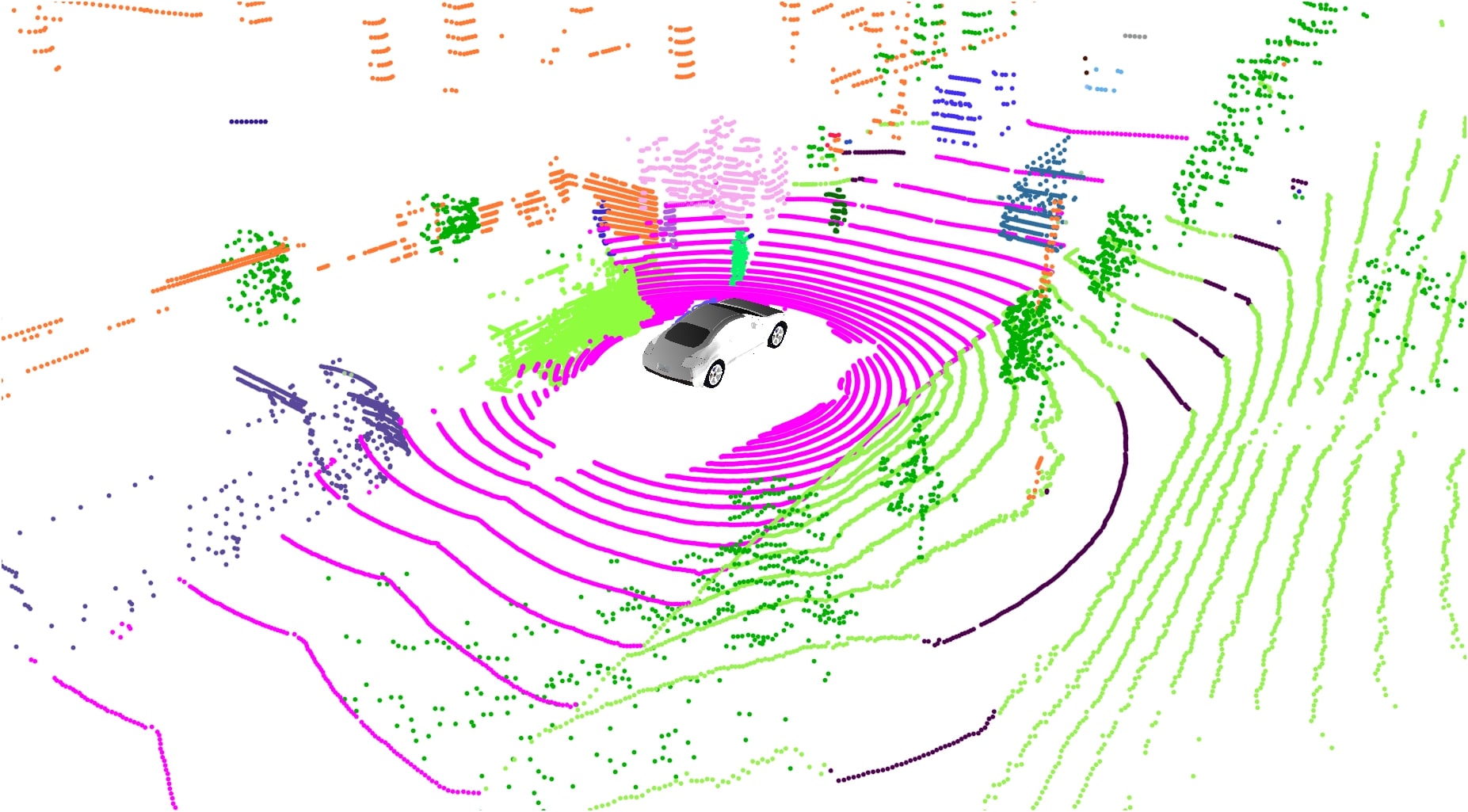}} & \raisebox{-0.4\height}{\includegraphics[width=\linewidth,frame]{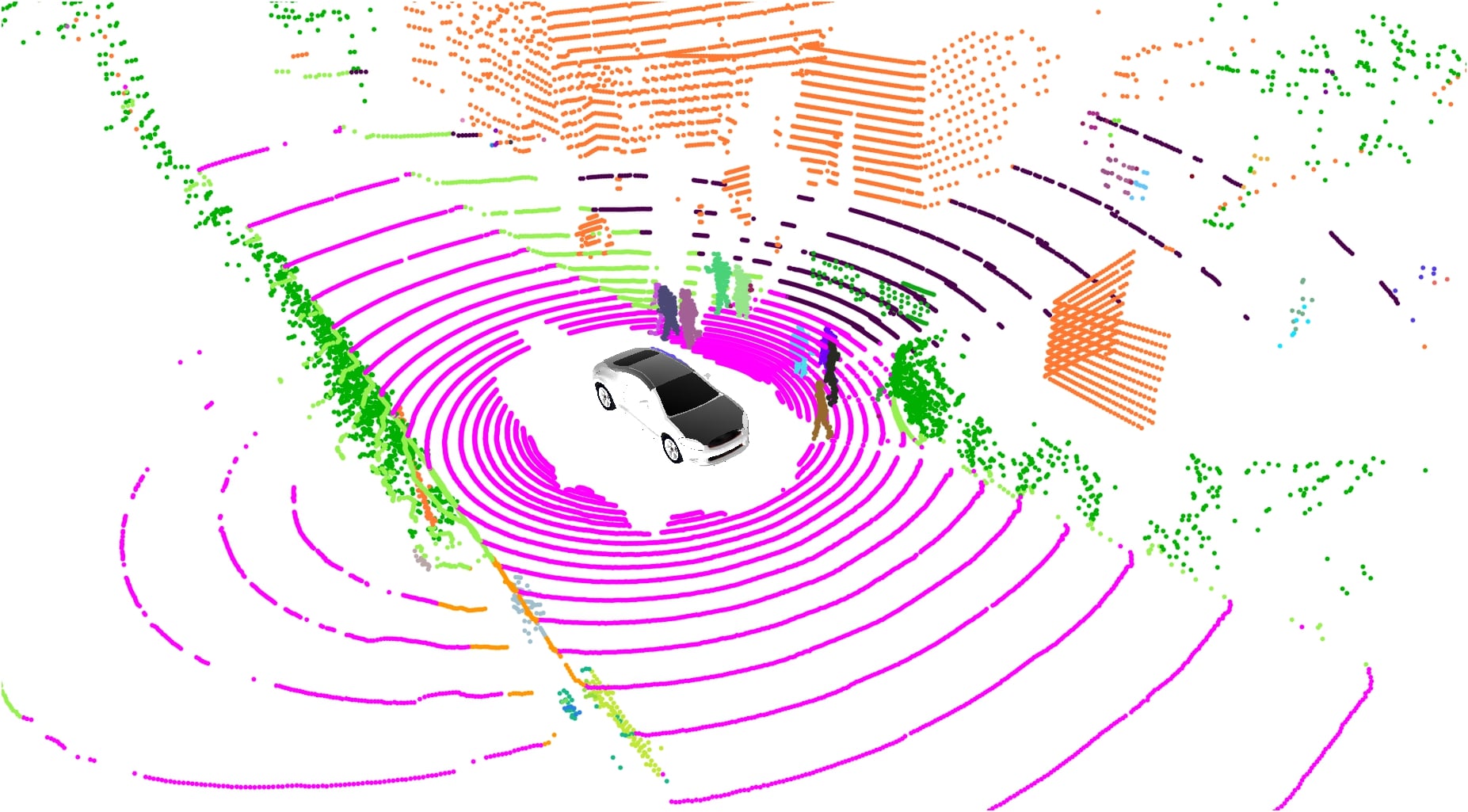}} & \raisebox{-0.4\height}{\includegraphics[width=\linewidth,frame]{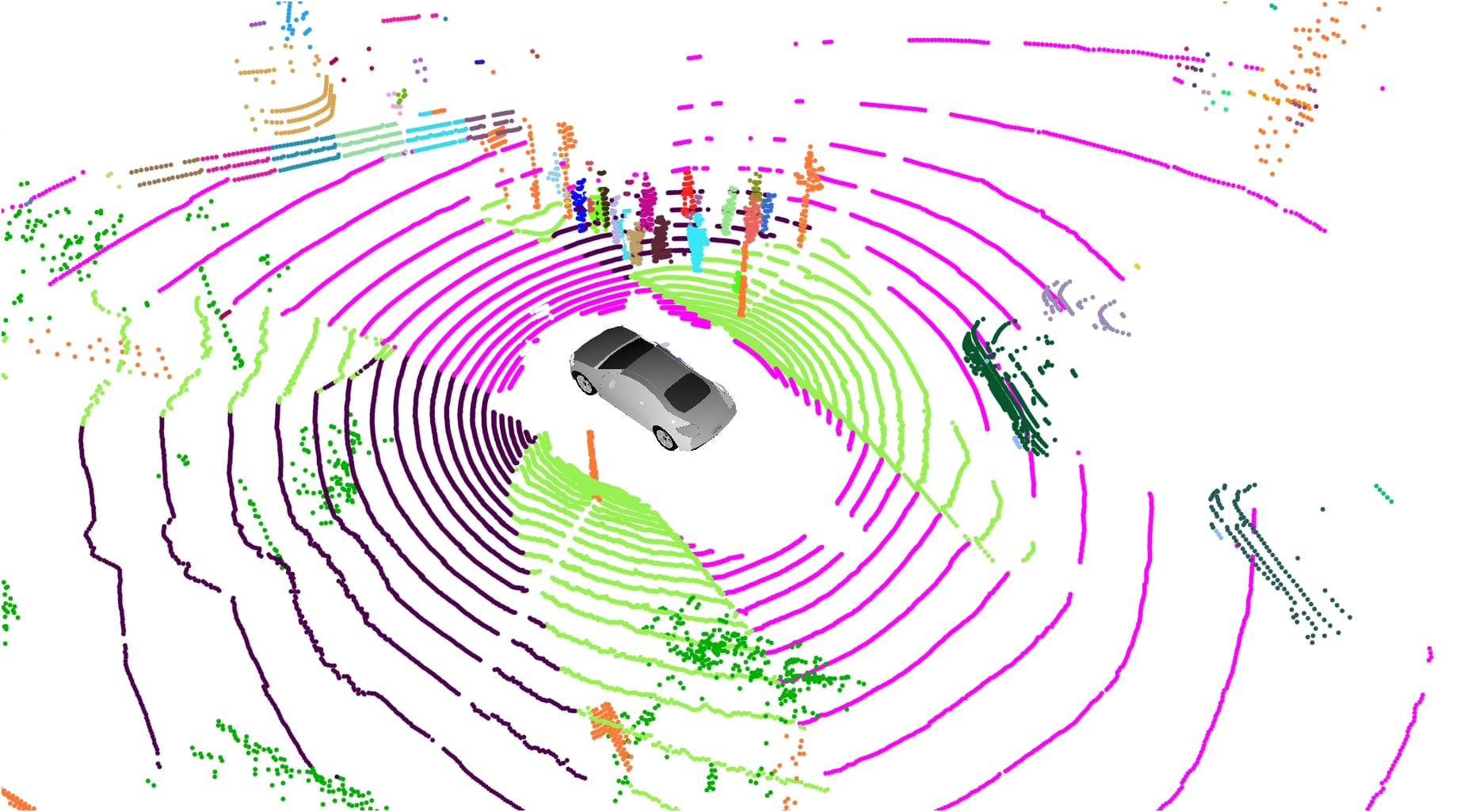}} \\
&\raisebox{-0.4\height}{(g)} & \raisebox{-0.4\height}{(h)} &  \raisebox{-0.4\height}{(i)} \\
\\
&\raisebox{-0.4\height}{\includegraphics[width=\linewidth,frame]{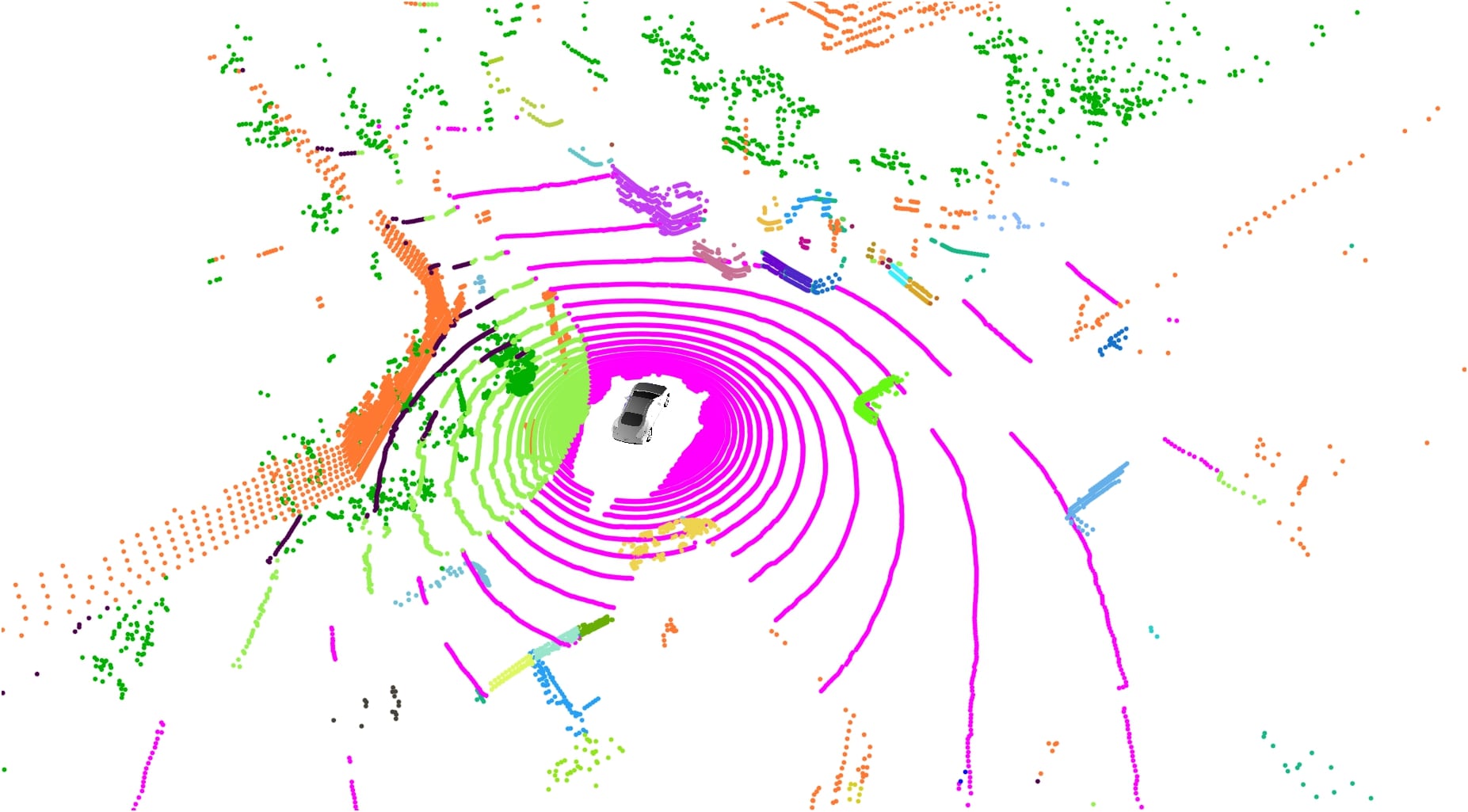}} & \raisebox{-0.4\height}{\includegraphics[width=\linewidth,frame]{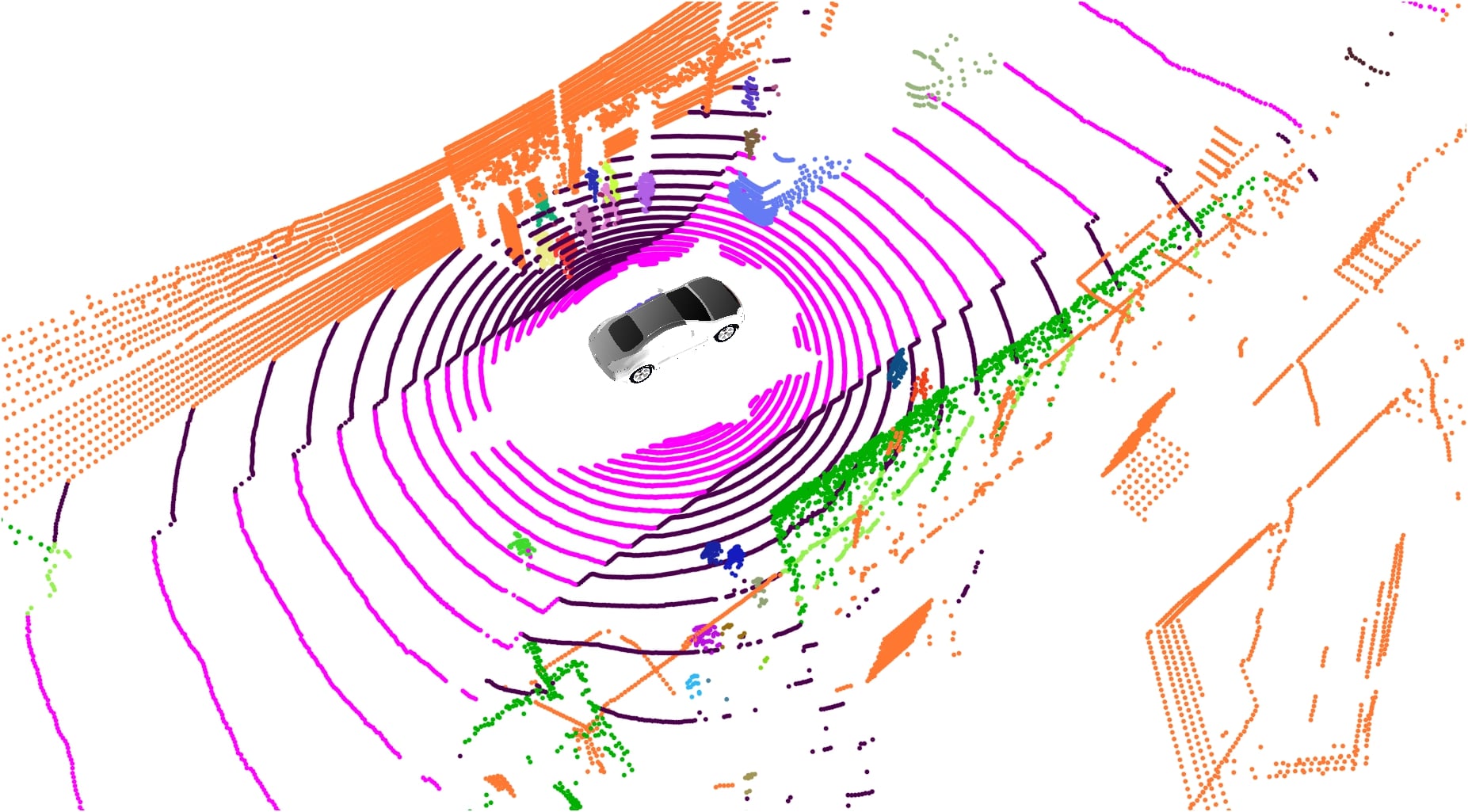}} & \raisebox{-0.4\height}{\includegraphics[width=\linewidth,frame]{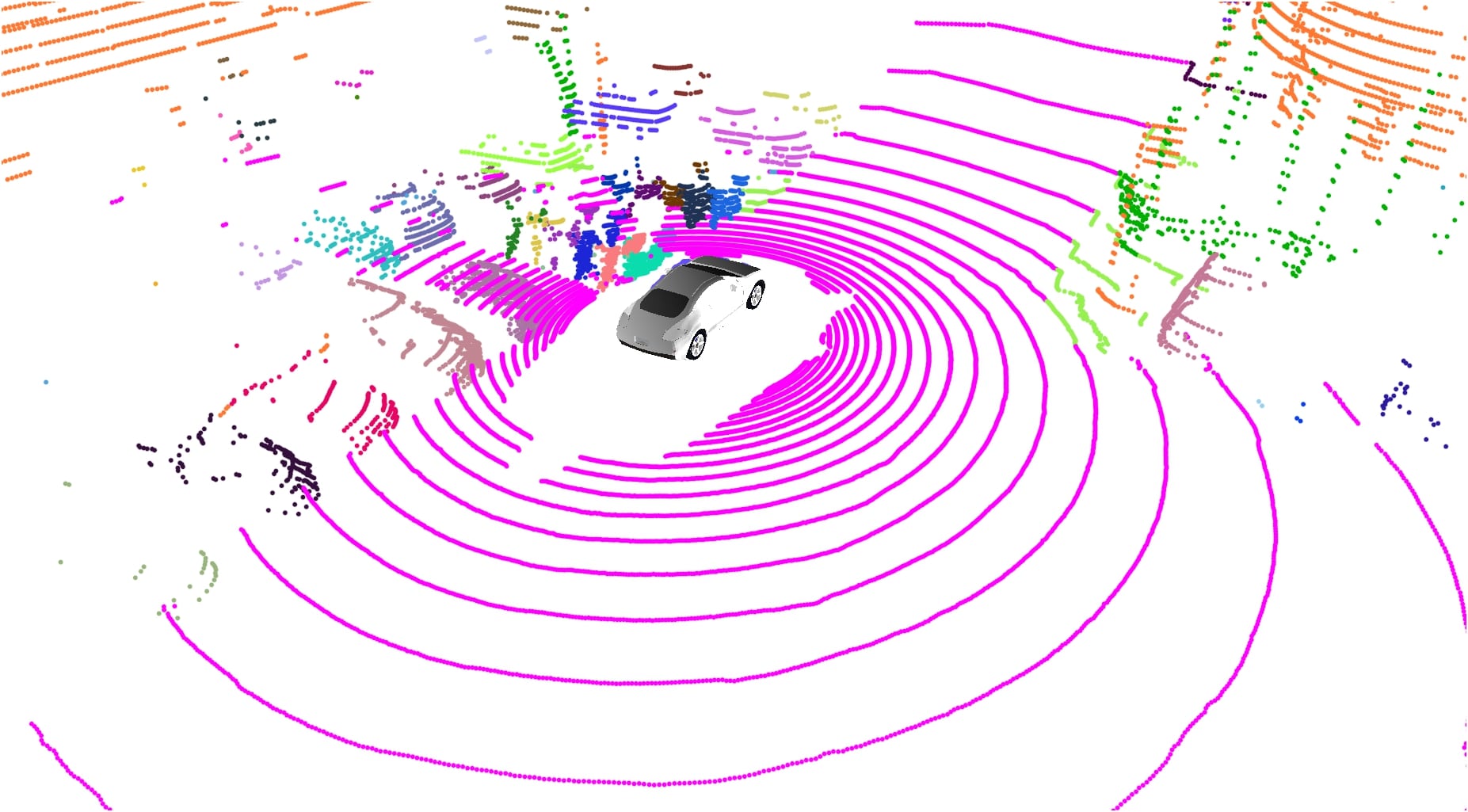}} \\
&\raisebox{-0.4\height}{(j)} & \raisebox{-0.4\height}{(k)} &  \raisebox{-0.4\height}{(l)} \\
\\\midrule
\multicolumn{4}{c}{Panoptic Tracking} \\
{\rotatebox[origin=c]{90}{(m)}}
&\raisebox{-0.4\height}{\includegraphics[width=\linewidth,frame]{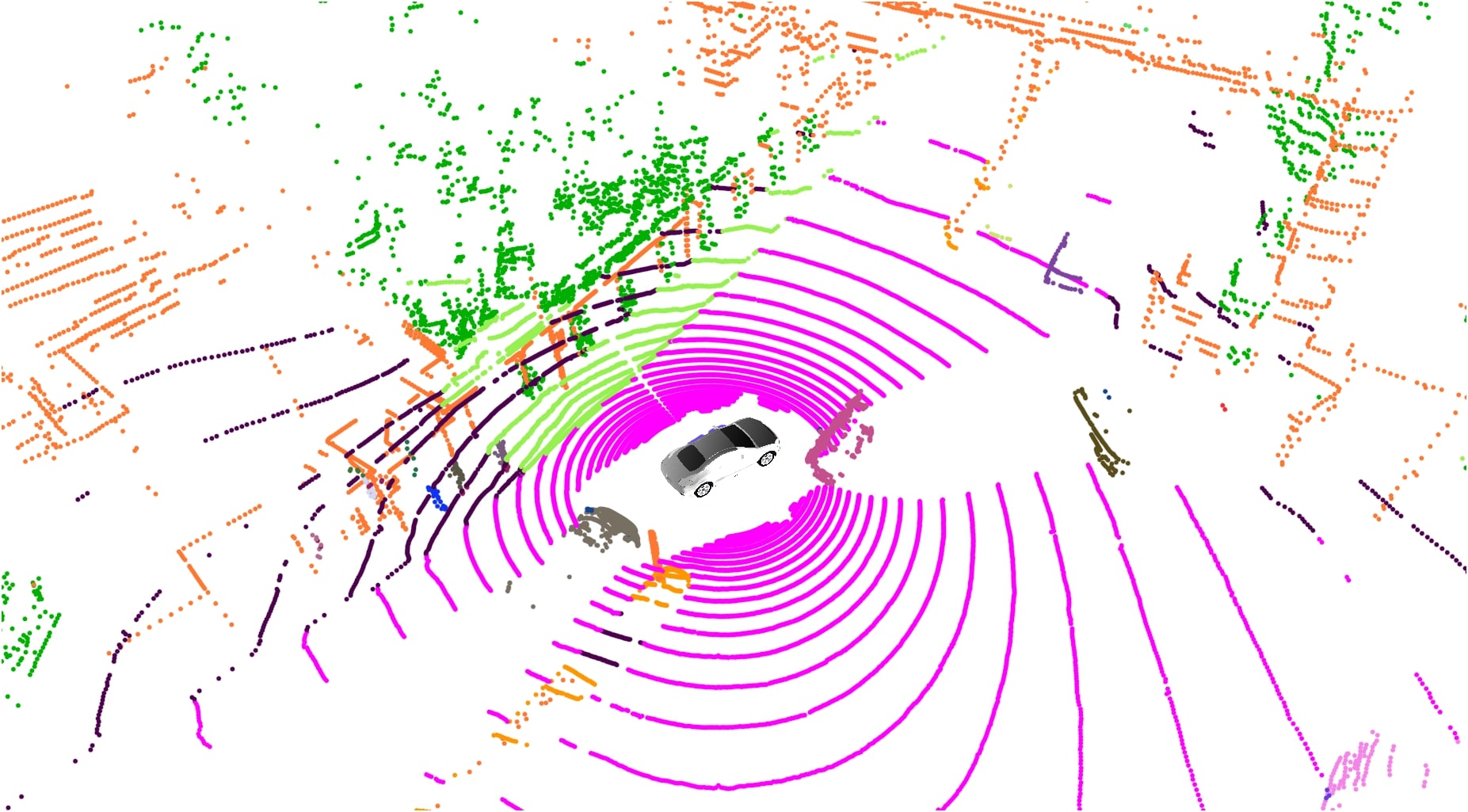}} & \raisebox{-0.4\height}{\includegraphics[width=\linewidth,frame]{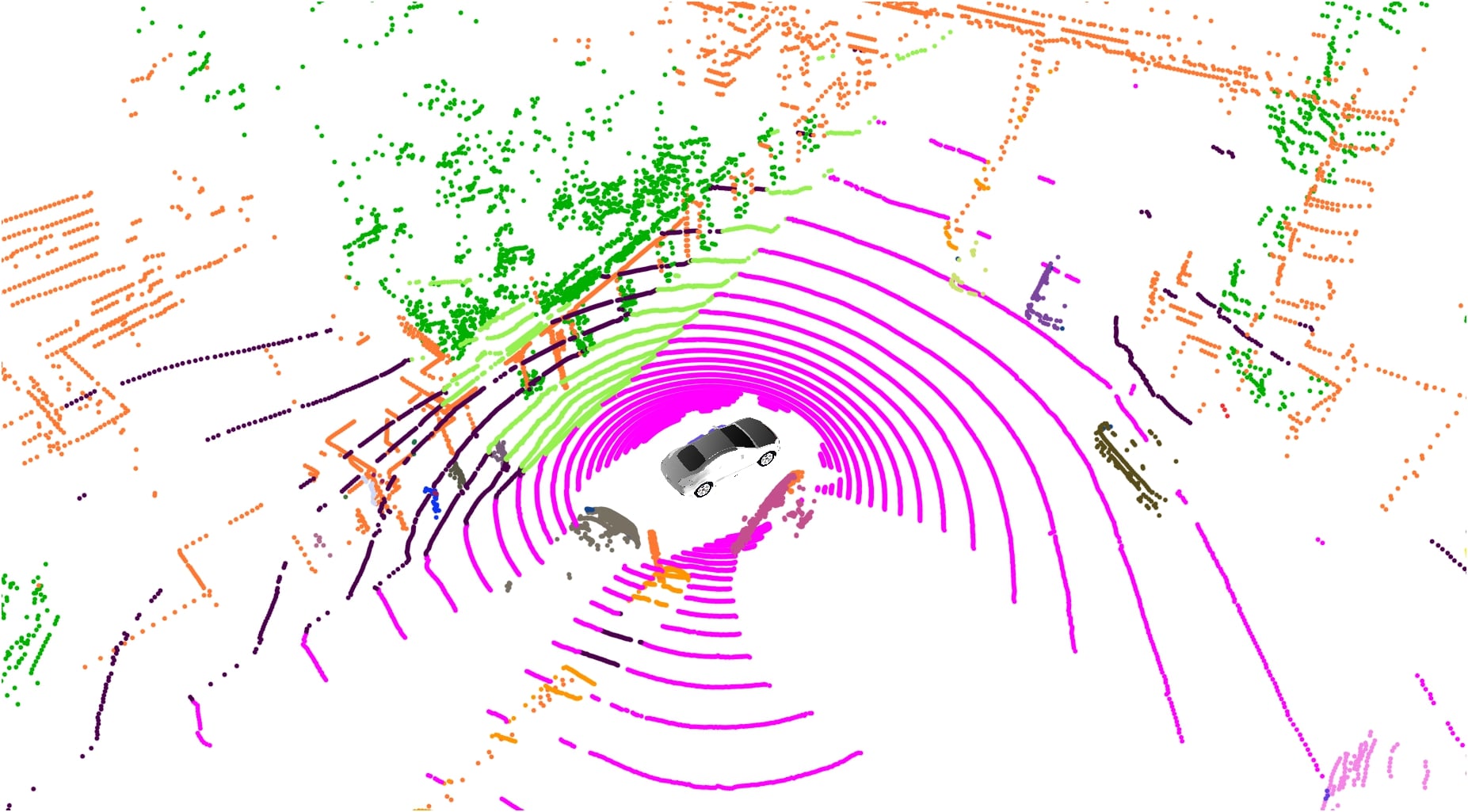}} & \raisebox{-0.4\height}{\includegraphics[width=\linewidth,frame]{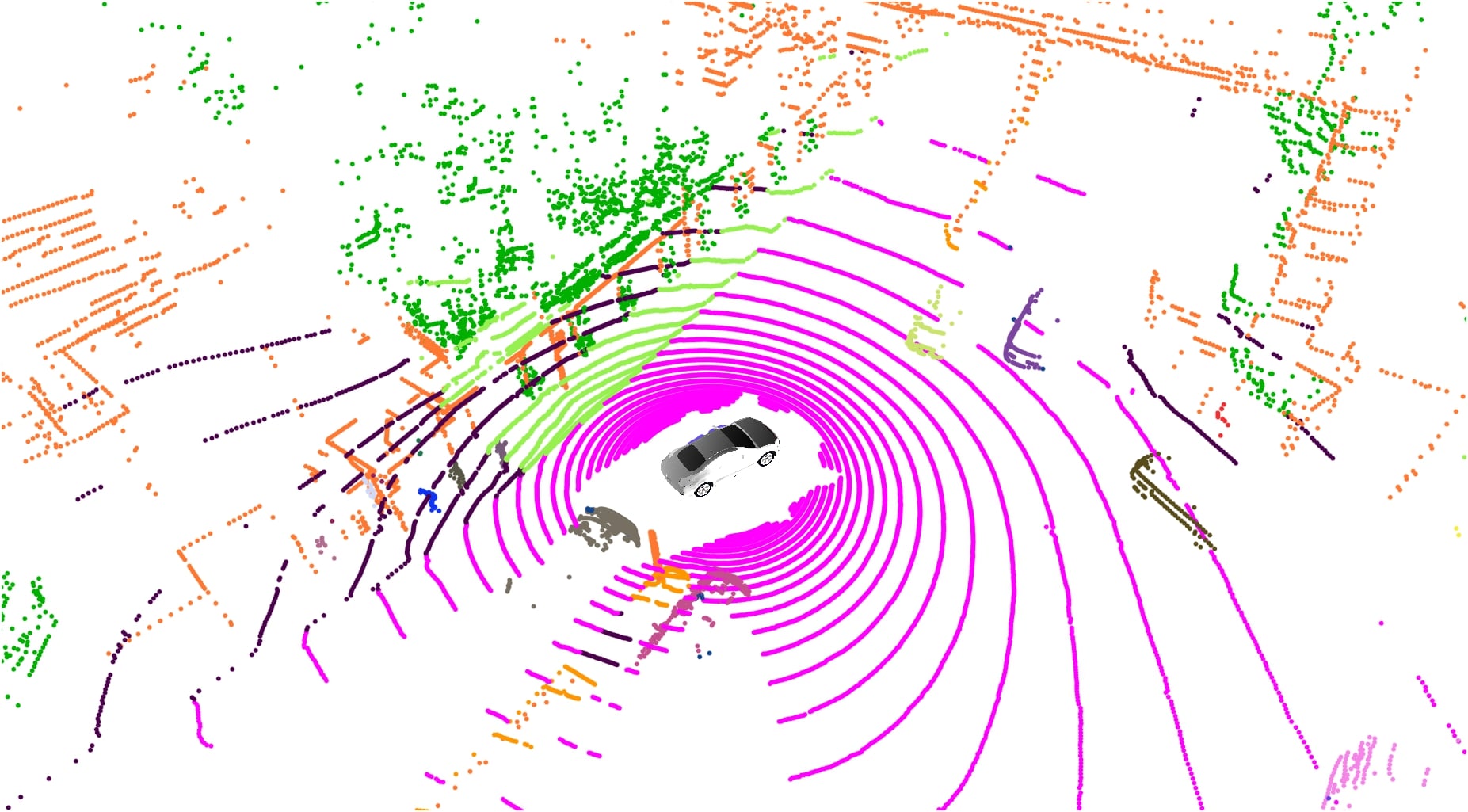}} \\
&\raisebox{-0.4\height}{t-2} & \raisebox{-0.4\height}{t-1} &  \raisebox{-0.4\height}{t} \\
\\
{\rotatebox[origin=c]{90}{(n)}}
&\raisebox{-0.4\height}{\includegraphics[width=\linewidth,frame]{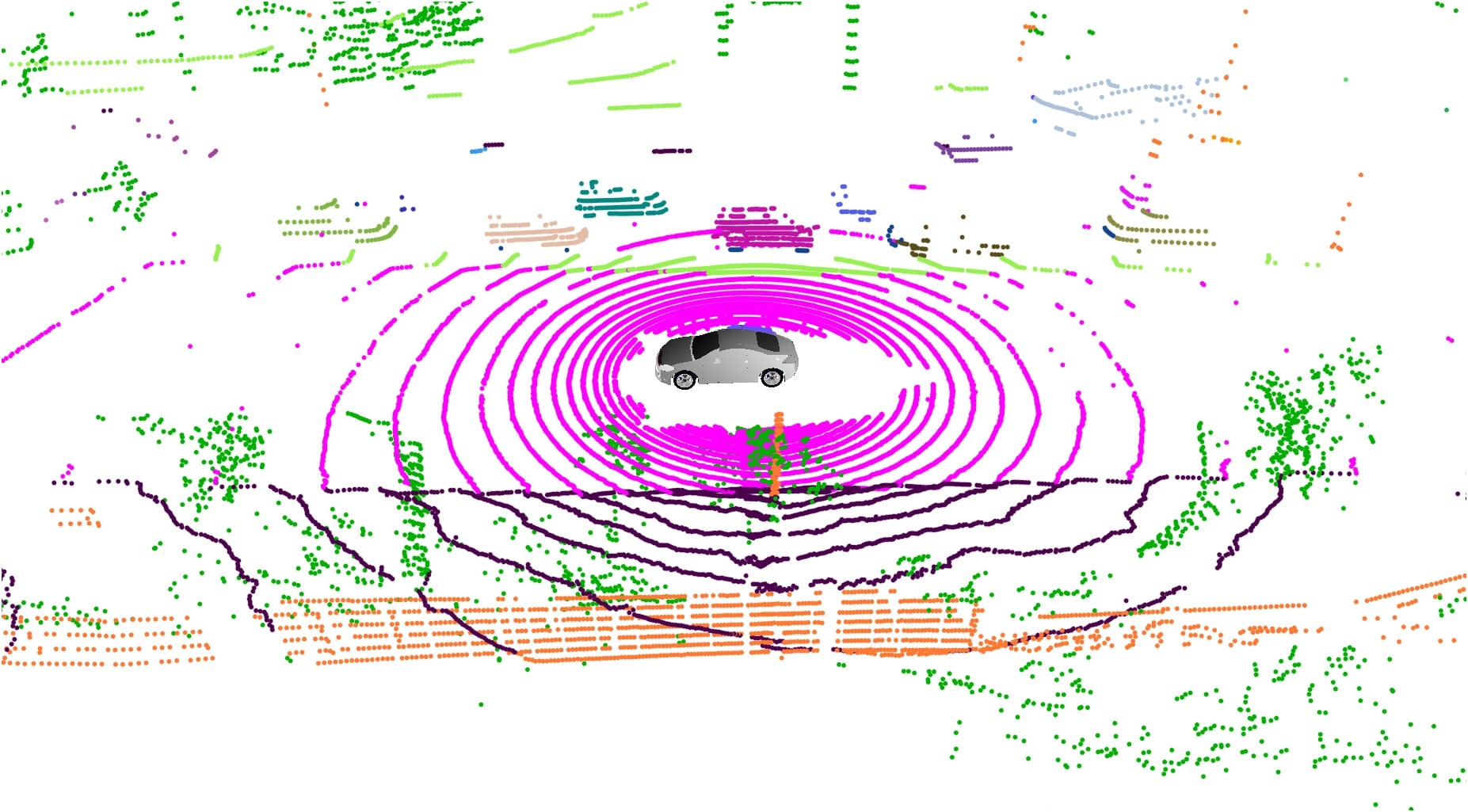}} & \raisebox{-0.4\height}{\includegraphics[width=\linewidth,frame]{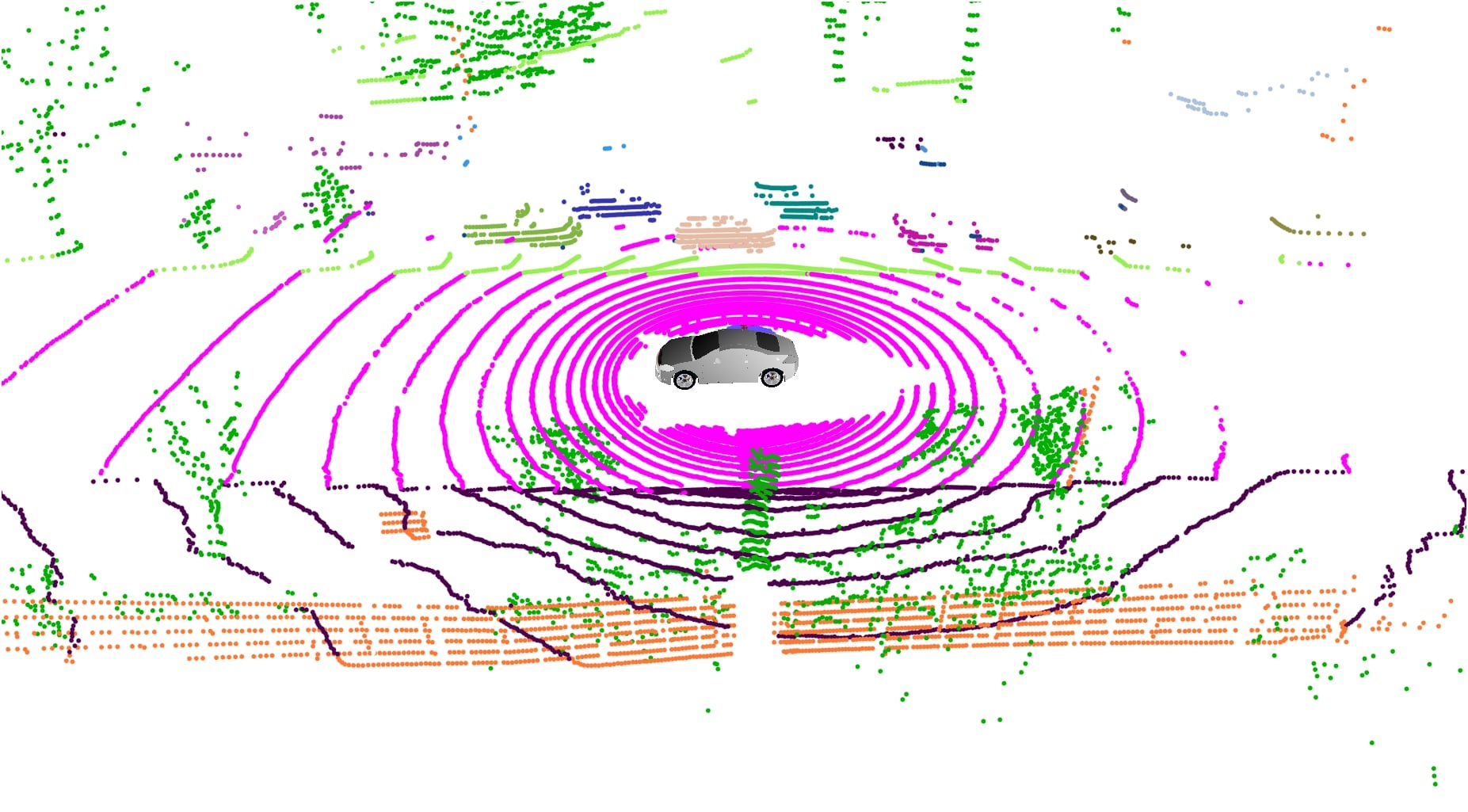}} & \raisebox{-0.4\height}{\includegraphics[width=\linewidth,frame]{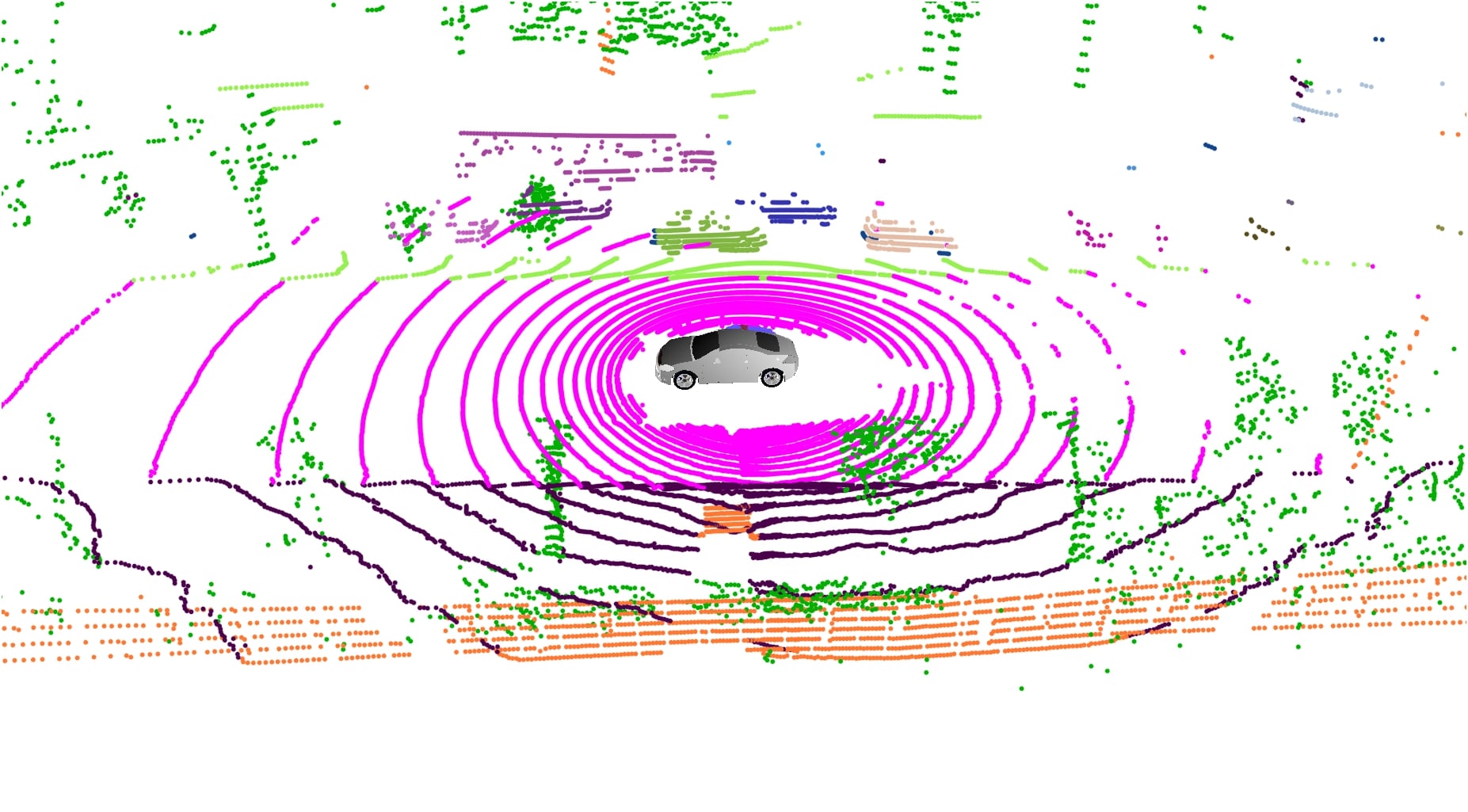}} \\
&\raisebox{-0.4\height}{t-2} & \raisebox{-0.4\height}{t-1} &  \raisebox{-0.4\height}{t} \\
\end{tabular}}
\caption{Example annotations from our Panoptic nuScenes dataset. Figures~(a)-(f) show semantic segmentation annotations, Figures~(g)-(h) show panoptic segmentation annotations, and Figures~(m) and (n) show panoptic tracking annotations.}
\label{fig:visual_ablation}
\end{figure*}

\begin{figure}
\centering
\includegraphics[width=\linewidth]{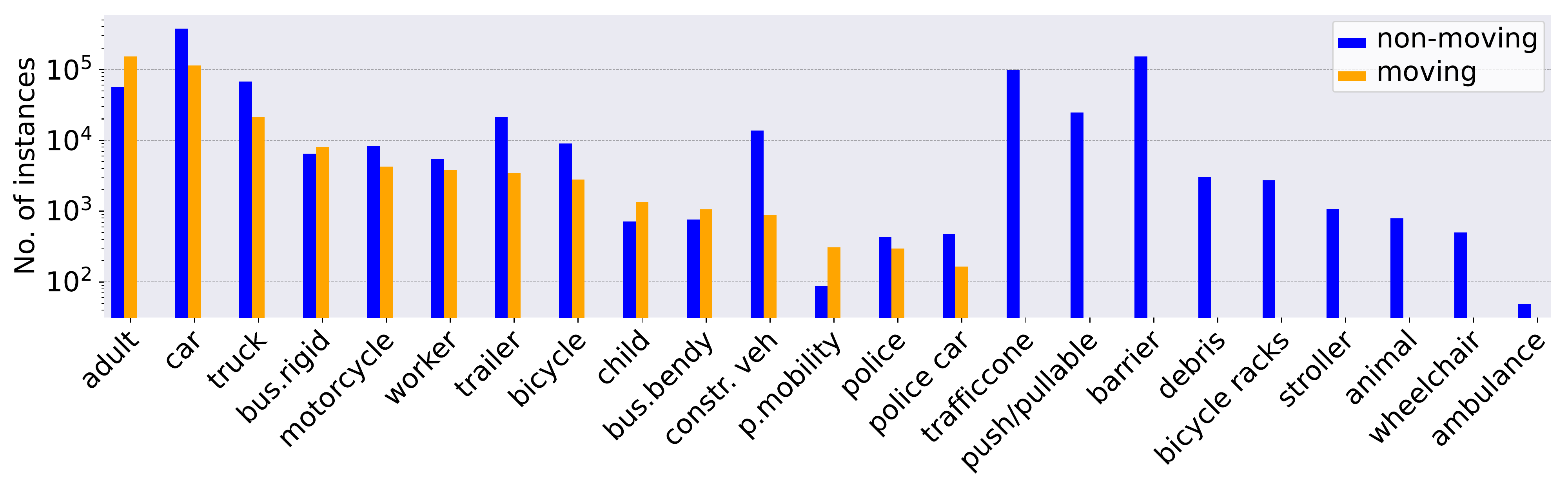}
\caption{Number of scan-wise instances, sorted by the number of moving instances. Note that the y-axis is in log scale.}
\label{fig:instances_scan_wise}
\end{figure}

\noindent\textit{Instance Statistics}: We present additional analysis of object instances that are present in the Panoptic nuScenes dataset, both from a scan-wise and sequence-wise perspective. \figref{fig:instances_scan_wise} shows the distribution of non-moving and moving scan-wise instances for various semantic classes. For common classes such as \emph{adult} and \emph{car}, we have 152k and 114k moving scan-wise instances. For rarer classes such as police and construction vehicles, there are 882 and 298 moving scan-wise instances, which is a non-trivial amount.

\begin{figure}
\centering
\includegraphics[width=\linewidth]{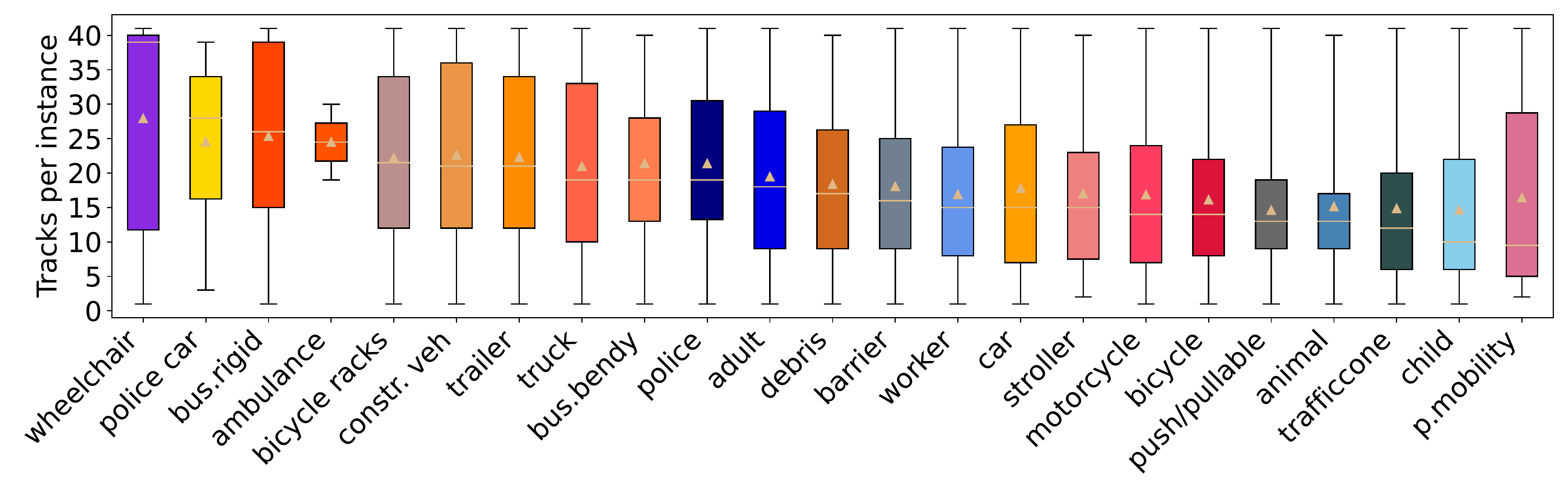}
\caption{Track length per instance for each \emph{thing} class, sorted by median track length. The triangle in a bar indicates the mean track length for the class, and the line indicates the median track length for that class.}
\label{fig:num_frames_per_instance_count}
\end{figure}

\begin{figure}
\centering
\includegraphics[width=\linewidth]{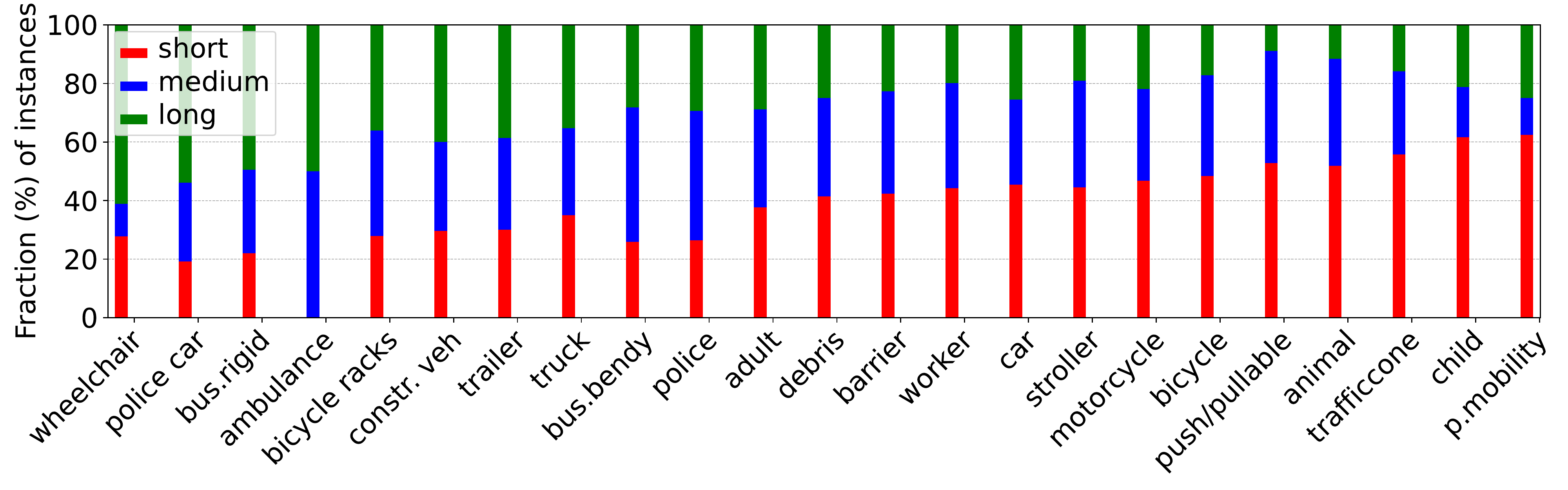}
\caption{Distribution of the track lengths for instances stratified by class. Short tracks are tracks which persist for less than 1/3 of a scene, medium tracks are those that persist for between 1/3 and 2/3 of a scene, and long tracks are those that persist for more than 2/3 of a scene.}
\label{fig:instances_short_medium_long_track_length}
\end{figure}

\figref{fig:num_frames_per_instance_count} further shows the distribution of the track lengths per sequence-wise instance for each semantic class. The median track lengths range from 9.5 frames to 39 frames across classes. For some of the less frequent classes such as \emph{wheelchair} and \emph{construction vehicle}, the median track length is on the higher end, which might be due to these being generally slower moving classes. \figref{fig:instances_short_medium_long_track_length} shows an overview of the track lengths for instances stratified by class. With the non-trivial amount of short, medium, and long tracks, Panoptic nuScenes provides diversity in object track length and across a wide variety of classes. This challenges panoptic tracking approaches to be able to track objects for a relatively sustained period of time, while also being able to handle situations when an object only appears briefly.

\noindent\textit{Panoptic nuScenes - Panoptic Segmentation and Tracking Challenges}: In Panoptic nuScenes, each LiDAR point is annotated as one of 32 semantic classes. However, for the Panoptic nuScenes panoptic segmentation and tracking challenges, we merge similar classes and remove rare or void classes, resulting in 10 \emph{thing} and 6 \emph{stuff} classes as discussed in \secref{sec:tasks_metrics}. \tableref{table:class_map} shows the class mapping from the general Panoptic nuScenes classes to the classes used in the panoptic segmentation and tracking challenges.
In addition, the rightmost column indicates the percentage of points that fall into overlapping bounding boxes for each \emph{thing} class. The panoptic labels for these points are assigned to \emph{noise}. \figref{fig:visual_ablation} presents visualizations of groundtruth annotation examples from our Panoptic nuScenes dataset for each of the three challenges. 

\begin{table}
\setlength\tabcolsep{2.0pt}
\footnotesize
\centering
\caption{Mapping from the Panoptic nuScenes classes to the challenge classes. Note that most prefixes for the former are omitted for brevity. ($^{\dagger}$) We use \emph{void} to denote classes that have been excluded from the challenges. ($^{\ddagger}$) The rightmost column shows the percentage of points within overlapping bounding boxes per class.}
\begin{tabular}{|l | C{2.0cm} | C{1.5cm}| C{1.5cm} |}
\toprule
General Panoptic Class &  Challenge Class$^{\dagger}$ &  Thing/Stuff & Overlap Percent ($\%$)$^{\ddagger}$ \\ 
\midrule
animal                                   &   void                  & thing    & 0      \\
personal mobility                        &   void                  & thing   & 0       \\
stroller                                 &   void                  & thing   & 0       \\
wheelchair                               &   void                  & thing  & 0        \\
debris                                   &   void                  & thing  & 0.12        \\
pushable                                 &   void                  & thing  & 0.7        \\
bicycle rack                             &   void                  & thing  & 0        \\ 
ambulance                                &   void                  & thing  & 0        \\ 
police                                   &   void                  & thing  & 0        \\
noise                                    &   void                  & -    & -          \\ 
other                                    &   void                  & -    & -          \\ 
ego                                      &   void                  & -    & -          \\ 
barrier                                  &   barrier               & thing  & 0.6        \\ 
bicycle                                  &   bicycle               & thing  & 0.72        \\ 
bus.bendy                                &   bus                   & thing   & 0.59       \\ 
bus.rigid                                &   bus                   & thing   & 0       \\ 
car                                      &   car                   & thing   & 0.00017       \\ 
construction                             &   construction vehicle  & thing   & 0.02       \\ 
motorcycle                               &   motorcycle            & thing   & 0.0007       \\ 
adult                                    &   pedestrian            & thing  & 0.33        \\ 
child                                    &   pedestrian            & thing  & 0.041        \\ 
construction worker                      &   pedestrian            & thing  & 0.146       \\ 
police officer                           &   pedestrian            & thing  & 0        \\ 
traffic cone                              &   traffic cone         & thing   & 0.07       \\ 
trailer                                  &   trailer               & thing   & 0.02       \\ 
truck                                    &   truck                 & thing   & 0.00249       \\ 
driveable surface                        &   driveable surface     & stuff  & -        \\ 
other                                    &   other flat            & stuff   & -       \\ 
sidewalk                                 &   sidewalk              & stuff   & -      \\ 
terrain                                  &   terrain               & stuff   & -       \\ 
manmade                                  &   manmade               & stuff   & -       \\ 
vegetation                               &   vegetation            & stuff  & -        \\
\bottomrule
\end{tabular}
\label{table:class_map}
\end{table}

\section{Additional Panoptic Segmentation Results}

In \tableref{tab:nuScenesClass}, we present a comparison of per-class panoptic segmentation results on the Panoptic nuScenes dataset. We compare the top three of our combination baselines that have published task-specific methods and the end-to-end baselines. Across all the classes, it can be seen that the independently combined baselines outperform the end-to-end methods. This is more pronounced for the \emph{thing} classes compared to the \emph{stuff} classes. Among the \emph{thing} classes, the average gap between the best-in-class result of the independently combined baselines compared to that of the end-to-end methods is 20.8 PQ, while for the \emph{stuff classes}, the average gap is 4.6 PQ. This is likely due to the role of the stronger task-specific detection methods that are used for the instance segmentation of the \emph{thing} classes.

\begin{table*}
\setlength\tabcolsep{3.7pt}
\centering
\caption{Class-wise panoptic segmentation results on the Panoptic nuScenes dataset. All scores are in [\%].}
\label{tab:nuScenesClass}
\begin{tabular}{ll|cccccccccccccccc|c}
\midrule
& Method & \begin{sideways}barrier\end{sideways} & \begin{sideways}bicycle\end{sideways} & \begin{sideways}bus\end{sideways} & \begin{sideways}car\end{sideways} & \begin{sideways}cvehicle\end{sideways} & \begin{sideways}motorcycle\end{sideways} & \begin{sideways}pedestrian\end{sideways} & \begin{sideways}traffic cone\end{sideways} & \begin{sideways}trailer\end{sideways} & \begin{sideways}truck\end{sideways} & \begin{sideways}driveable\end{sideways} & \begin{sideways}other flat\end{sideways} & \begin{sideways}sidewalk\end{sideways} & \begin{sideways}terrain\end{sideways} & \begin{sideways}man-made\end{sideways} & \begin{sideways}vegetation\end{sideways}  & PQ \\
\midrule
\multirow{3}{*}{\rotatebox[origin=c]{90}{val set}} & PanopticTrackNet~\cite{hurtado2020mopt} & 47.1 & 27.9 & 57.9 & 66.3 & 22.8 & 51.1 & 42.8 & 46.8 & 38.9 & 51.0 & 77.5 & 41.6 & 59.7 & 42.3 & 60.1 & 68.8 & 51.4  \\
& EfficientLPS~\cite{sirohi2021efficientlps} & 56.8& 37.8 & 52.4 & \textbf{75.6} & 32.1 & 65.1 & \textbf{74.9} & 73.5 & \textbf{49.9} & 49.7 & \textbf{95.2} & 43.9 & \textbf{67.5} & \textbf{52.8} & 81.8 & \textbf{82.4} & 62.0  \\
& PolarSeg-Panoptic~\cite{zhou2021panoptic} & \textbf{56.9} & \textbf{41.1} & \textbf{65.5} & 75.4 & \textbf{38.9} & \textbf{67.7} & 73.2 & \textbf{74.9} & 40.4 & \textbf{58.6} & 94.2 & \textbf{46.6} & 66.2 & 49.4 & \textbf{83.9} & 82.0 & \textbf{63.4} \\
\midrule
\multirow{6}{*}{\rotatebox[origin=c]{90}{test set}} 
& PanopticTrackNet~\cite{hurtado2020mopt} & 44.3 & 23.1 & 45.8 & 71.5 & 21.8 & 42.7 & 62.3 & 63.7 & 40.2 & 43.6 & 93.0 & 23.8 & 51.2 & 42.6 & 75.9 & 79.3 & 51.6\\
& EfficientLPS~\cite{sirohi2021efficientlps} & 57.1 & 38.1 & 53.6 & 75.8 & 31.6 & 66.0 & 75.2 & 74.3 & 50.6 & 49.4 & 95.3 & 44.1 & 68.2 & 53.3 & 82.3 & 83.4 & 62.4 \\
& PolarSeg-Panoptic~\cite{zhou2021panoptic} & 55.9 & 39.6 & 55.7 & 77.4 & 37.4 & 67.6 & 76.5 & 73.7 & 52.4 & 53.5 & 95.0 & 44.6 & 67.4 & 53.2 & 82.9 & 84.4 & 63.6 \\
& SPVNAS~\cite{tang2020searching} + CenterPoint~\cite{yin2021center} & 78.1 & 62.3 & 59.4 & 91.2 & 46.5 & 83.2 & 92.9 & 90.7 & 49.0 & 63.4 & 97.4 & 49.4 & 72.6 & 54.5 & 85.4 & 79.9 & 72.2 \\
& Cylinder3D++~\cite{zhu2021cylindrical} + CenterPoint~\cite{yin2021center} & 79.2 & 70.5 & 63.8 & \textbf{92.8} & 56.8 & 83.3 & 93.6 & 92.8 & 66.1 & 69.0 & \textbf{97.6} & \textbf{55.1} & \textbf{74.1} & \textbf{57.1} & \textbf{87.0} & \textbf{85.1} & 76.5 \\
& (AF)\textsuperscript{2}-S3Net~\cite{cheng20212} + CenterPoint~\cite{yin2021center} & \textbf{80.8} & \textbf{79.4} & \textbf{67.3} & 91.7 & \textbf{63.8} & \textbf{86.4} & \textbf{93.8} & \textbf{93.7} & \textbf{68.7} & \textbf{72.9} & 97.1 & 45.6 & 69.4 & 51.5 & 84.6 & 82.3 & \textbf{76.8} \\
\bottomrule
\end{tabular}
\end{table*}

\section{Correlation Analysis}

In this section, we extend the analysis of the combination baselines to study the relationships between (i) panoptic segmentation performance and the component semantic segmentation and detection performance, and (ii) panoptic tracking performance and the component semantic segmentation and tracking performance.

\begin{figure*}
    \footnotesize
    \centering
    \begin{tabular}{c c}
    \includegraphics[width=0.45\linewidth]{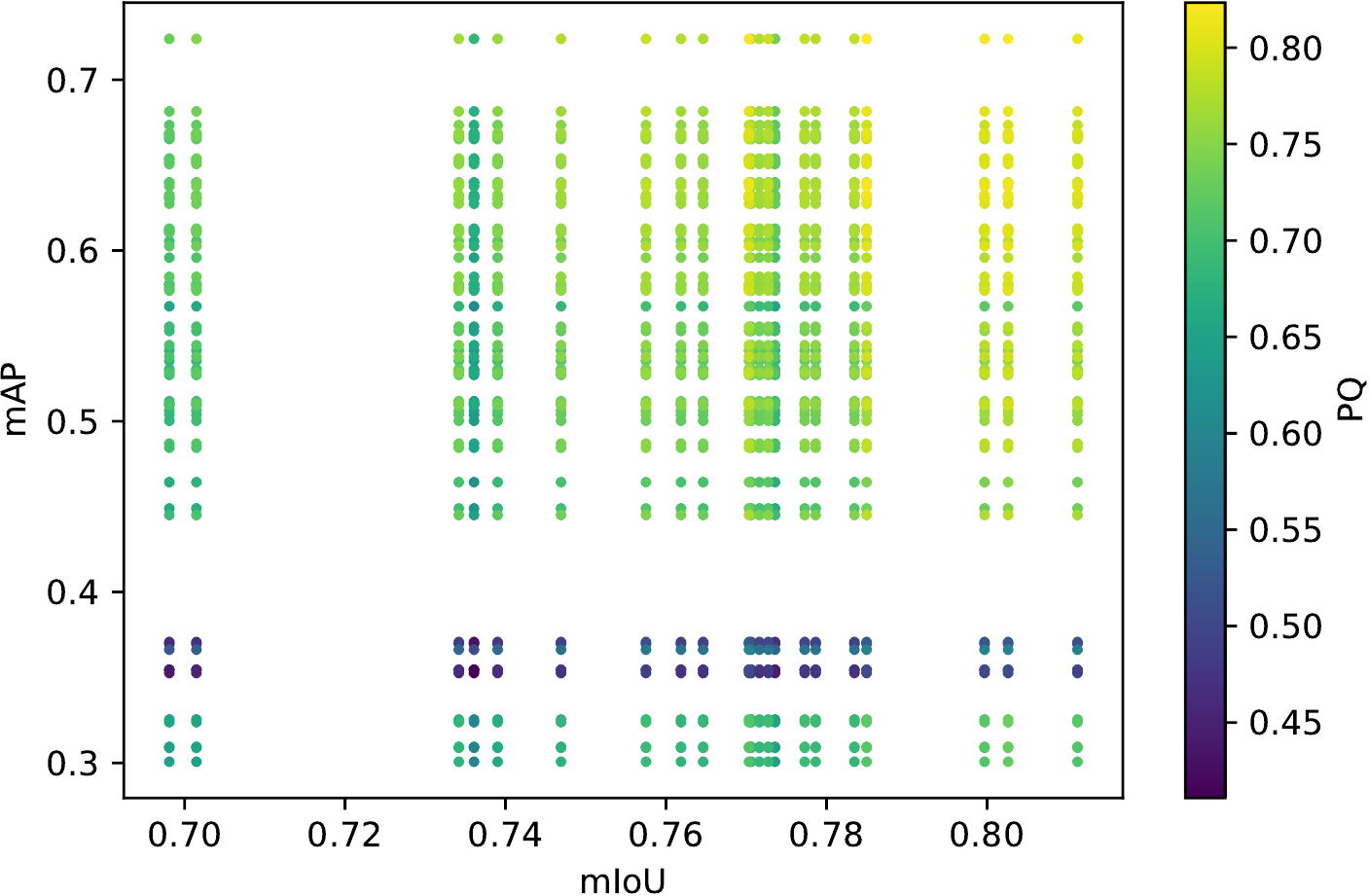} & \includegraphics[width=0.45\linewidth]{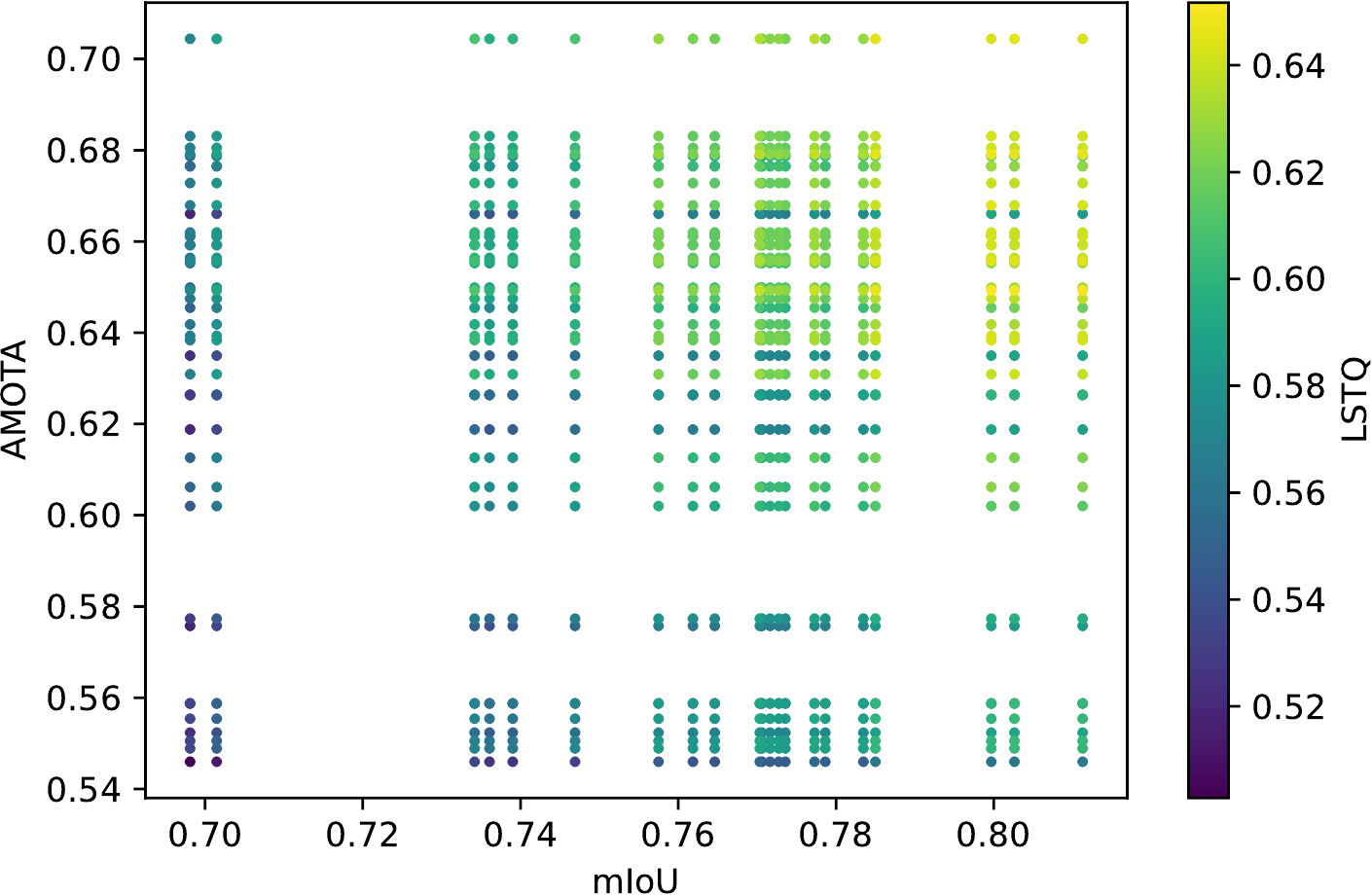} \\
    \\
    (a) Panoptic segmentation performance (PQ) & (b) Panoptic tracking performance (LSTQ) \\
    \end{tabular}
    \caption{Performance of panoptic segmentation and panotic tracking in relation to LiDAR semantic segmentation (mIoU) and detection (mAP) / tracking (AMOTA) performance respectively based on the combined baselines from LiDAR semantic segmentation and LiDAR detection / tracking methods.}
    \label{fig:mIoU_mAP_AMOTA_PQ_LSTQ}
\end{figure*}

\noindent\textit{Panoptic Segmentation}: As discussed in \secref{subsec:baseline_results}, panoptic segmentation can be achieved by combining the submissions of LiDAR semantic segmentation and LiDAR detection from the nuScenes challenges. We generate 1470 of these combinations and in \figref{fig:mIoU_mAP_AMOTA_PQ_LSTQ}(a), we show an overview of the resulting panoptic segmentation performance as measured by the PQ score, for the combinations. Within each column, a perceptual change in color can be observed, implying that for the same LiDAR semantic segmentation method, the change in detection method influences the PQ. However, across each row, the change in color is insignificant, indicating that varying the LiDAR semantic segmentation method does not impact PQ as much as the LiDAR detection method.

\begin{table}
    \footnotesize
    \centering
    \caption{Correlation of LiDAR semantic segmentation methods (measured by mIoU) and tracking methods (measured by AMOTA) with various panoptic tracking metrics. The correlations are calculated from the 924 panoptic tracking baselines that we generated by combining various LiDAR semantic segmentation and tracking methods.}
    \begin{tabular}{l|c c}
    \toprule
         Metric         & AMOTA  & mIoU  \\
         \midrule
         LSTQ~\cite{aygun20214d}           & 0.55   & 0.62  \\
         PTQ~\cite{hurtado2020mopt}            & 0.23   & 0.69  \\
         PAT (Ours)           & 0.48   & 0.57  \\
         \bottomrule
    \end{tabular}
    \label{tab:correlations}
    \vspace{1cm}
\end{table}

\begin{figure*}
\centering
\footnotesize
{\renewcommand{\arraystretch}{1}
\begin{tabular}{P{0.4cm}P{5.5cm}P{5.5cm}P{5.5cm}}
{\rotatebox[origin=c]{90}{(a)}}&\raisebox{-0.4\height}{t-2} & \raisebox{-0.4\height}{t-1} &  \raisebox{-0.4\height}{t} \\
\\
{\rotatebox[origin=c]{90}{Ground Truth}}&\raisebox{-0.4\height}{\includegraphics[width=\linewidth,frame]{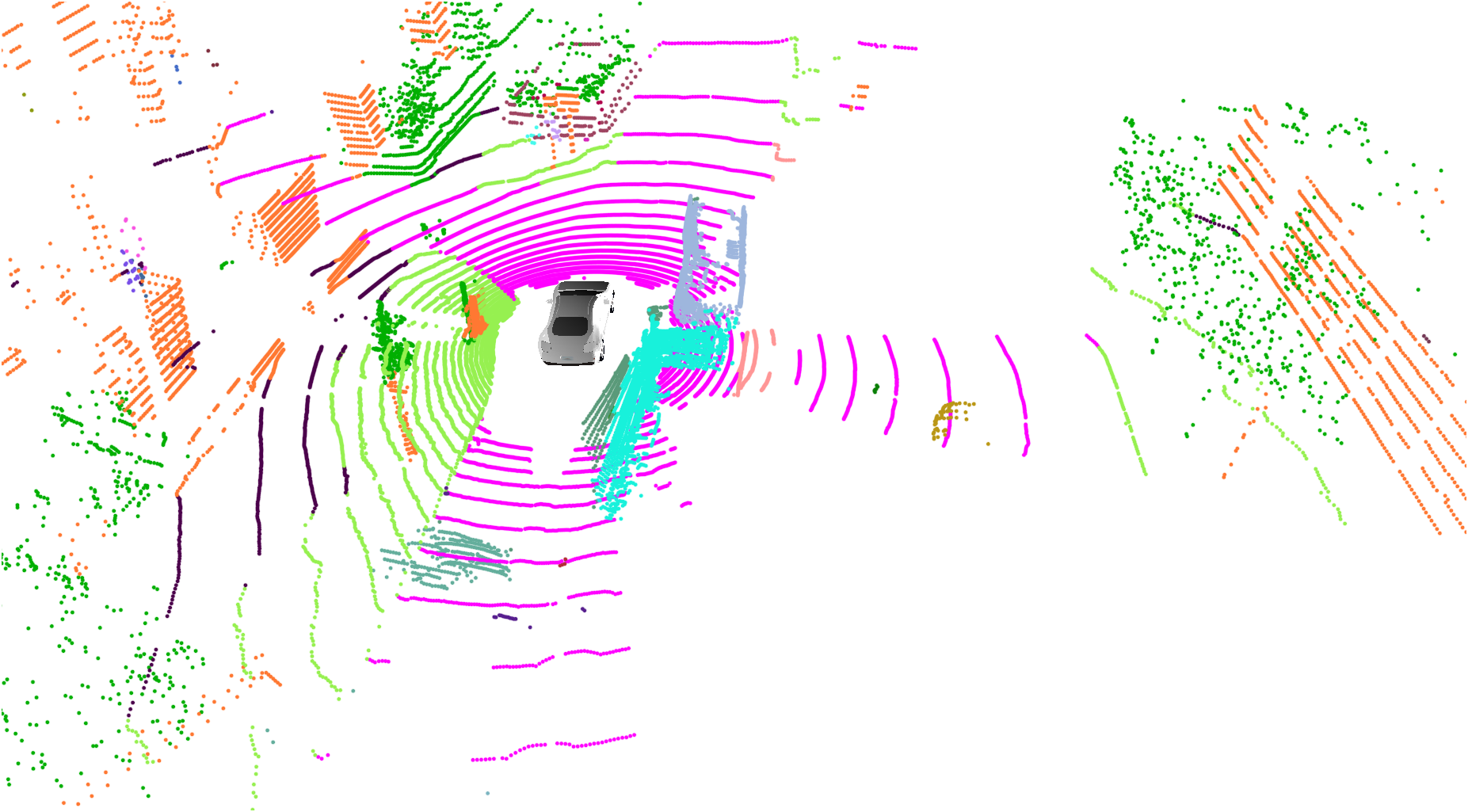}} & \raisebox{-0.4\height}{\includegraphics[width=\linewidth,frame]{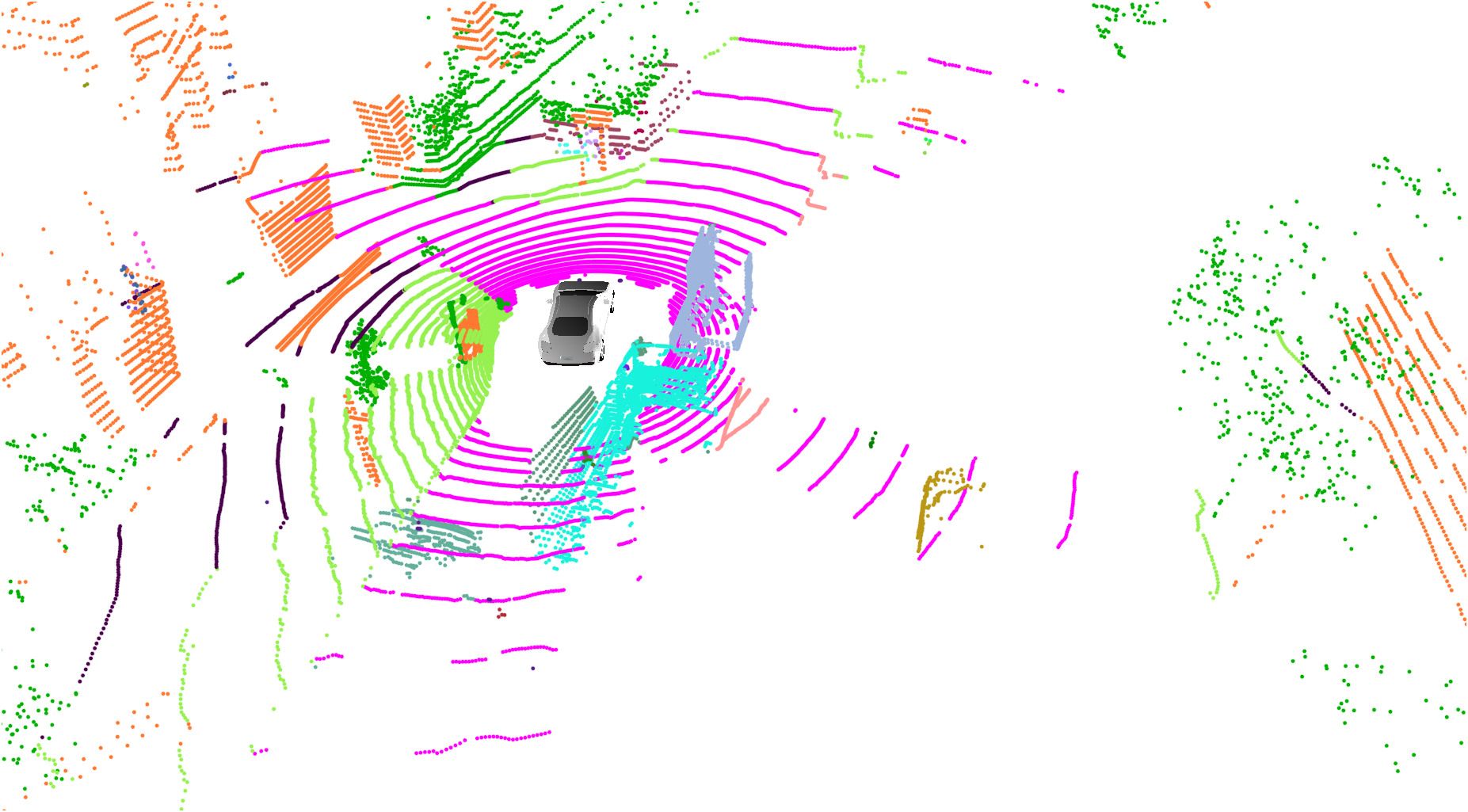}} & \raisebox{-0.4\height}{\includegraphics[width=\linewidth,frame]{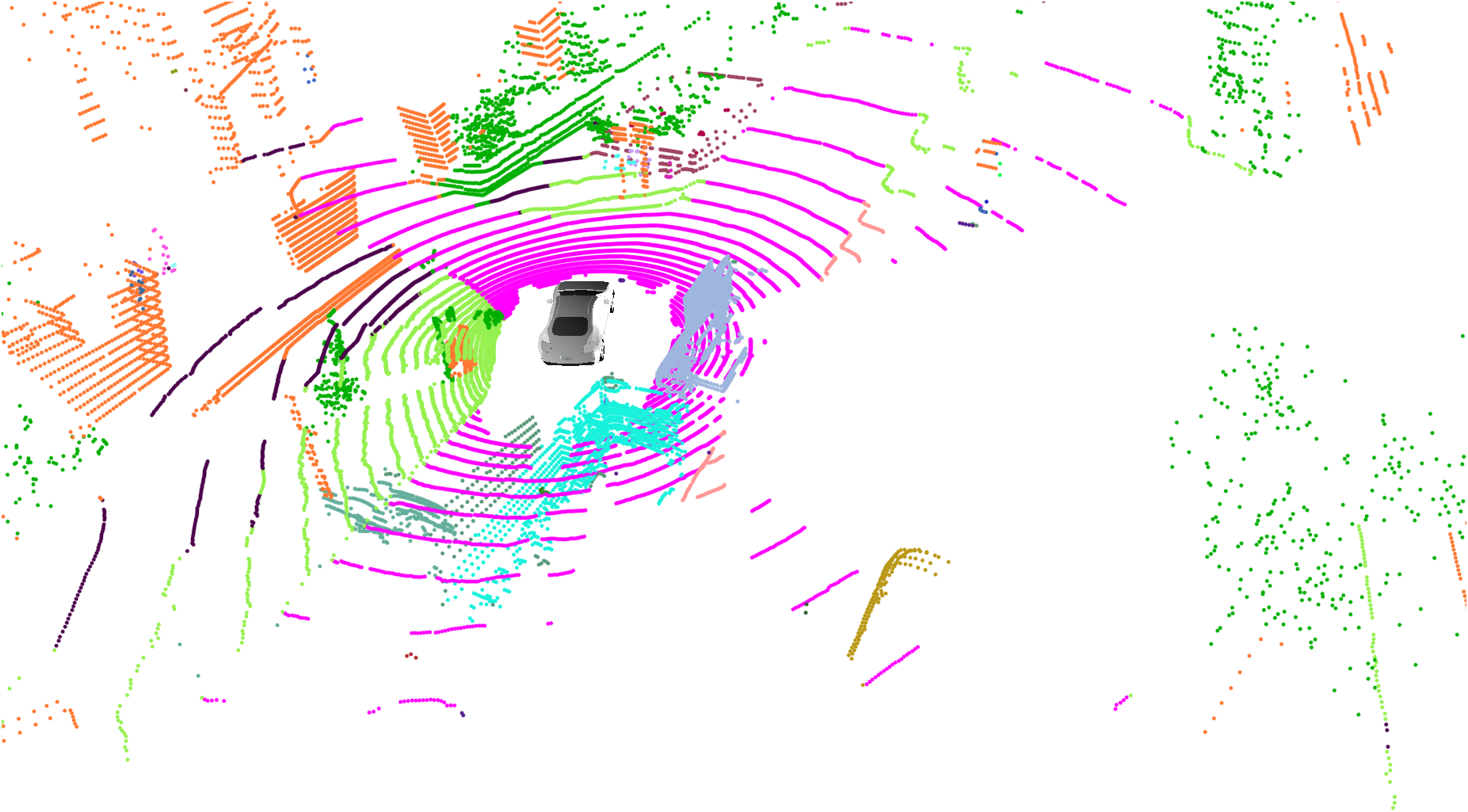}} \\
\\
{\rotatebox[origin=c]{90}{4D-PLS}}&\raisebox{-0.4\height}{\includegraphics[width=\linewidth,frame]{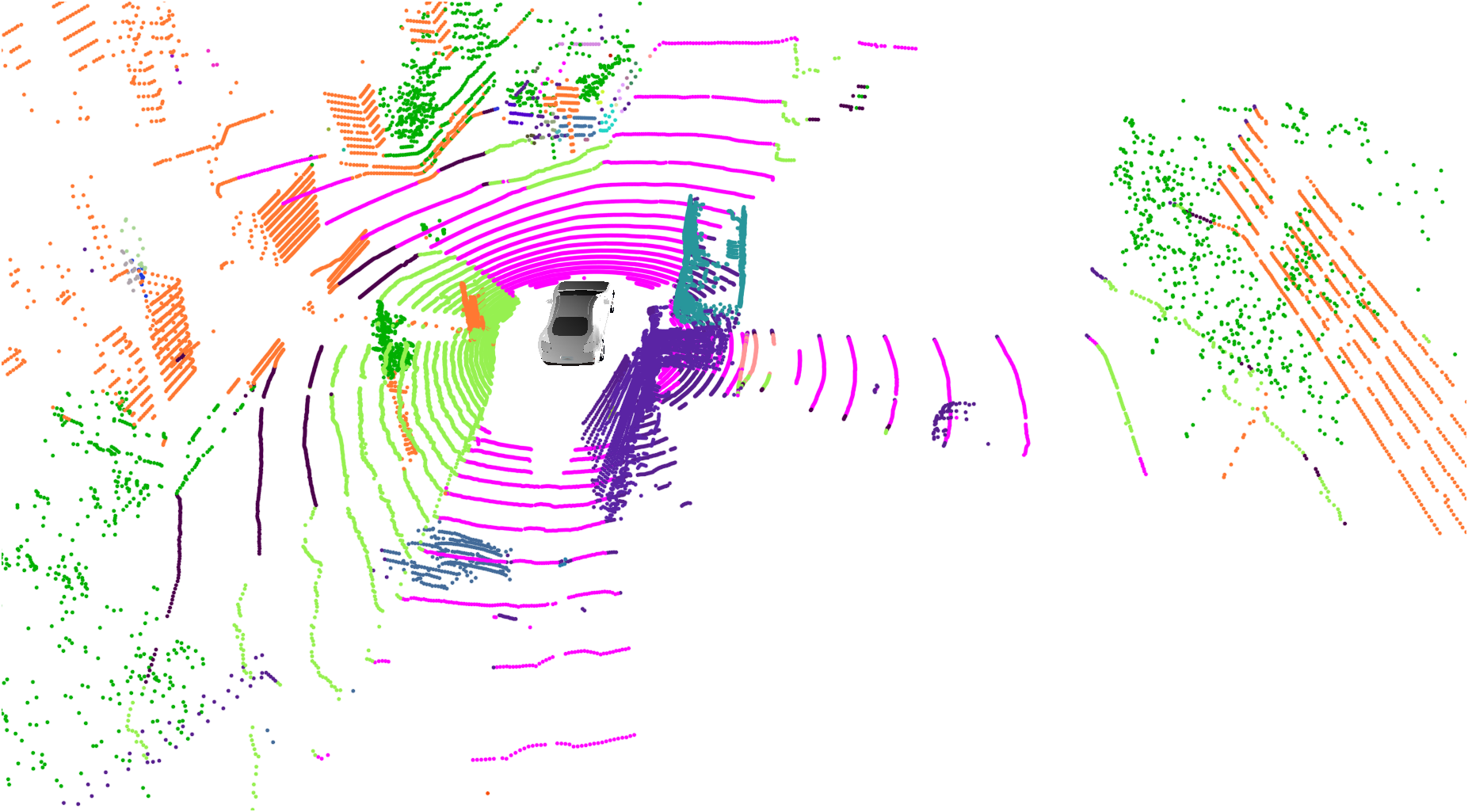}} & \raisebox{-0.4\height}{\includegraphics[width=\linewidth,frame]{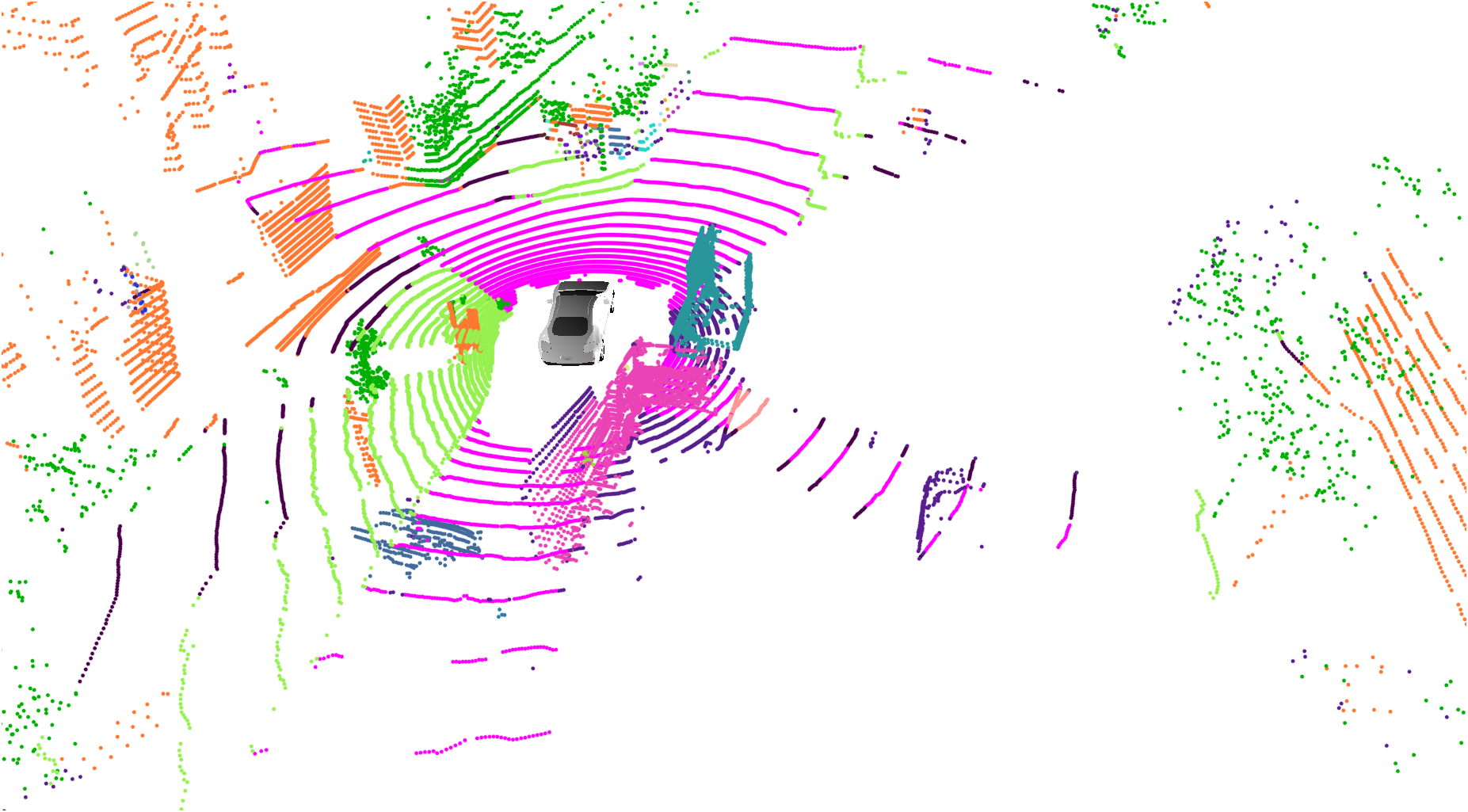}} & \raisebox{-0.4\height}{\includegraphics[width=\linewidth,frame]{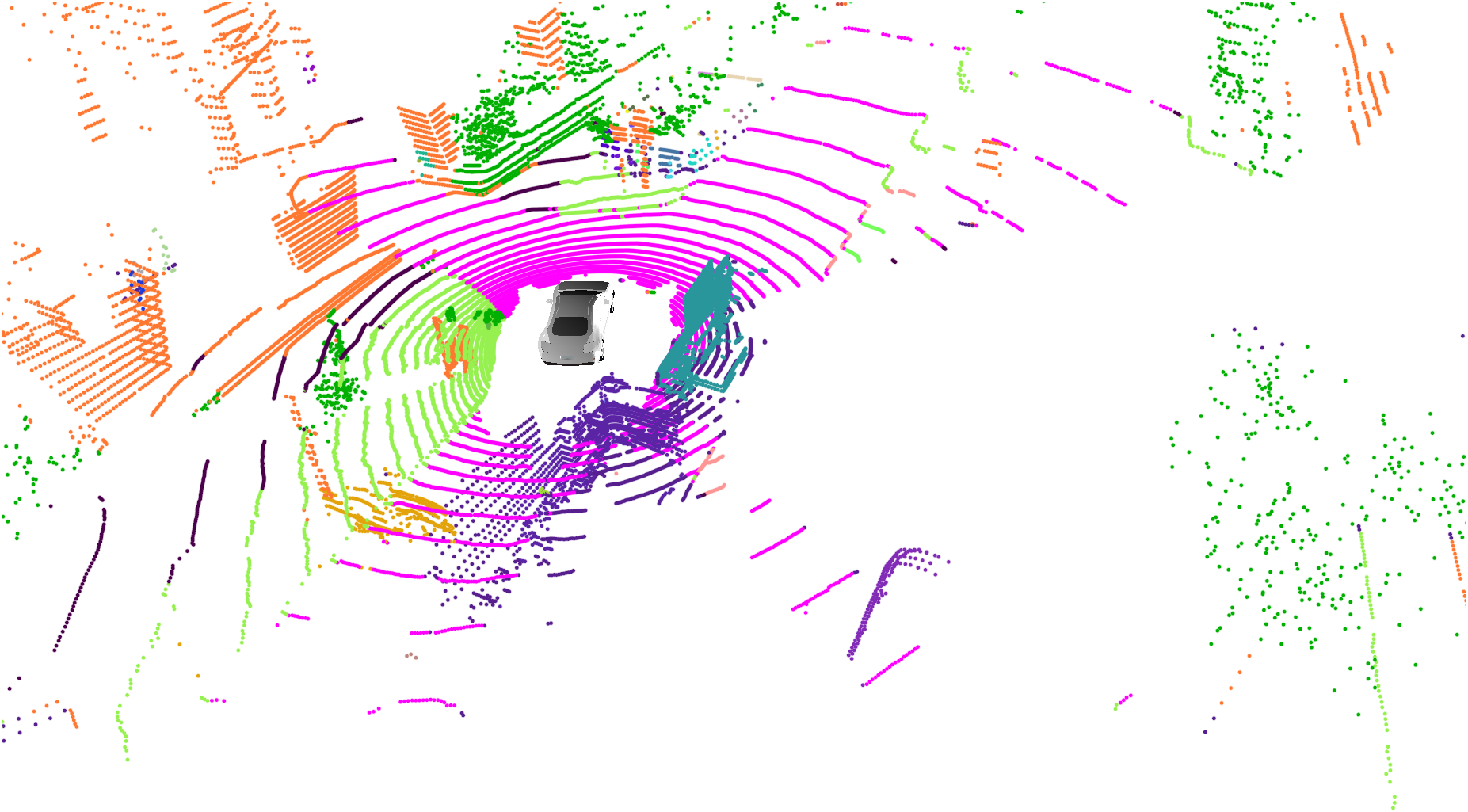}} \\
\\
{\rotatebox[origin=c]{90}{EfficientLPS + Kalman Filter}}&\raisebox{-0.4\height}{\includegraphics[width=\linewidth,frame]{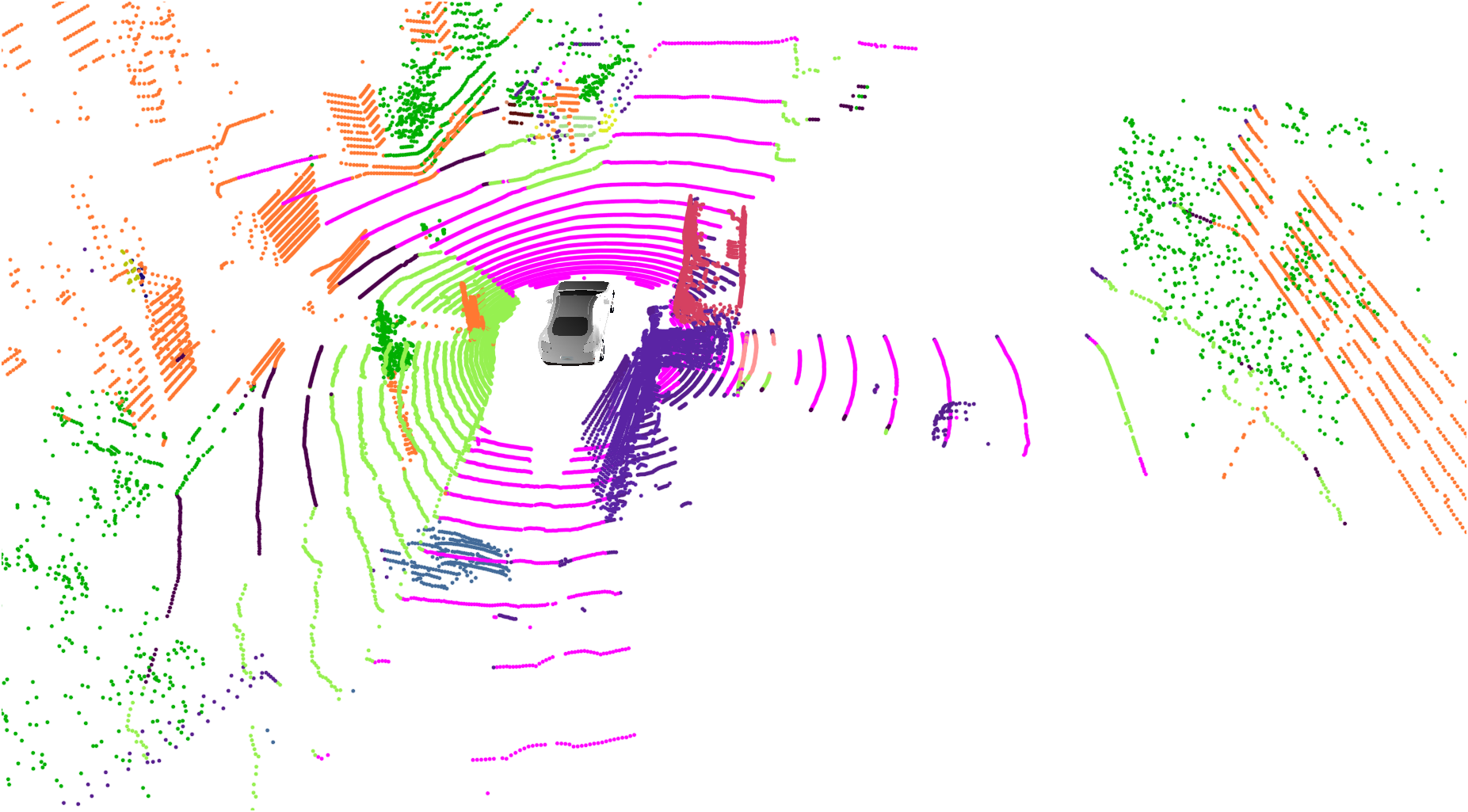}} & \raisebox{-0.4\height}{\includegraphics[width=\linewidth,frame]{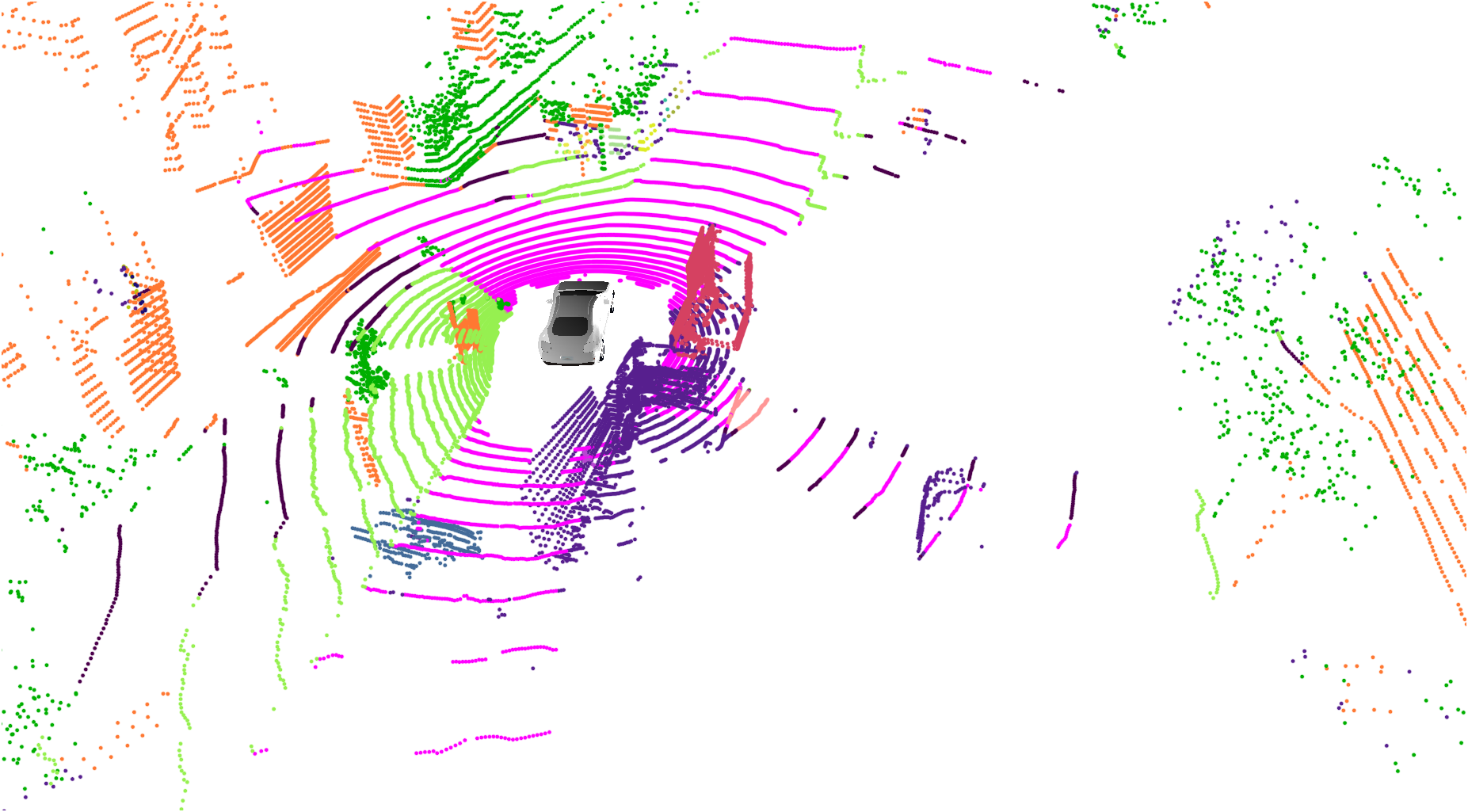}} & \raisebox{-0.4\height}{\includegraphics[width=\linewidth,frame]{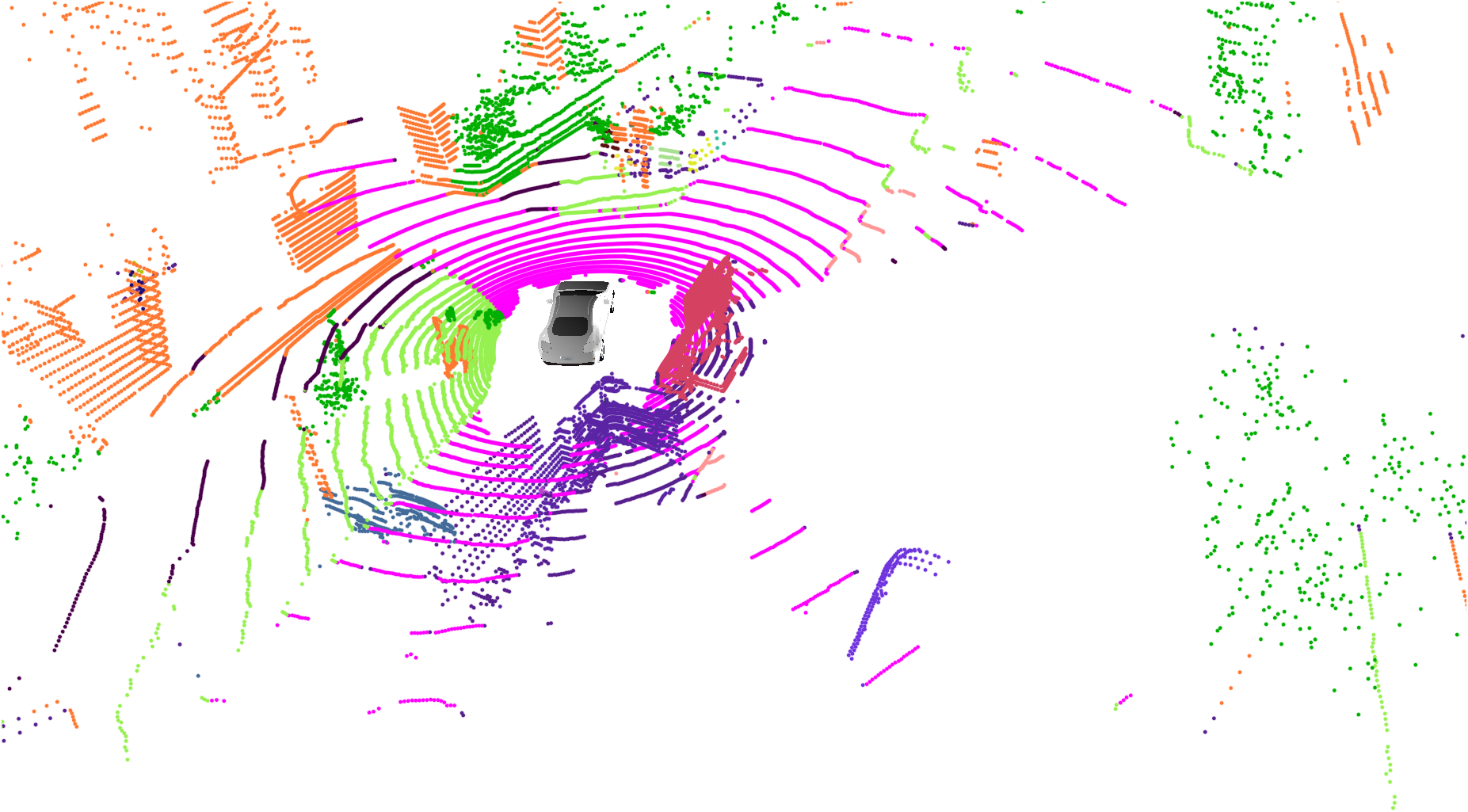}} \\
\\\midrule
\\
{\rotatebox[origin=c]{90}{(b)}}&\raisebox{-0.4\height}{t-2} & \raisebox{-0.4\height}{t-1} &  \raisebox{-0.4\height}{t} \\
\\
{\rotatebox[origin=c]{90}{Ground Truth}}&\raisebox{-0.4\height}{\includegraphics[width=\linewidth,frame]{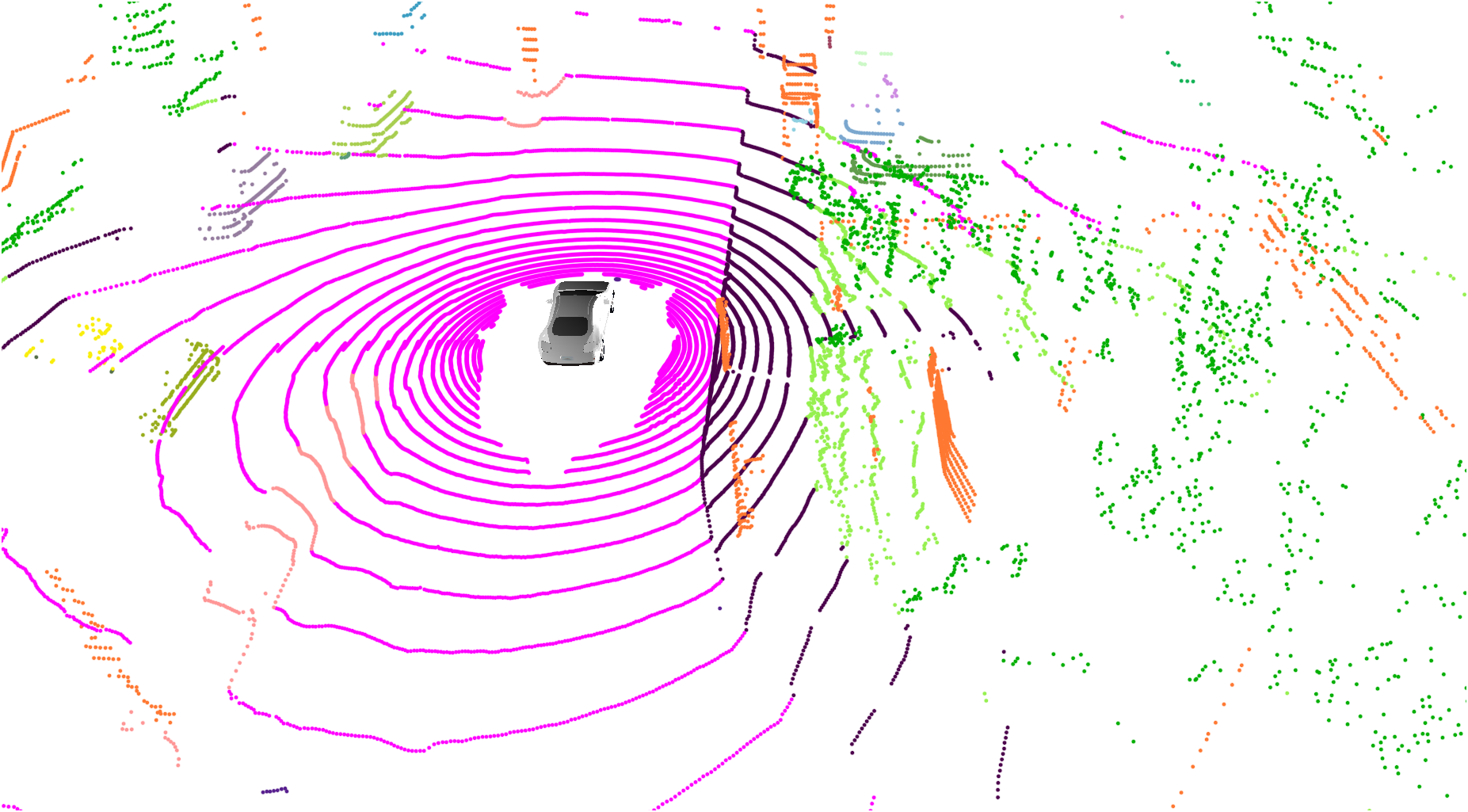}} & \raisebox{-0.4\height}{\includegraphics[width=\linewidth,frame]{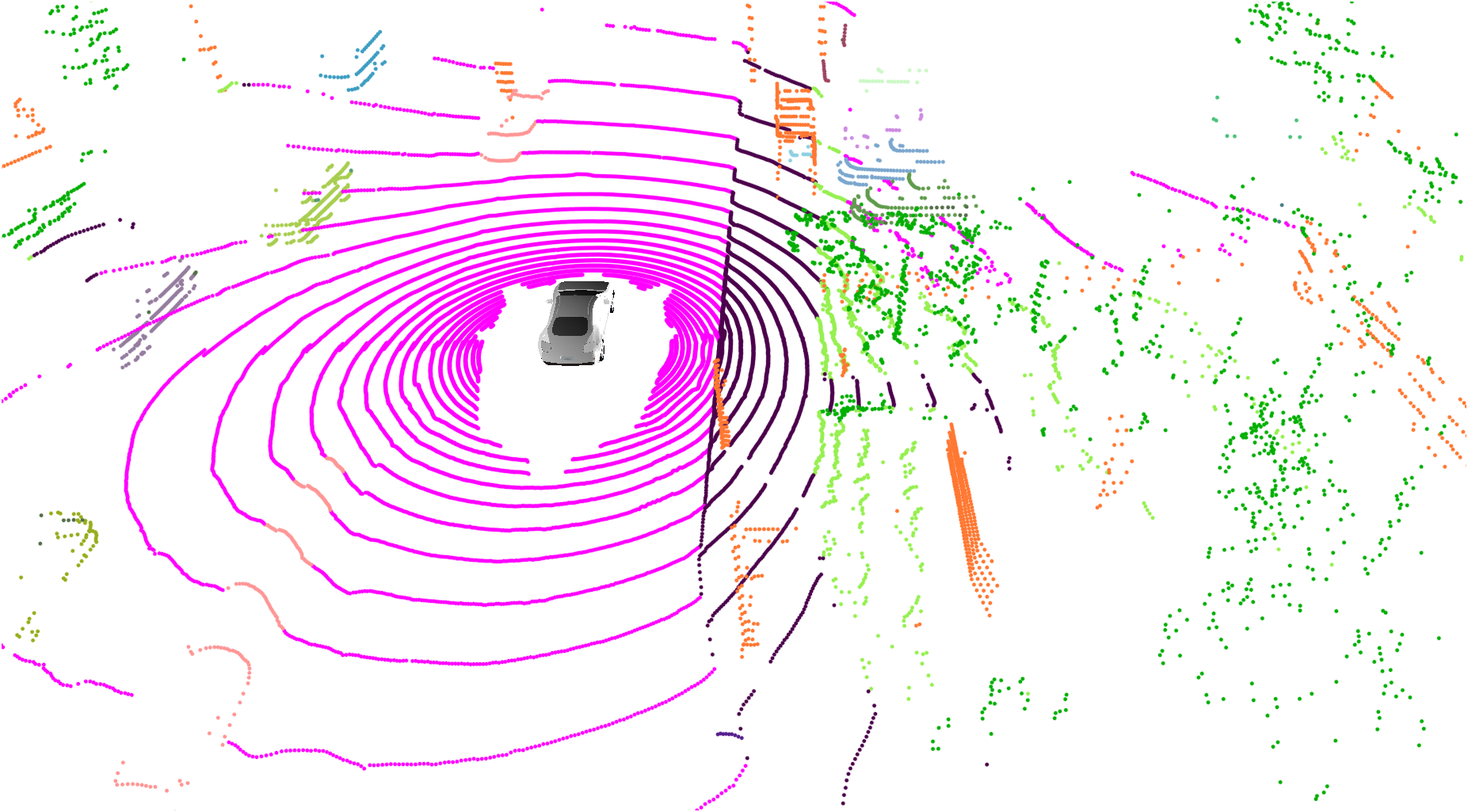}} & \raisebox{-0.4\height}{\includegraphics[width=\linewidth,frame]{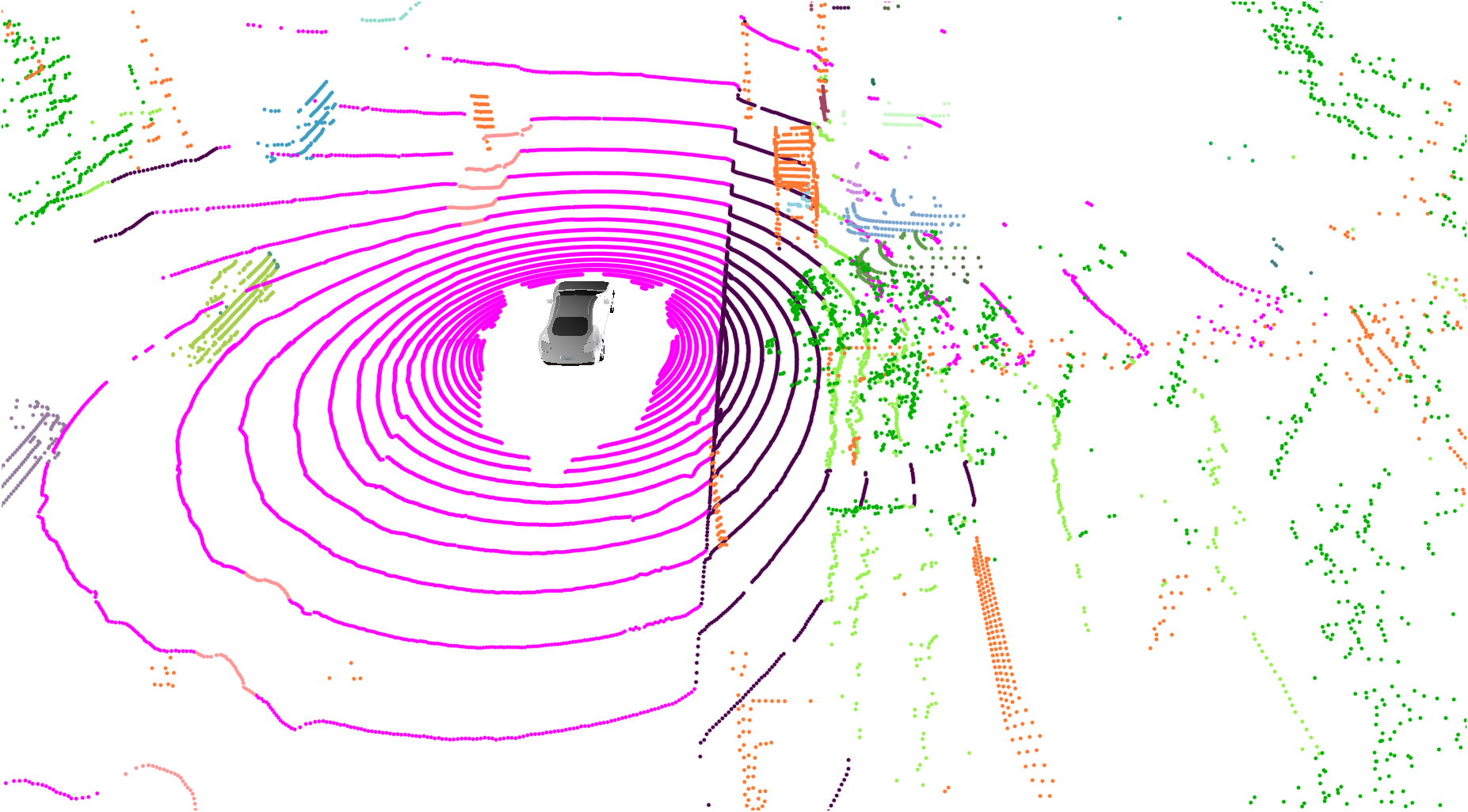}} \\
\\
{\rotatebox[origin=c]{90}{4D-PLS}}&\raisebox{-0.4\height}{\includegraphics[width=\linewidth,frame]{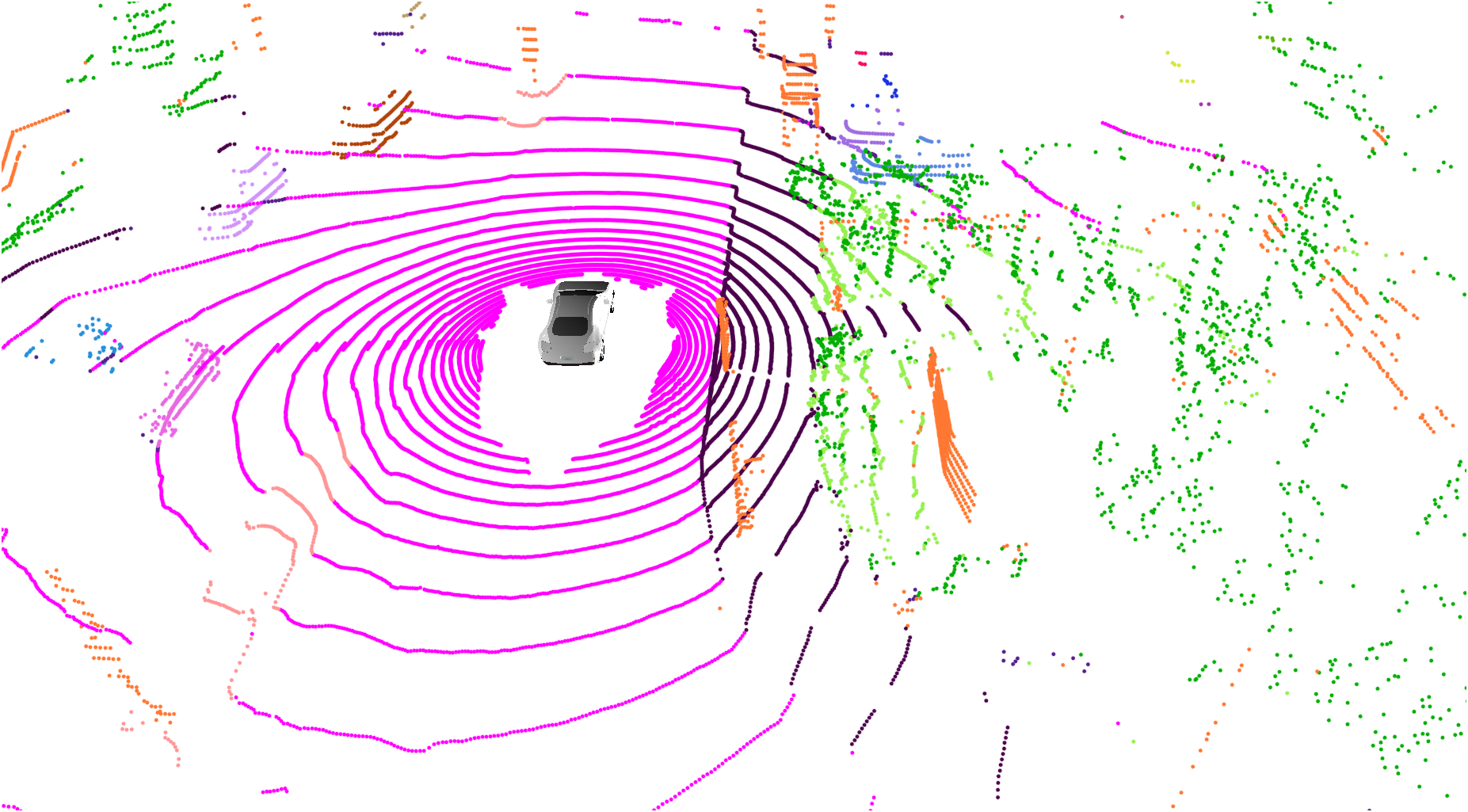}} & \raisebox{-0.4\height}{\includegraphics[width=\linewidth,frame]{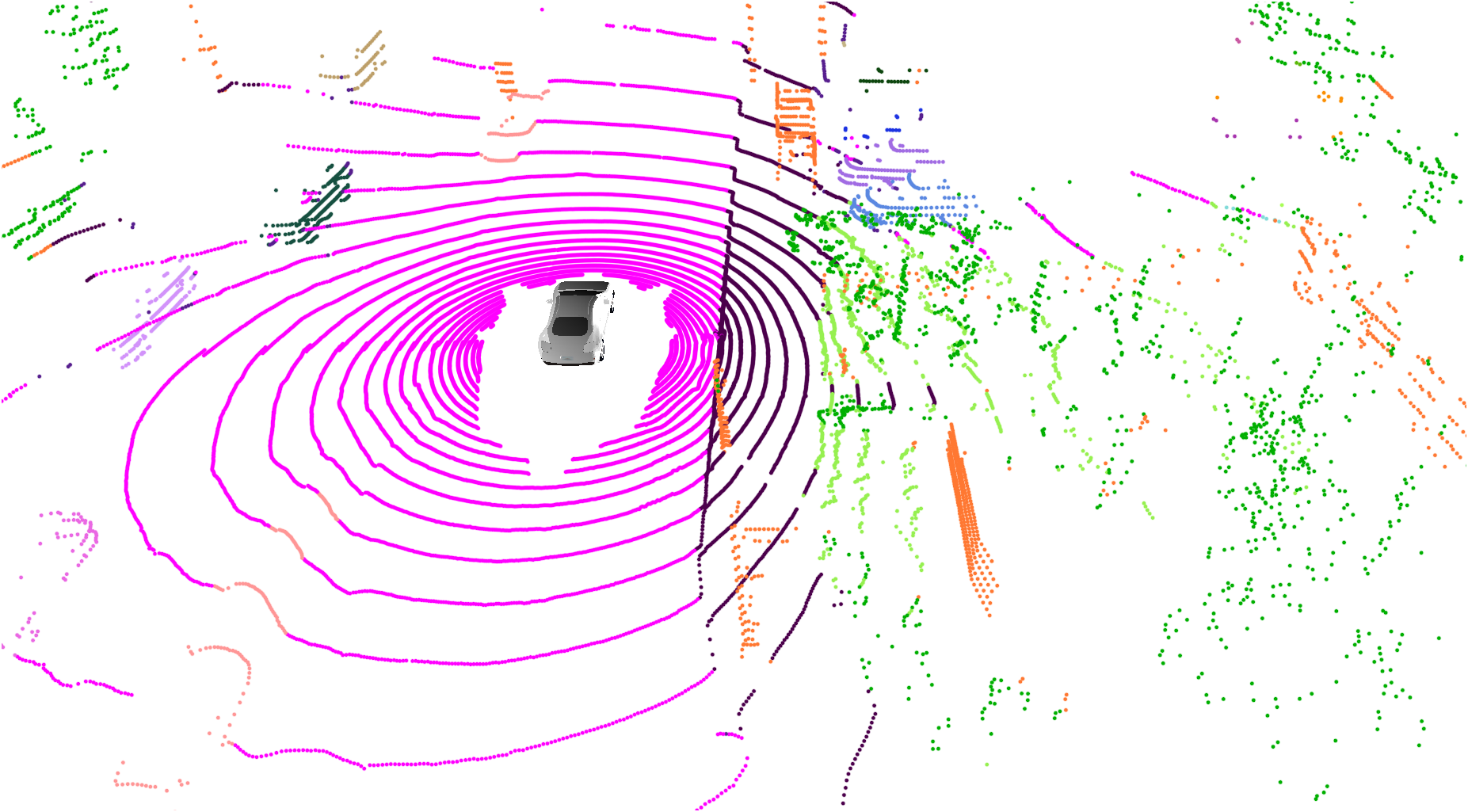}} & \raisebox{-0.4\height}{\includegraphics[width=\linewidth,frame]{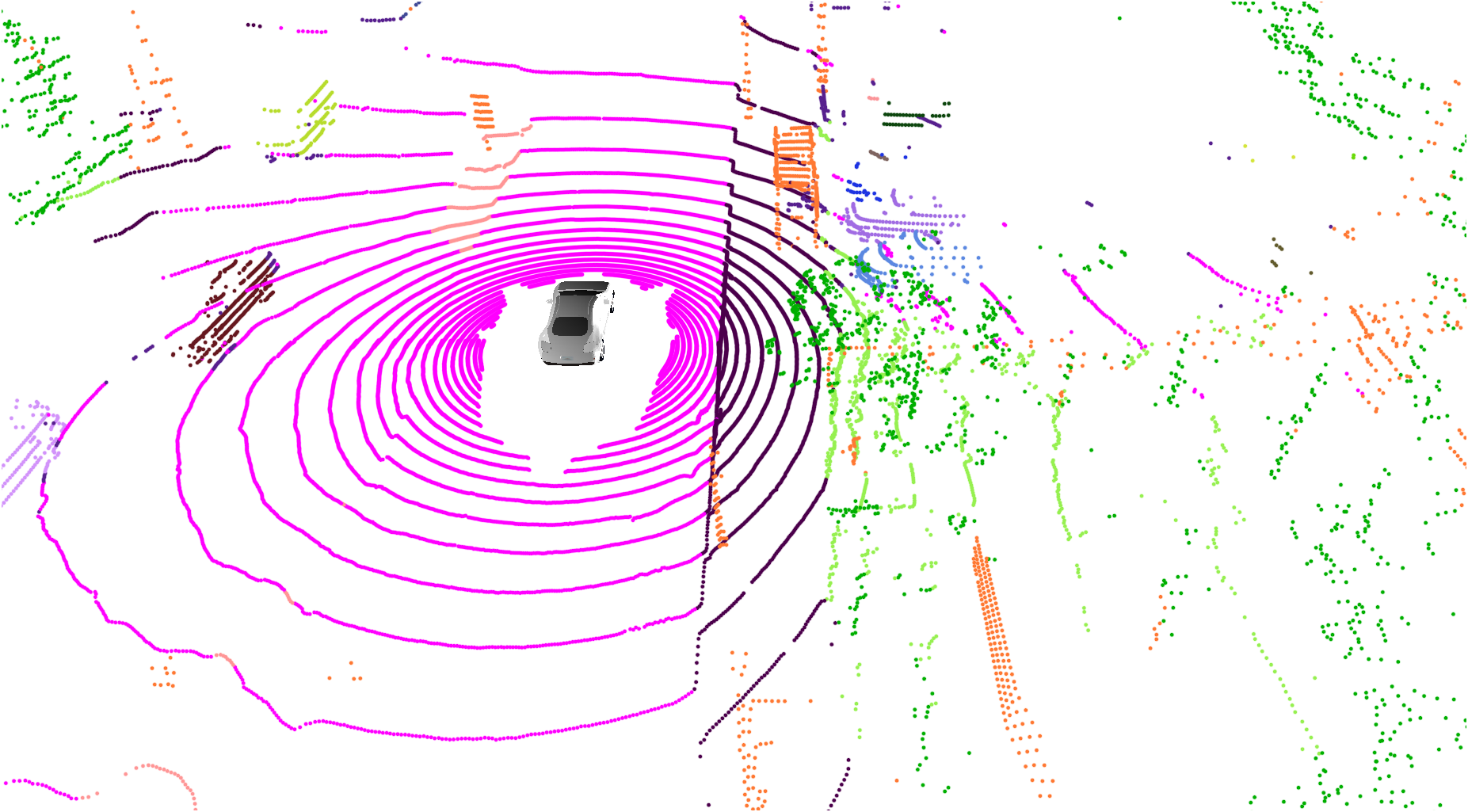}} \\
\\
{\rotatebox[origin=c]{90}{EfficientLPS + Kalman Filter}}&\raisebox{-0.4\height}{\includegraphics[width=\linewidth,frame]{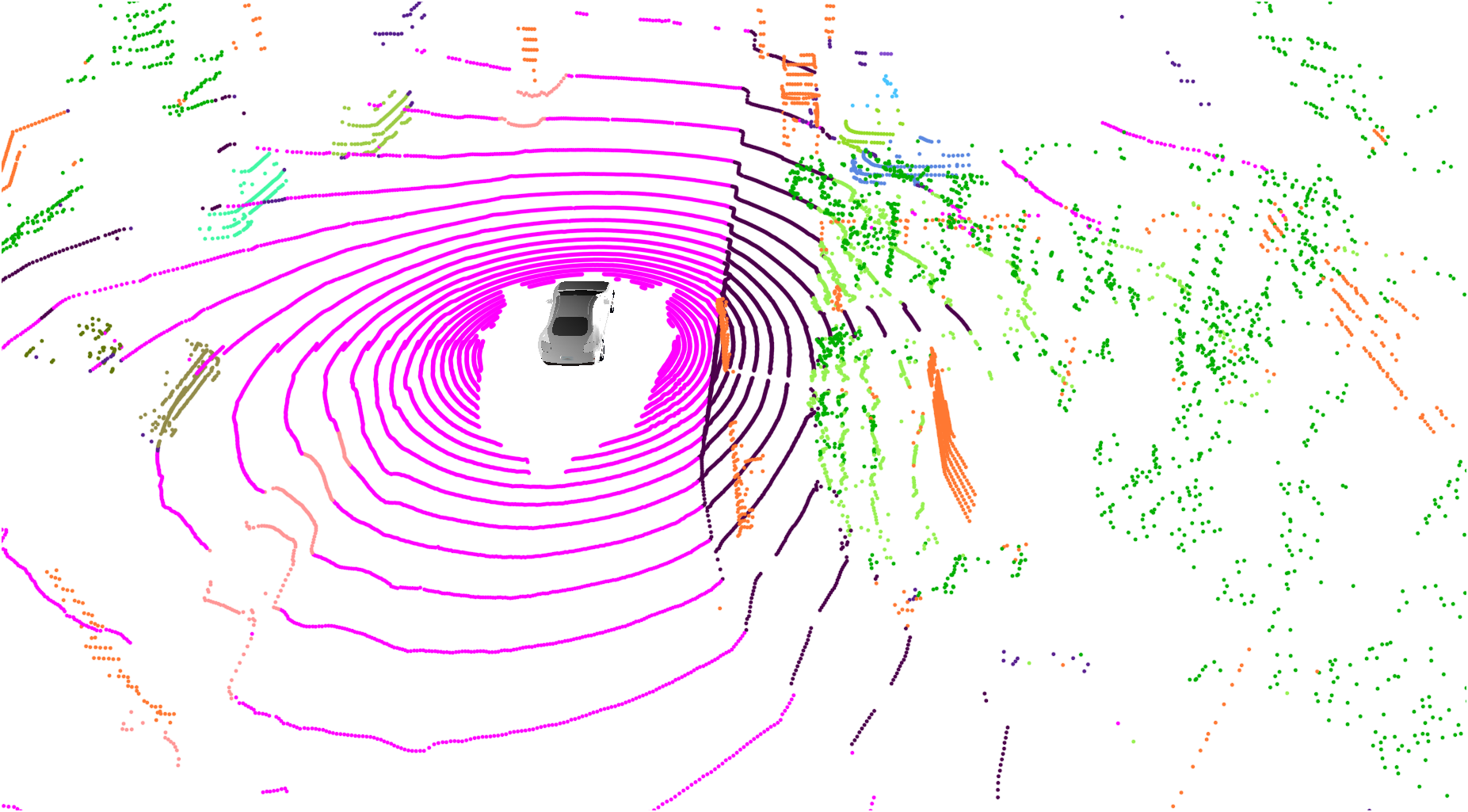}} & \raisebox{-0.4\height}{\includegraphics[width=\linewidth,frame]{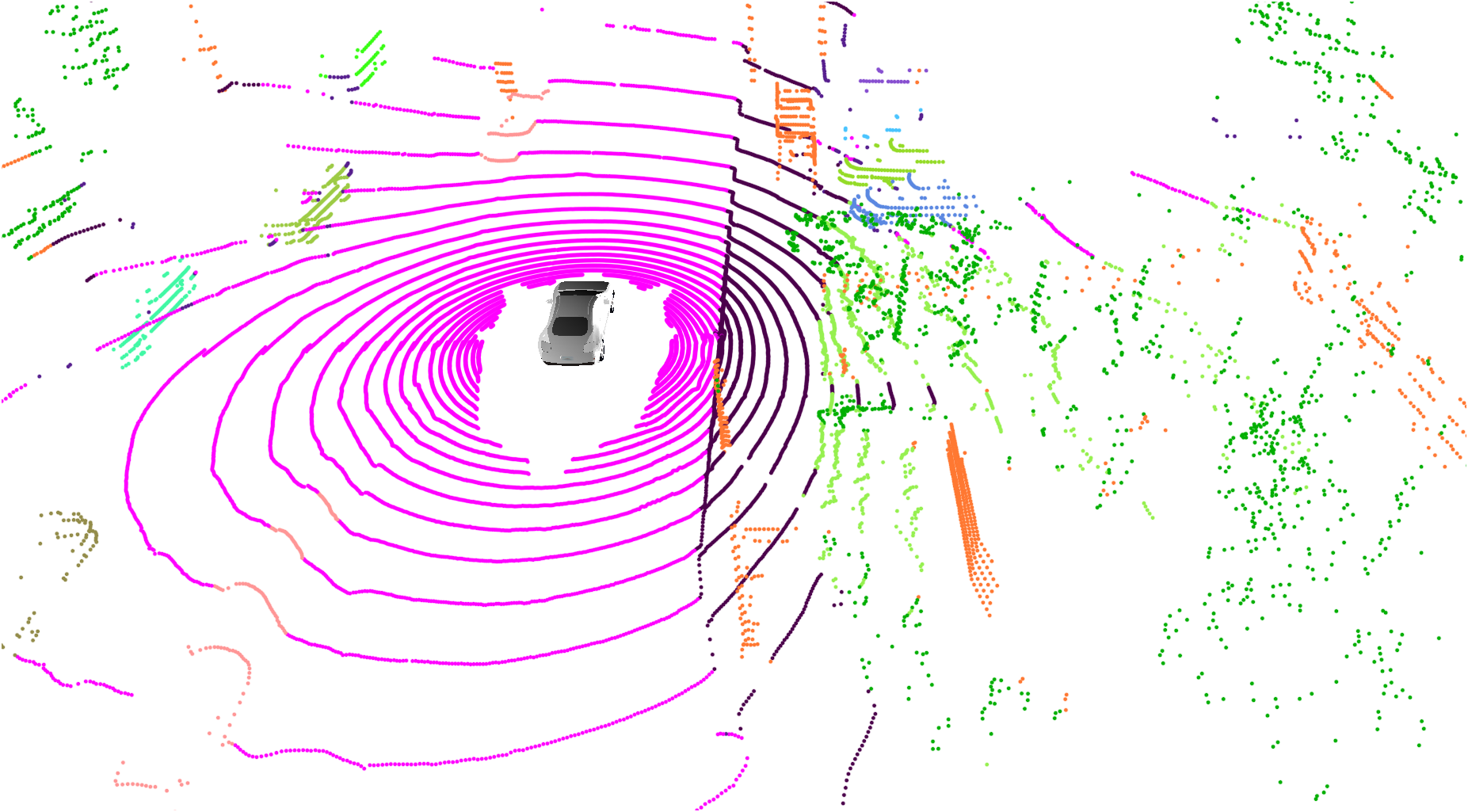}} & \raisebox{-0.4\height}{\includegraphics[width=\linewidth,frame]{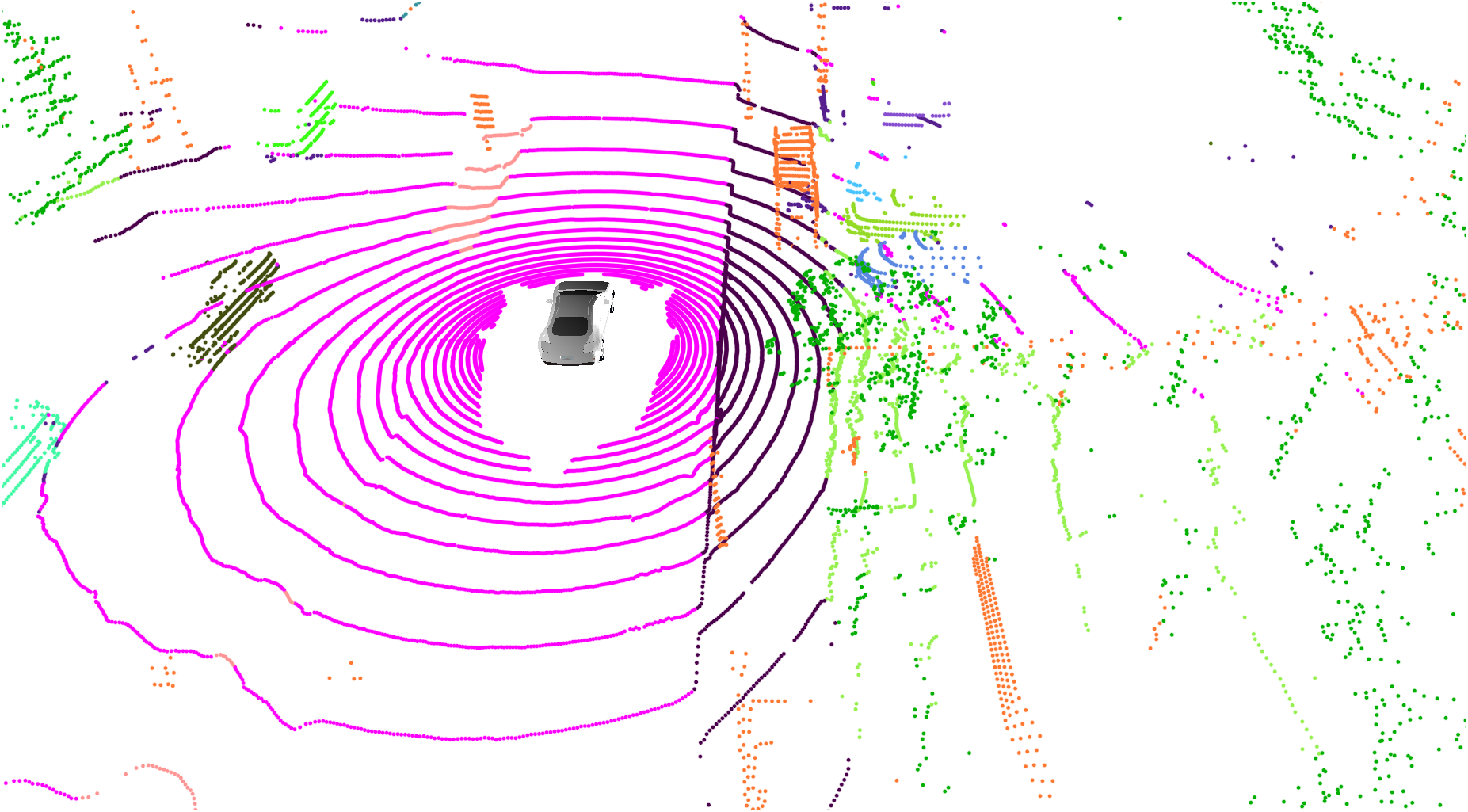}} \\
\end{tabular}}
\caption{Qualitative comparison of LiDAR panoptic tracking performance. We compare the best two end-to-end approaches, namely 4D-PLS and EfficientLPS + Kalman Filter. In example (a), 4D-PLS fails to track the parked truck and truck behind the ego car consistently, contrary to EfficientLPS + Kalman Filter that successfully tracks them. Similarly, in example (b), 4D-PLS fails to track the light green car on the left in groundtruth while EfficientLPS + Kalman Filter accurately does so. This shows that pairing a panoptic segmentation network with a filtering algorithm is a simple yet effective solution. Best viewed at $\times 4$ zoom.\looseness=-1}
\label{fig:tracking_ablation}
\end{figure*}

\noindent\textit{Panoptic Tracking}: In \secref{subsec:new_metric_analysis}, we compare our proposed metric, PAT, with other existing metrics which has been proposed for panoptic tracking. Here, we provide additional analysis for the LSTQ metric with respect to mIoU and AMOTA metrics. \figref{fig:mIoU_mAP_AMOTA_PQ_LSTQ}(b) gives a qualitative overview of how LSTQ varies with different panoptic tracking baselines which we generate by combining LiDAR semantic segmentation and tracking methods. As per \secref{subsec:baseline_results}, we used mIoU and AMOTA as a performance measure for the LiDAR semantic segmentation methods and the tracking methods respectively. From the 924 independently combined panoptic tracking baselines, we can see that there is generally a perceptual change in color across each row and column. This implies that LSTQ takes into account both the LiDAR semantic segmentation performance as well as the tracking performance. In \tableref{tab:correlations}, we show more quantitatively how various panoptic tracking metrics consider LiDAR semantic segmentation and tracking. PTQ demonstrates more bias towards LiDAR semantic segmentation as observed from the significantly stronger correlation between mIoU and PTQ in contrast to that between AMOTA and PTQ. On the other hand, LSTQ and PAT are more balanced metrics, with a smaller disparity in the respective correlations.

\section{Analysis of Panoptic Tracking Metrics}

The panoptic tracking task was initially introduced as MOPT~\cite{hurtado2020mopt} and simultaneously as video panoptic segmentation, 
each work proposes a new metric PTQ and VPQ, respectively. We decide to exclude VPQ for LiDAR panoptic tracking as it relies on 3D IoUs. Using more than four frames increases the difficulty of 3D IoU matching, and since this metric is initially proposed for images, this presents an additional limitation for LiDAR panoptic tracking evaluation. On the other hand, PTQ presents some drawbacks as pointed by \cite{weber2021step} and they introduce a new metric later adapted to LiDAR point clouds called LSTQ. This metric works at the pixel (or LiDAR point) level. We propose the PAT metric to address the drawbacks of our previously proposed PTQ metric and presents an opportunity to measure the task performance at the instance level. Furthermore, our PAT metric is able to penalize fluctuating track IDs as shown in \tabref{tab:metric}.

 \begin{figure*}
\centering
\footnotesize
{\renewcommand{\arraystretch}{1}
\begin{tabular}{P{0.4cm}P{5.5cm}P{5.5cm}P{5.5cm}}
&\raisebox{-0.4\height}{Ground Truth} & \raisebox{-0.4\height}{EfficientLPS} &  \raisebox{-0.4\height}{PolarSeg-Panoptic} \\
\\
{\rotatebox[origin=c]{90}{(a)}}&\raisebox{-0.4\height}{\includegraphics[width=\linewidth,frame]{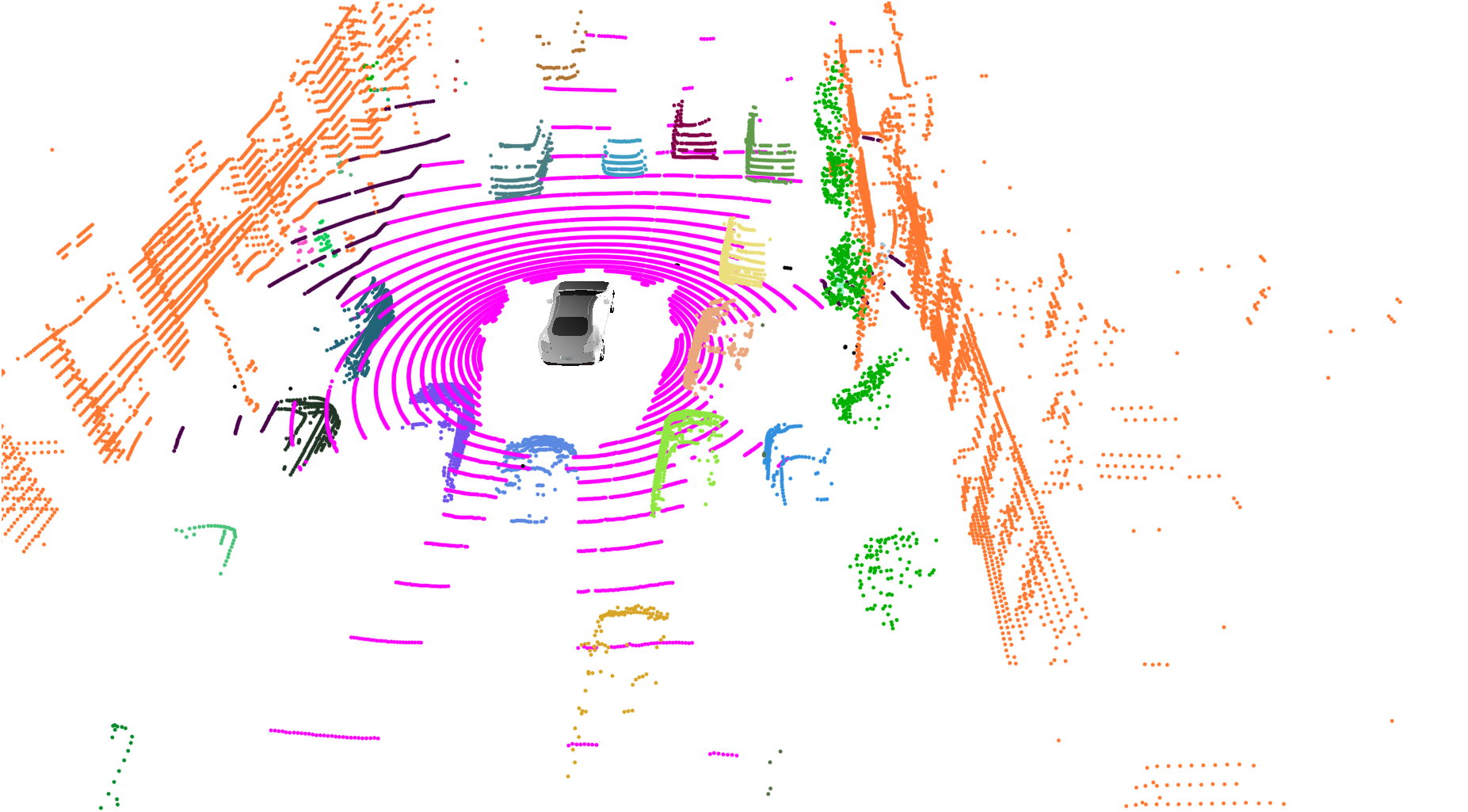}} & \raisebox{-0.4\height}{\includegraphics[width=\linewidth,frame]{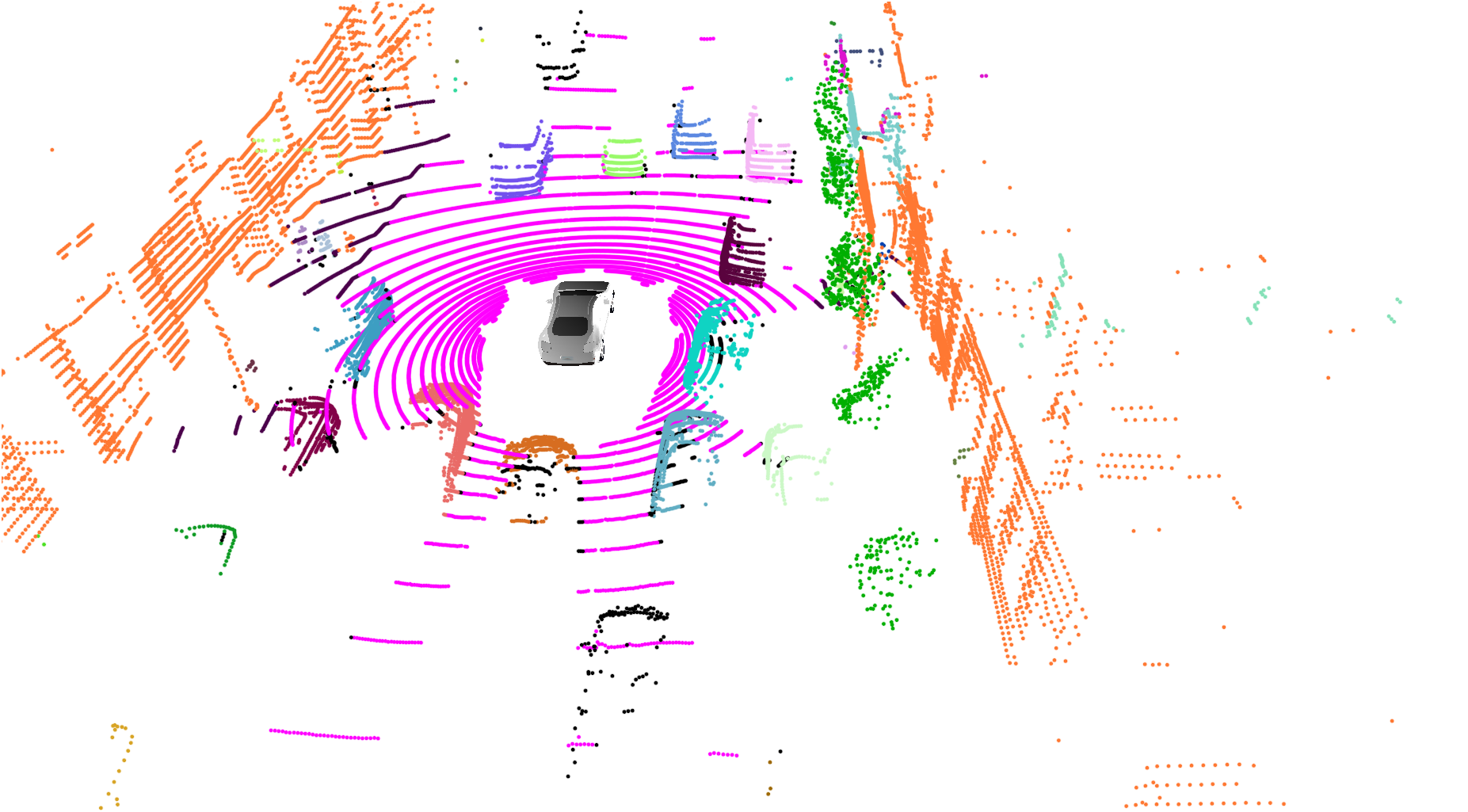}} & \raisebox{-0.4\height}{\includegraphics[width=\linewidth,frame]{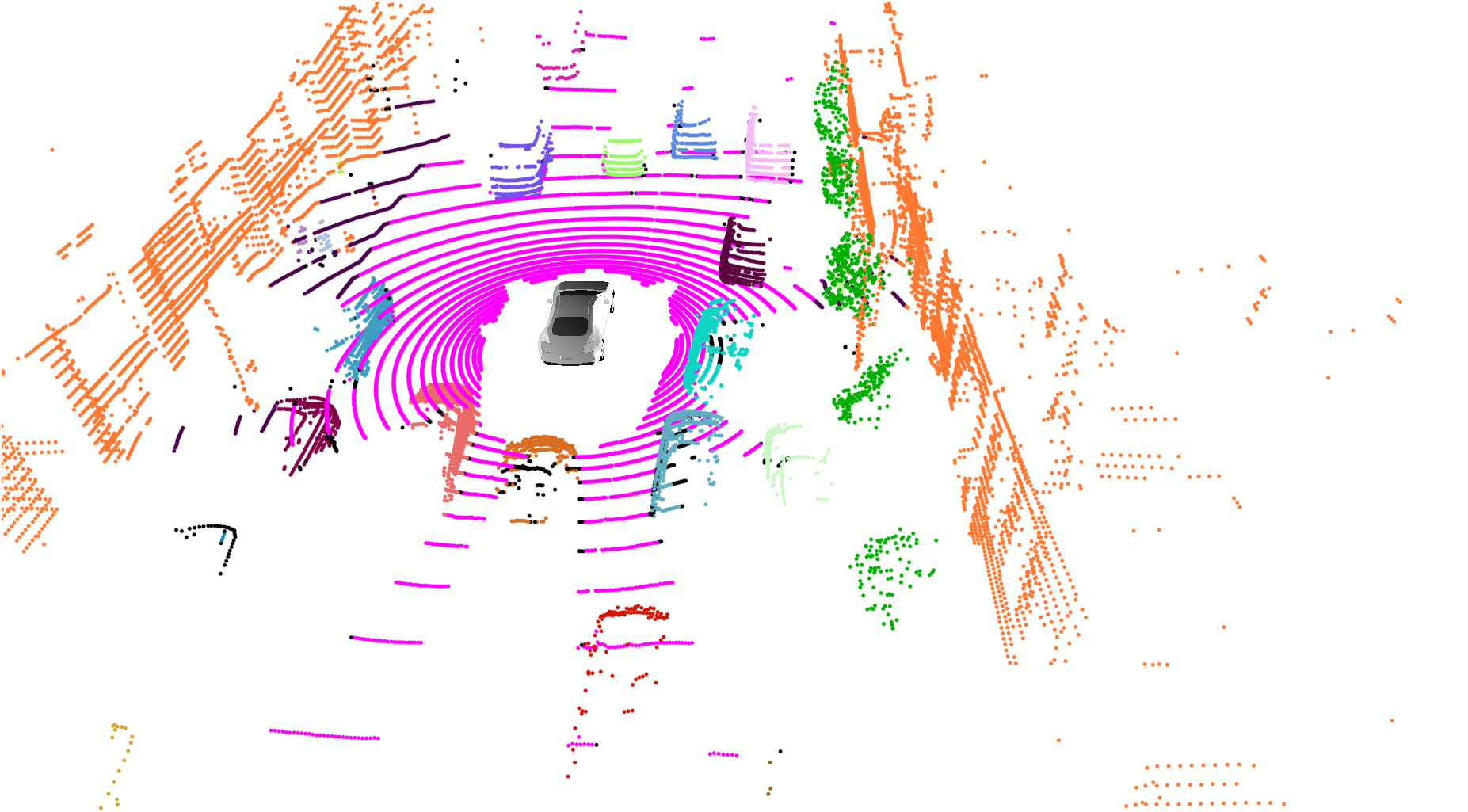}} \\
\\
{\rotatebox[origin=c]{90}{(b)}}&\raisebox{-0.4\height}{\includegraphics[width=\linewidth,frame]{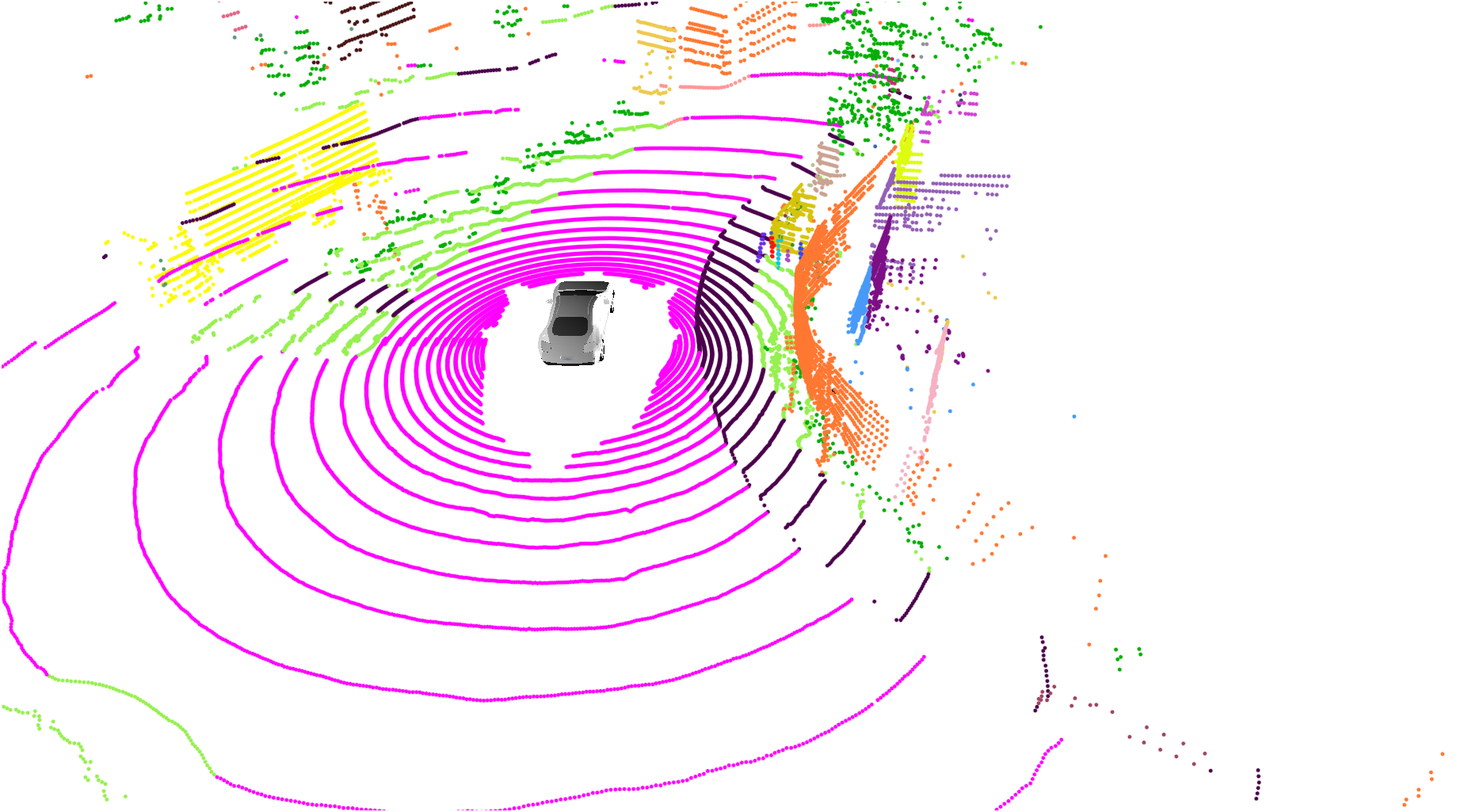}} & \raisebox{-0.4\height}{\includegraphics[width=\linewidth,frame]{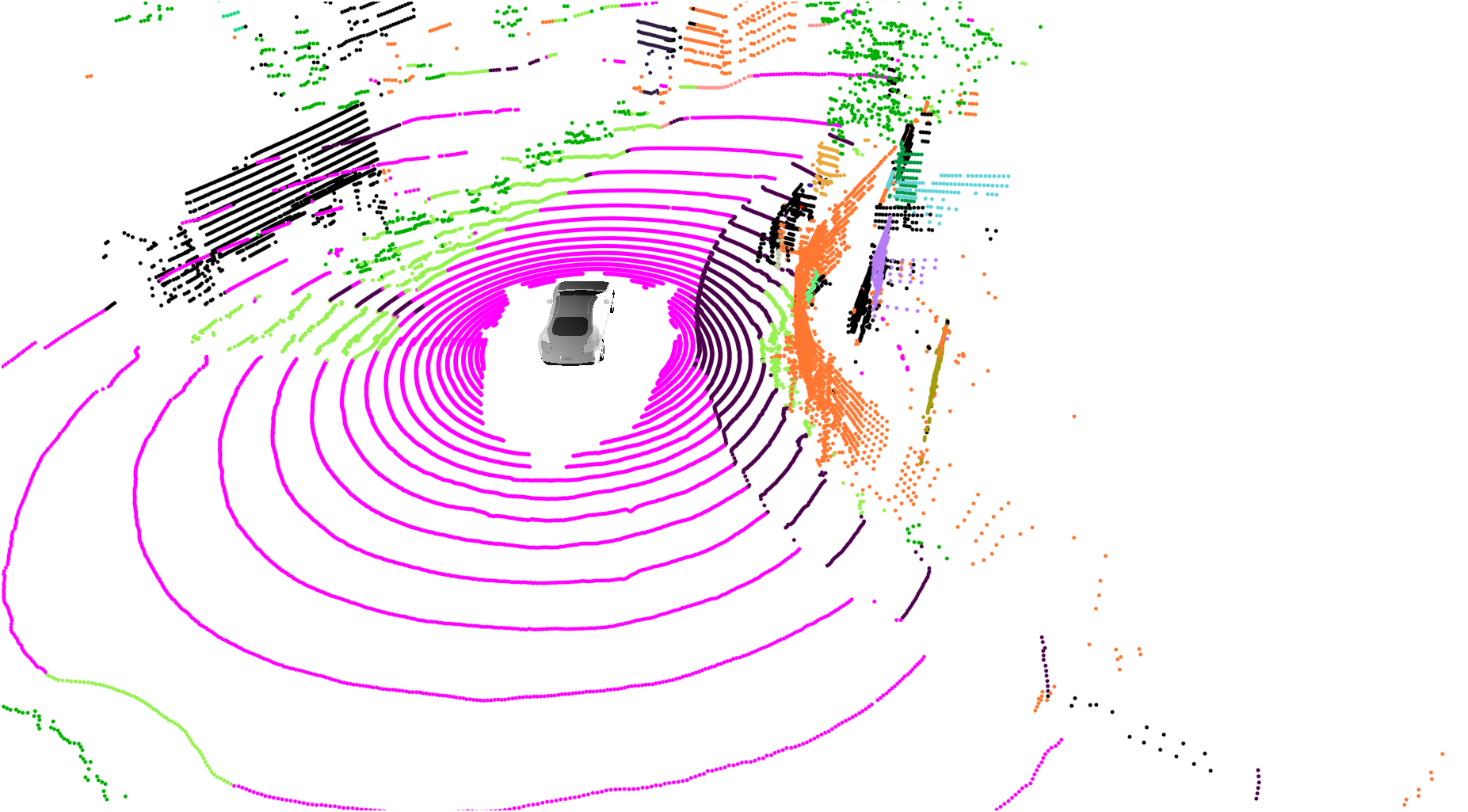}} & \raisebox{-0.4\height}{\includegraphics[width=\linewidth,frame]{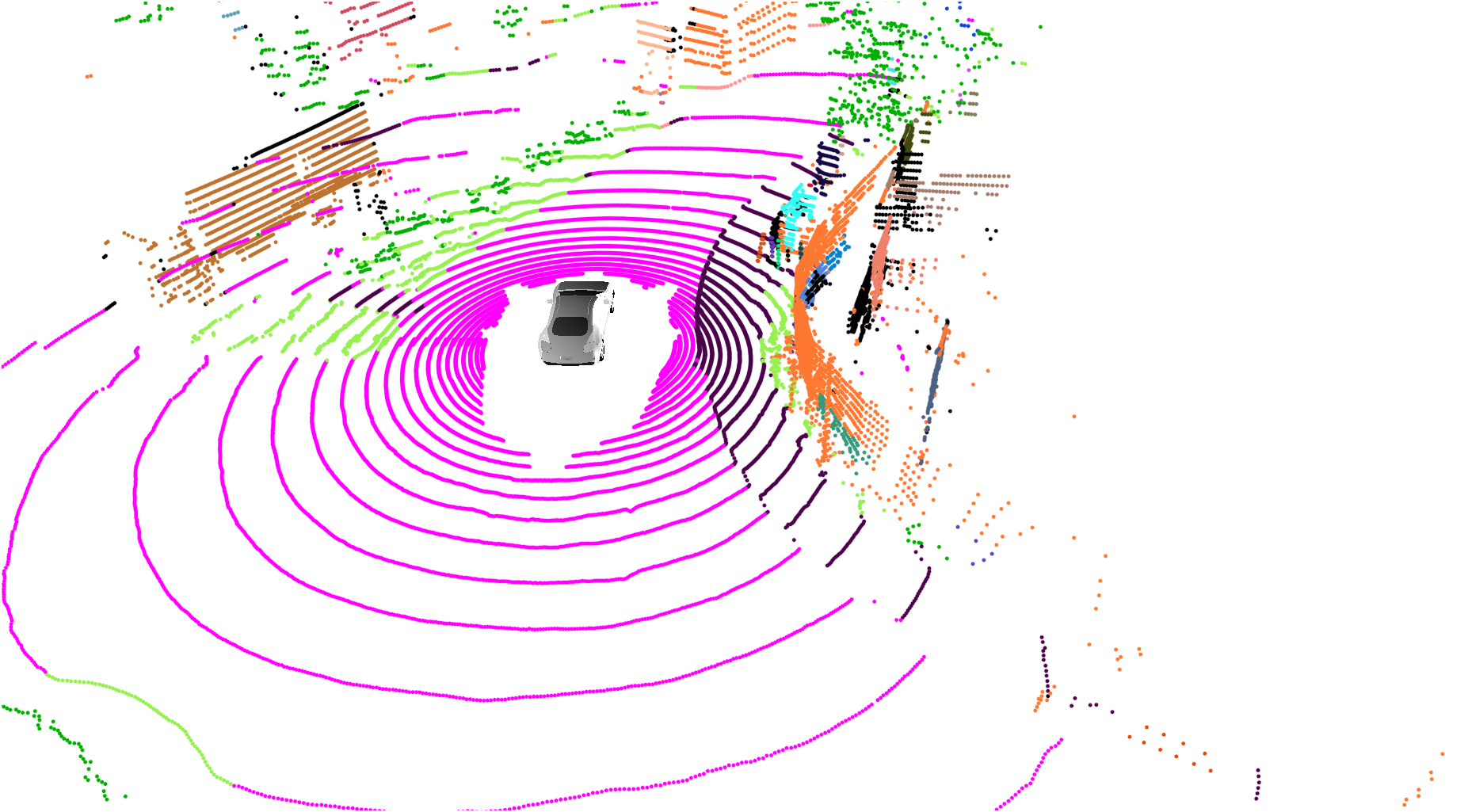}} \\
\\
\end{tabular}}
\caption{Qualitative comparison of LiDAR panoptic segmentation performance. We compare the two best end-to-end approaches, namely EfficientLPS and PolarSeg-Panoptic. In example (a), EfficientLPS misses the detection of the cars near the top and bottom end denoted by black point colours, whereas PolarSeg-Panoptic successfully segments them. Similarly, in (b), EfficientLPS misses the truck, whereas PolarSeg-Panoptic correctly segments it which shows the superiority of polar Bird's Eye View representation over scan unfolding projection. Best viewed at $\times 4$ zoom.}
\label{fig:panoptic_ablation}
\end{figure*}

To facilitate the comparison and analysis of our proposed \newmet metric, we provide a comparison of metrics related to panoptic tracking in \tableref{tab:metrics_abbre}.

\begin{table*}
\centering
\caption{Panoptic segmentation and tracking metrics, PQ is the panoptic segmentation metric, PTQ, LSTQ, and PAT are the panoptic tracking metrics.}
\label{tab:metrics_abbre}
\footnotesize
\begin{tabular}
{lllll}
\toprule
Metric & Decomposition & Description & Formula \\
\toprule
\multirow{2}{*}{PQ~\cite{kirillov2019panoptic}} & SQ & \textbf{S}emantic \textbf{Q}uality & $\sum_{(p,g)\in {TP}} IoU(p, g) \/ {|TP|}$ \\ 
& RQ & \textbf{R}ecognition \textbf{Q}uality & $|TP| \/ {|TP| + 0.5 |FP| + 0.5 |FN|}$ \\ 
& PQ & \textbf{P}anoptic \textbf{Q}uality  & $SQ \times RQ$ \\
\midrule
\multirow{2}{*}{PTQ~\cite{hurtado2020mopt}}
& IDS & \textbf{ID} \textbf{S}witches  & by counting \\
& PTQ & \textbf{P}anoptic \textbf{T}racking \textbf{Q}uality & $\frac{{\sum_{(p,g)\in {TP}} IoU(p, g)} - |IDS|}{|TP| + 0.5 |FP| + 0.5 |FN|}$ \\
\midrule
\multirow{3}{*}{LSTQ~\cite{aygun20214d}}
& $S_{cls}$ & \textbf{C}lassification \textbf{S}core  &  $\frac{1}{C}\sum_{c=1}^{C} IoU(c)$, C classes \\
& $S_{assoc}$ & \textbf{A}ssociation \textbf{S}core & Eq.~(4)-(7) in \cite{aygun20214d}    \\
& LSTQ & \textbf{L}idar \textbf{S}egmentation and \textbf{T}racking \textbf{Q}uality & $\sqrt{S_{score} \times S_{cls}}$    \\
\midrule
\multirow{3}{*}{\textbf{PAT}} & \textbf{PQ} & \textbf{P}anoptic \textbf{Q}uality & Eq.~(\ref{eq:track_quality_1}) and (~\ref{eq:track_quality_2})\\
& \textbf{TQ} & \textbf{T}racking \textbf{Q}uality & $\frac{2 \times PQ \times TQ}{PQ + TQ}$ \\
& \textbf{PAT} (Ours) & \textbf{PA}noptic \textbf{T}racking &  \\
\bottomrule
\end{tabular}
\end{table*}

\begin{table}
\centering
\caption{Comparison of results using different mean between PQ and TQ. The harmonic mean is the more strict operation.}
\label{tab:mean}
\footnotesize
\begin{tabular}{l|p{0.6cm}p{0.6cm}| p{0.6cm}p{0.6cm}| p{0.6cm}p{0.6cm}}
\toprule
 & PQ   50.0 & TQ   50.0 & PQ   90.0 & TQ   10.0  & PQ   90.0 & TQ   80.0 \\
\bottomrule
Mean & \multicolumn{2}{c|}{50.0} & \multicolumn{2}{c|}{50.0} & \multicolumn{2}{c}{85.0} \\
Geometric Mean & \multicolumn{2}{c|}{50.0} & \multicolumn{2}{c|}{30.0} & \multicolumn{2}{c}{84.9} \\
Harmonic Mean & \multicolumn{2}{c|}{50.0} & \multicolumn{2}{c|}{18.0} & \multicolumn{2}{c}{84.7} \\
\bottomrule
\end{tabular}
\begin{tabular}{p{1cm}}
\vspace{0.2cm}\\
\end{tabular}
\end{table}

The main differences between the LSTQ and PAT metrics are:
\begin{enumerate}
    \item PAT provides the evaluation at the instance level while LSTQ works at the point level.
    \item Both metrics are based on separately obtaining panoptic and tracking evaluations. This allows straightforward interpretation in comparison to PTQ. Subsequently, PAT computes the harmonic mean, whereas LSTQ computes the geometric mean. As shown in \tabref{tab:mean}, PAT is more strict when one of the values is low.
    \item As demonstrated in \tabref{tab:metric}, we propose the Tracking Quality score for PAT, that penalizes fragmented tracks while LSTQ is not able to penalize for this mistakes.
\end{enumerate}

Both the PAT and LSTQ metrics are suitable for penalizing different aspects of the task. Therefore, both metrics are useful in different scenarios, similar to tracking task where we have HOTA and MOTA metrics.

\section{Impact of Diverse Scenes in the Training Set}

In this section, we provide more details on our study of generalization ability of an approach trained on a dataset consisting of diverse scenes and objects in~\secref{sec:generalize}. For this experiment, we employ EfficientLPS as our panoptic segmentation network and train two instances of it on SemanticKITTI and Panoptic nuScenes datasets, respectively. We use the same training protocols as described in~\cite{mohan2020efficientps}. We evaluate the two EfficientLPS models on the unseen validation set of PandaSet. Panoptic nuScenes employs 32 channels LiDAR sensor with point cloud density of $35k$, whereas Semantic KITTI employs 64 channels LiDAR sensor with point cloud density of $122k$. Moreover, PandaSet uses 64 channels with point cloud density of $169k$. To facilitate use of all the datasets, we restrict the classes for the panoptic segmentation task to a subset of classes that are common among these datasets. Thus, for this experiment, the \textit{thing} classes comprise bicycle, bus, car, motorcycle, person, and truck, while road, sidewalk, vegetation and building form the \textit{stuff} classes. It should be noted that PandaSet provides 3D bounding box and semantic segmentation labels but not official panoptic segmentation labels. Hence, we compute the panoptic segmentation labels for this dataset by following the annotation protocol described in~\secref{sec:annotation_protocol}. To describe briefly, we combine the point-level labels with the 3D bounding boxes to obtain instance labels for each point. Here, an instance consists of the points that fall within a 3D bounding box and have the same segmentation type as the box. In case of overlapping points due to bounding box overlaps, we treat them as noise or void labels and are not considered during evaluation. 

\section{Qualitative Evaluations}

In this section, we qualitatively compare the best two end-to-end approaches for both panoptic segmentation and panoptic tracking tasks. \figref{fig:panoptic_ablation} presents the qualitative evaluation for the task of panoptic segmentation. For the aforementioned task, PolarSeg-Panoptic achieves the highest PQ score followed by EfficientLPS among the end-to-end approaches. In~\figref{fig:panoptic_ablation}~(a) EfficientLPS fails to detect the two cars, one towards the top and the other towards the bottom end, whereas PolarSeg-Panoptic correctly segments them. Similarly, in~\figref{fig:panoptic_ablation}~(b) EfficientLPS misses the truck instance on the left contrary to PolarSeg-Panoptic that segments it. The two examples of missed detection reasonably imply that using polar Bird's Eye View representation over scan unfolding projection is an optimal choice. Following, \figref{fig:tracking_ablation} presents the qualitative comparison for panoptic tracking task. EfficientLPS + Kalman Filter approach achieves the highest PAT score for the task of panoptic tracking, whereas 4D-PLS attains the second highest score. Thus, we compare these two solutions. In~\figref{fig:tracking_ablation}~(a), 4D-PLS fails to track the parked truck and truck behind the ego car in frame t-1 and t, respectively. Similarly, in \figref{fig:tracking_ablation}~(b), 4D-PLS is not able to track the light green car on the left in the groundtruth. In contrast, EfficientLPS + Kalman Filter performs panoptic tracking consistently for both the examples. From the results, we can safely infer that using filtering algorithm for tracking in conjunction with a panoptic segmentation network can yield a simple yet effective panoptic tracking solution.

\end{document}